\PassOptionsToPackage{table}{xcolor}

\documentclass{egpubl}
\usepackage{eg2027}[preprint]
 
\ConferenceSubmission   
\usepackage[T1]{fontenc}
\usepackage{microtype}
\usepackage{dfadobe}  

\usepackage{cite}  
\BibtexOrBiblatex
\electronicVersion
\PrintedOrElectronic
\ifpdf \usepackage[pdftex]{graphicx} \pdfcompresslevel=9
\else \usepackage[dvips]{graphicx} \fi

\usepackage{egweblnk} 
\usepackage{booktabs}
\usepackage{multirow}
\usepackage{subcaption}
\usepackage{tikz}
\usepackage{tabularx}
\usetikzlibrary{spy}

\usepackage[table]{xcolor}
\definecolor{tabfirst}{rgb}{1, 0.7, 0.7} 
\definecolor{tabsecond}{rgb}{1, 0.85, 0.7} 
\definecolor{tabthird}{rgb}{1, 1, 0.7} 

\usepackage{pifont}
\newcommand{\cmark}{\ding{51}}%

\usepackage{amsmath}
\usepackage{amssymb}
\usepackage{bm}
\usepackage{xspace}
\newcommand{\ie}{\textit{i.e.}\xspace}

\newcommand{\cf}{cf.\xspace}
\renewcommand{\paragraph}{\noindent\textbf}

\title[Efficient Confidence-Aware Densification for 3DGS]%
      {Efficient Confidence-Aware Densification for 3DGS}

\author[M. Hofer~\textit{et al.}]
{\parbox{\textwidth}{
\centering Manuel Hofer$^{1}$\orcid{0009-0006-1463-6589}
        \enspace Thomas K\"ohler$^{1}$\orcid{0009-0004-2685-0502}
        \enspace Felix Windisch$^{1}$\orcid{0009-0001-8174-9907}
        \enspace Bernhard Kerbl$^{2}$\orcid{0000-0002-5168-8648}
        \enspace Markus Steinberger$^{1,3}$\orcid{0000-0001-5977-8536}
        \enspace Lukas Radl$^{1}$\orcid{0009-0008-4075-5877}
        }
        \\
{\parbox{\textwidth}{\centering 
$^1$Graz University of Technology, Austria\enspace
$^2$University of Copenhagen, Denmark\enspace
$^3$Huawei Technologies, Austria
       }
}
}

\begin{document}

\teaser{
\vspace{-0.75cm}
\includegraphics[width=0.995\linewidth]{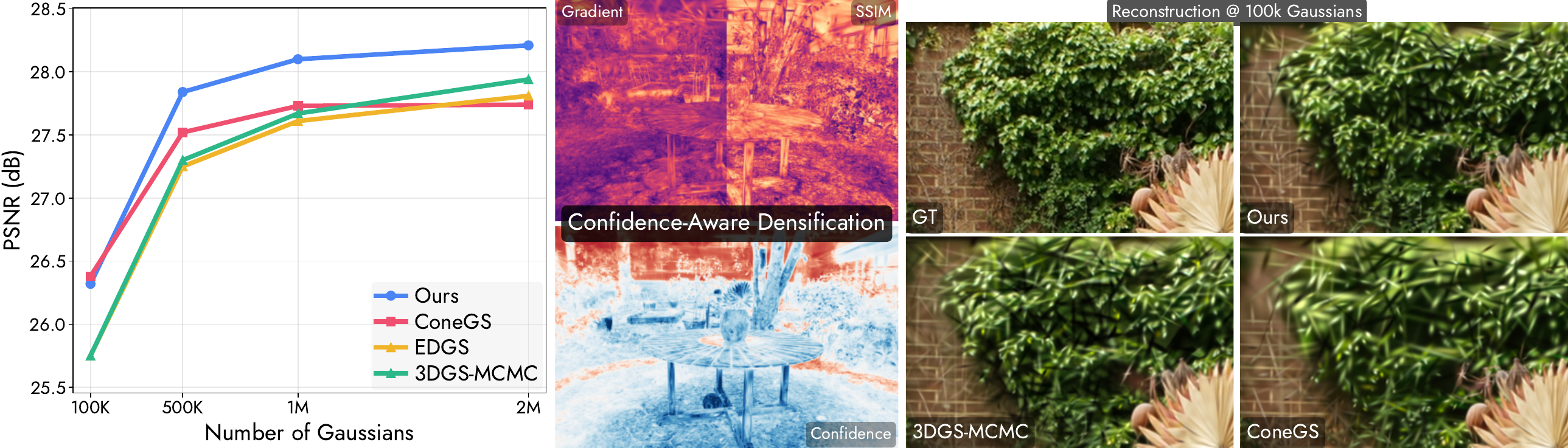}
 \centering
  \caption{
\textbf{Confidence-Aware Densification}:
Left: average PSNR over the primitive budget for the Mip-NeRF 360 dataset, for our method, 3DGS-MCMC~\cite{kheradmand2024mcmc}, EDGS~\cite{Kotovenko2026edgs} and ConeGS~\cite{baranowski2026conegs}; our gain over 3DGS-MCMC is largest at the two lowest budgets and persists at $\mathrm{2M}$, where both prior-based methods fall below 3DGS-MCMC.
The middle panels show where our densification for 3D Gaussian Splatting~\cite{kerbl20233dgs} adds primitives, based on absolute screen-space gradients, a per-primitive SSIM error and a learned confidence.
Right: With $\mathrm{100k}$ primitives, our method preserves visibly higher detail in a background region.
}
\label{fig:teaser}
}

\maketitle
\thispagestyle{plain}
\makeatletter
\def\@oddhead{}\def\@evenhead{}
\def\@oddfoot{\hfil\thepage\hfil}\def\@evenfoot{\hfil\thepage\hfil}
\makeatother
\begin{abstract}
3DGS-MCMC has replaced the gradient-driven density control of 3D Gaussian Splatting with opacity-based sampling and relocation, and is now the de facto standard for prior-free densification under a fixed primitive budget.
Because opacity stays high where the scene is already well reconstructed, this sampling sends most new primitives to the foreground and leaves the background under-reconstructed.
At low budgets, prior-based methods such as ConeGS and EDGS outperform it, at the cost of a pre-trained radiance field or dense image matches and a separate initialization stage.
We argue for a comeback of error-driven densification within the fixed-budget MCMC framework.
Our confidence-aware densification keeps the fixed budget and gradual growth of 3DGS-MCMC, but places primitives by the image error instead of the opacity.
New primitives are drawn mostly where screen-space gradients are large.
We favor primitives with a low learned confidence, which marks regions the reconstruction still explains poorly, and relocate the primitives that contribute least to the rendered images to these places.
Averaged over 24 scenes, our method outperforms 3DGS-MCMC in PSNR, SSIM and LPIPS at every budget from $\mathrm{100k}$ to $\mathrm{2M}$ primitives.
At $\mathrm{100k}$, without a prior or an initialization stage, it recovers 0.84\,dB of the 1.05\,dB PSNR by which ConeGS leads 3DGS-MCMC, and from $\mathrm{500k}$ on, it has the highest PSNR and SSIM of all fixed-budget methods we compare with, including ConeGS and EDGS.
While inherently prior-free, our densification can also exploit priors: from the ConeGS or EDGS initialization, it matches ConeGS in PSNR and SSIM at $\mathrm{100k}$ and is best in all three metrics at $\mathrm{500k}$.

\ccsdesc[500]{Computing methodologies~Reconstruction}
\ccsdesc[300]{Computing methodologies~Rasterization}
\ccsdesc[100]{Computing methodologies~Parallel algorithms}

\printccsdesc   
\end{abstract}  

\section{Introduction}

Following Neural Radiance Fields~\cite{Mildenhall2020NeRF}, 3D Gaussian Splatting (3DGS)~\cite{kerbl20233dgs} has become a standard representation for novel view synthesis, as its explicit primitives render high-quality images at real-time frame rates.
Much of the quality of 3DGS comes from adaptive density control (ADC)~\cite{kerbl20233dgs}, which clones and splits primitives where the screen-space gradients exceed a pre-defined threshold.
Later work refined this error signal with absolute gradients~\cite{yu2024gof, ye2024absgs} and a per-primitive SSIM error~\cite{bulo2024revising}.
ADC is robust across a wide range of scenes, but it relies on an initialization from sparse structure-from-motion (SfM) points.
In contrast to Neural Radiance Fields, 3DGS only receives gradients where primitives exist, and because ADC clones and splits existing primitives, it cannot place new ones far from them.
ADC also offers no direct control over the final number of primitives.

When the number of primitives is capped by a fixed budget, quality depends on where densification places them.
3DGS-MCMC~\cite{kheradmand2024mcmc} showed that densification can succeed even from random initialization.
It reframes densification as the relocation of low-opacity primitives to the positions of high-opacity ones, with noise added to the positions of all primitives, scaled so that mainly low-opacity primitives move, to drive exploration.
Due to its robustness and its explicit control over the primitive count, this framework has become the de facto standard for prior-free densification and is adopted by recent works~\cite{condor2026nht, liu2025beta}.
At low budgets, however, prior-based methods outperform it: ConeGS~\cite{baranowski2026conegs} initializes from a pre-trained Instant-NGP~\cite{mueller2022instant} (iNGP) radiance field and uses it to place primitives at high-error pixels, and EDGS~\cite{Kotovenko2026edgs} initializes from dense image matches, both at the cost of a prior and a separate initialization stage.
With more primitives, this advantage turns into a disadvantage: on Mip-NeRF 360, both ConeGS and EDGS end up below 3DGS-MCMC at $\mathrm{2M}$ (\cf Figure~\ref{fig:teaser}).
A prior thus gives a head start when few primitives have to be placed well, but not a better final result.
With few primitives, we find that 3DGS-MCMC under-reconstructs the background.
It draws the source primitives, at which new primitives are placed, in proportion to opacity, and it relocates primitives whose opacity has decayed below a threshold, so neither rule is tied to the image error.
Because opacities stay high in regions that many training views observe, both rules favor the already well-reconstructed foreground, and the imbalance compounds while the budget fills (\cf Figure~\ref{fig:sampling_bias}).

We therefore argue for a comeback of error-driven densification within the MCMC framework.
We keep the fixed budget, the gradual growth and the relocation step of 3DGS-MCMC, and change the two decisions it bases on opacity: where new primitives go, and which make room for them.
New primitives go where the image error is large, which we measure mainly with screen-space gradients and partly with a structural error.
The gradients are weighted toward primitives of low confidence, a per-primitive value that CoMe~\cite{radl2026come} learns for surface reconstruction and that drops where the reconstruction remains poor.
Room is made by the primitives that contribute least to the rendered images, rather than by those whose opacity has decayed.
The two changes only work together: either one alone barely improves 3DGS-MCMC (Section~\ref{sec:analysis}).
New and relocated primitives are placed once by sampling around their source and take their color from a training image, which replaces the continuous position noise of 3DGS-MCMC.
Apart from one confidence per primitive and one loss term, we need no additional model or training stage.

Our method outperforms 3DGS-MCMC at every budget we test, most clearly when primitives are scarce, and it reconstructs backgrounds in more detail (\cf Figure~\ref{fig:teaser}): on Mip-NeRF 360, most of its gain over 3DGS-MCMC lies in the background (Tables~\ref{tab:background} and~S.11).
Without a prior, it closes most of the gap to ConeGS at low budgets, and for larger budgets, it reaches the highest PSNR and SSIM of all fixed-budget methods we compare with. %
Priors and our densification are not mutually exclusive: started from the initialization of a prior-based method, our method matches or exceeds that method in PSNR and SSIM.
It also keeps the robustness of 3DGS-MCMC to random initialization.
Because it needs no initialization stage, it trains in under 3\,min at $\mathrm{500k}$ primitives on Mip-NeRF 360, while ConeGS needs 14\,min, 2.2\,min of it for initialization.

\noindent Our contributions are:\vspace{-4pt}
\begin{itemize}
    \item An analysis showing that the opacity-driven sampling and relocation of 3DGS-MCMC disproportionately concentrate primitives in the well-reconstructed foreground and leave backgrounds under-reconstructed.
    \item Confidence-aware densification for the fixed budget of 3DGS-MCMC, which places new primitives by the image error, weighted by a learned confidence, and relocates the primitives that contribute least, together with an ablation showing that neither change improves 3DGS-MCMC on its own.
    \item Position sampling from the source Gaussian with color initialization from the training images, which replaces the continuous position noise of 3DGS-MCMC.
\end{itemize}

\section{Related Work}

\paragraph{Radiance Fields.}
3D Gaussian Splatting~\cite{kerbl20233dgs} has become the standard representation used for novel view synthesis.
Recent follow-up works address artifact-free rendering~\cite{radl2024stopthepop, steiner2025aaagaussians, Yu2024MipSplatting, hahlbohm2025perspective}, better view-dependent appearance~\cite{disario2025sphericalvoronoi, condor2026nht, liu2025beta}, more lightweight models~\cite{Fang2024minisplatting2, papantonakis2024reducing, fan2023lightgaussian} and faster optimization~\cite{chen2025dashgaussian, hahlbohm2026fastergs}.
Beyond novel view synthesis, 3DGS is used for surface extraction~\cite{radl2026come, Radl2025SOF, yu2024gof, guedon2025milo}.
Alternative representations that target fully correct volumetric rendering~\cite{mai2024ever, mai2026radiance, govindarajan2025radfoarm, govindarajan2026powerfoam} have also been investigated;
note that their quality also depends on densification.

{
\newcommand{\overlaylabel}[2]{%
    \begin{tikzpicture}
        \node[anchor=south west, inner sep=0] (image) at (0,0) {%
            \includegraphics[width=0.49\linewidth]{#1}%
        };
        \begin{scope}[x={(image.south east)},y={(image.north west)}]
            \node[anchor=south east, 
                  fill=black, fill opacity=0.45, 
                  inner sep=2.5pt, 
                  text opacity=0.8,
                  rounded corners=1.5pt,
                  align=right] 
            at (0.98, 0.02) {%
                {\sffamily\scriptsize\textbf{\textcolor{white}{#2}}}
            };
        \end{scope}
    \end{tikzpicture}%
}
\begin{figure}[t!]
    \centering
    \overlaylabel{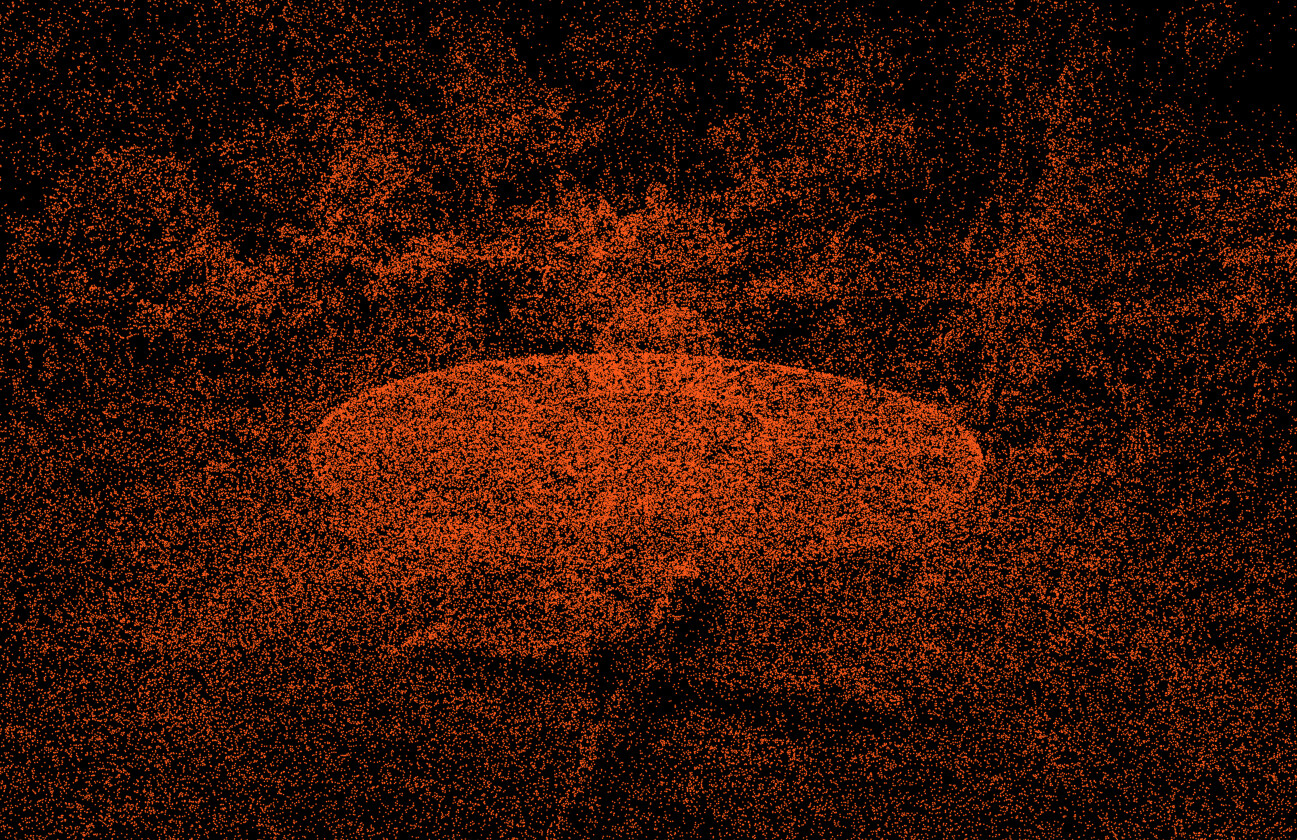}{Ours}
    \overlaylabel{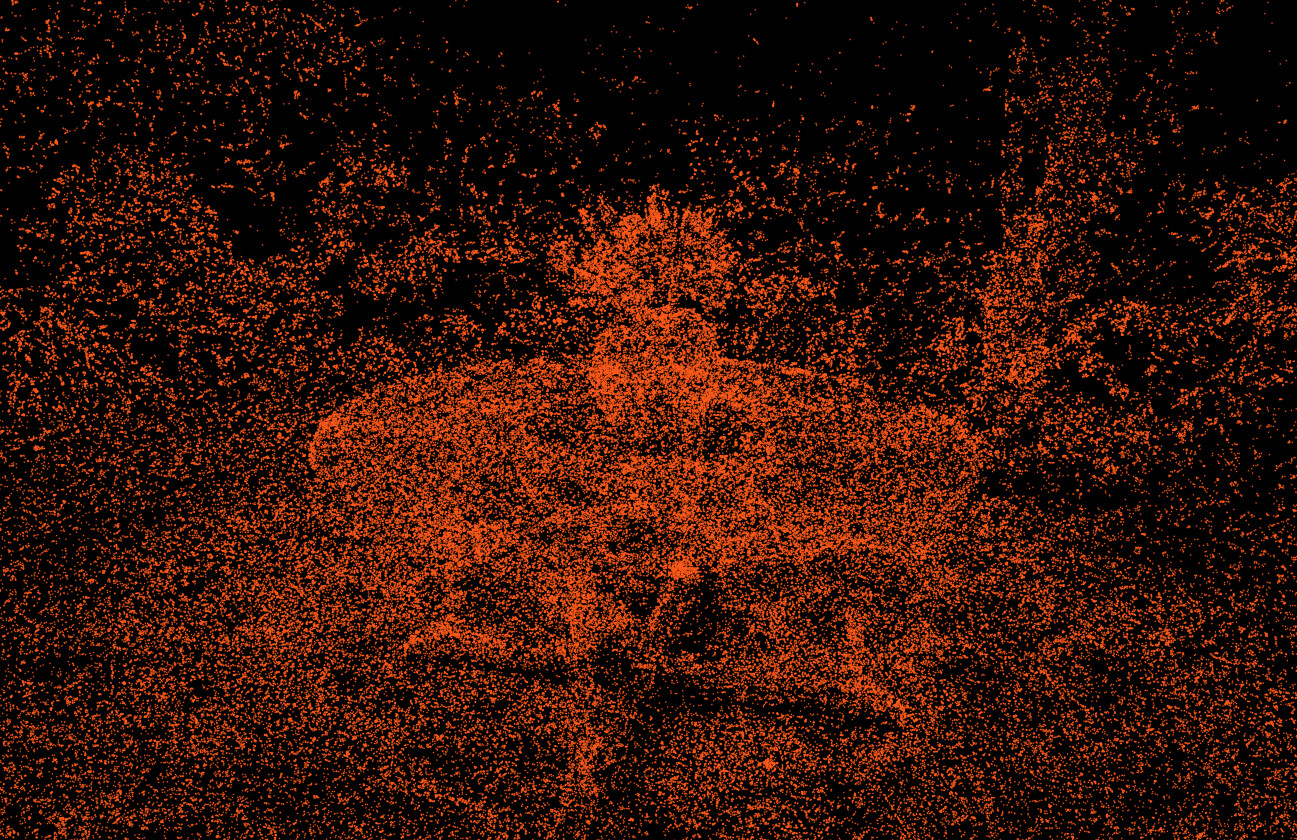}{3DGS-MCMC}
    \caption{
\textbf{Biased sampling in 3DGS-MCMC.}
We render the positions of all source primitives drawn during densification on the Mip-NeRF 360 Garden scene as a point cloud, for our method (left) and 3DGS-MCMC (right).
3DGS-MCMC draws its sources in clusters around the central object and leaves large parts of the background with few sources, while ours cover the background evenly.
    }
    \label{fig:sampling_bias}
\end{figure}
}

\paragraph{Densification.}
Starting from a sparse SfM point cloud, 3DGS relies on densification, \ie adding and removing primitives during optimization.
The ADC of Kerbl~\emph{et al.}~\cite{kerbl20233dgs} splits or clones primitives with large image-space gradients and prunes primitives with low opacity.
Later works refine this error signal with absolute screen-space gradients~\cite{ye2024absgs, yu2024gof} and per-primitive SSIM errors~\cite{bulo2024revising}, which we reuse together with the confidence that CoMe~\cite{radl2026come} learns for mesh extraction.
Taming 3DGS~\cite{mallick2024taming} splits and clones by tailored scores until it reaches a budget.
Simplification methods~\cite{Fang2024minisplatting2, zhang2025gaussianspa} first optimize a large set of primitives and then prune it step by step to a compact, better distributed one.
Prior-based methods start from a denser initialization, obtained from a pre-trained radiance field~\cite{niemeyer2025radsplat, baranowski2026conegs} or from dense correspondences~\cite{Kotovenko2026edgs}, and ConeGS~\cite{baranowski2026conegs} also inserts primitives at high-error pixels, at the depth of its radiance field.
This head start requires a separate initialization stage, which for ConeGS takes nearly as long as our entire training.
3DGS-MCMC~\cite{kheradmand2024mcmc}, the prior-free baseline we build on, grows towards a fixed budget, samples, relocates and perturbs primitives by their opacity, and works even from random initialization.
We keep its budget, growth and relocation but change which primitives are drawn and relocated and where new ones go.

\section{Preliminaries and Analysis}
\label{sec:prelim}

\paragraph{3D Gaussian Splatting.}
\label{sec:prelim:3dgs}
3DGS~\cite{kerbl20233dgs} models a 3D scene as a set of ellipsoids with a Gaussian falloff.
Each primitive $i$ has a center $\bm{\mu}_i$, rotation $\bm{R}_i$, scale $\bm{s}_i$, opacity $o_i$ and a view-dependent color $\bm{c}_i$, represented by spherical harmonics (SH).
To render a view $k$ of the training set $\mathcal{K}$, the primitives are splatted~\cite{zwicker2002ewa} onto the image plane and alpha-blended.
With the $N$ primitives that overlap pixel $\bm{p}$ sorted front to back along its ray, the color of $\bm{p}$ is
\begin{equation}
    \bm{C}(\bm{p}) = \sum_{i=0}^{N-1} \alpha_i(\bm{p})\, \bm{c}_i \prod_{j=0}^{i-1} (1 - \alpha_j(\bm{p})).
\end{equation}
Here, $\alpha_i(\bm{p}) = G_{2D}^i(\bm{p})\, o_i$ is the alpha of the projected Gaussian $G_{2D}^i$, and the product is the transmittance $T_{i-1}(\bm{p})$ in front of primitive $i$, with $T_i = \prod_{j=0}^{i} (1 - \alpha_j)$ and $T_{-1} = 1$.
For primitive $i$ in training view $k$, $b_{i}^{k}(\bm{p}) = \alpha_i(\bm{p})\, T_{i-1}(\bm{p})$ is its blending weight at pixel $\bm{p}$.
The primitives are optimized against the ground-truth image $\bm{C}_\text{GT}$ with the photometric loss, where $\lambda_\text{SSIM} = 0.2$:
\begin{equation}
    \mathcal{L}_\text{rgb} = (1 - \lambda_\text{SSIM}) \mathcal{L}_1 (\bm{C}, \bm{C}_\text{GT}) + \lambda_\text{SSIM} (1 - \mathrm{SSIM}(\bm{C}, \bm{C}_\text{GT})). \label{eq:photoloss}
\end{equation}

\paragraph{Learned Confidence.}
CoMe~\cite{radl2026come} adds a self-supervised confidence~\cite{wang2024duster} to 3DGS.
Each primitive carries a confidence $\bar{\gamma}_i = \exp(\gamma_i)$ with learnable $\gamma_i$, and alpha-blending these confidences gives the confidence map $\hat{C}(\bm{p}) \in \mathbb{R}_+$.
It is trained per pixel with
\begin{equation}
    \mathcal{L}_\mathrm{conf}(\bm{p}) = \mathcal{L}_\mathrm{rgb}(\bm{p}) \cdot \hat{C}(\bm{p}) - \beta_\text{conf} \log \hat{C}(\bm{p}),\label{eq:confidence}
\end{equation}
where $\mathcal{L}_\mathrm{rgb}(\bm{p})$ is the photometric error of Equation~\eqref{eq:photoloss} at pixel $\bm{p}$ and $\beta_\text{conf}$ is a hyperparameter.
This loss is minimal at $\hat{C}(\bm{p}) = \beta_\text{conf} / \mathcal{L}_\mathrm{rgb}(\bm{p})$, so $\beta_\text{conf}$ is the error level at which the confidence equals one, and the confidence map falls below 1 wherever the photometric error exceeds it.
Since $\hat{C}$ is blended from the per-primitive confidences, primitives that cover such pixels learn $\bar{\gamma}_i < 1$, which marks under-reconstructed regions, while $\bar{\gamma}_i > 1$ marks well-reconstructed ones.
We use these confidences as one of the error signals that decide where new primitives go (Section~\ref{sec:method:sampling}).

\paragraph{Fixed-Budget Densification.}
\label{sec:mcmc}
3DGS-MCMC~\cite{kheradmand2024mcmc} replaces the splitting and cloning of earlier methods~\cite{kerbl20233dgs, bulo2024revising} with relocation within a fixed budget of primitives.
Every 100 iterations, it draws source primitives by opacity-proportional sampling, places new primitives at their positions and grows the number of primitives by 5\% until the budget is reached.
It adjusts the opacity and scale of each source and of the primitives placed at it so that the rendered image stays approximately unchanged, which lets training continue smoothly across densification steps.
A primitive whose opacity $o_i$ drops to $0.005$ or below is moved to such a source (opacity-threshold relocation).
Opacities decrease mainly due to the regularizer
\begin{align}
    \mathcal{L}_\text{reg} = \lambda_\text{reg} \sum_i \left( |o_i| + \|\bm{s}_i\|_1\right),\label{eq:reg}
\end{align}
with weight $\lambda_\text{reg}$.
3DGS-MCMC also adds continuous position noise at every iteration, shaped by the covariance of each primitive and gated so that mainly low-opacity primitives move.

\begin{figure}[t!]
    \centering
    \includegraphics[width=\linewidth]{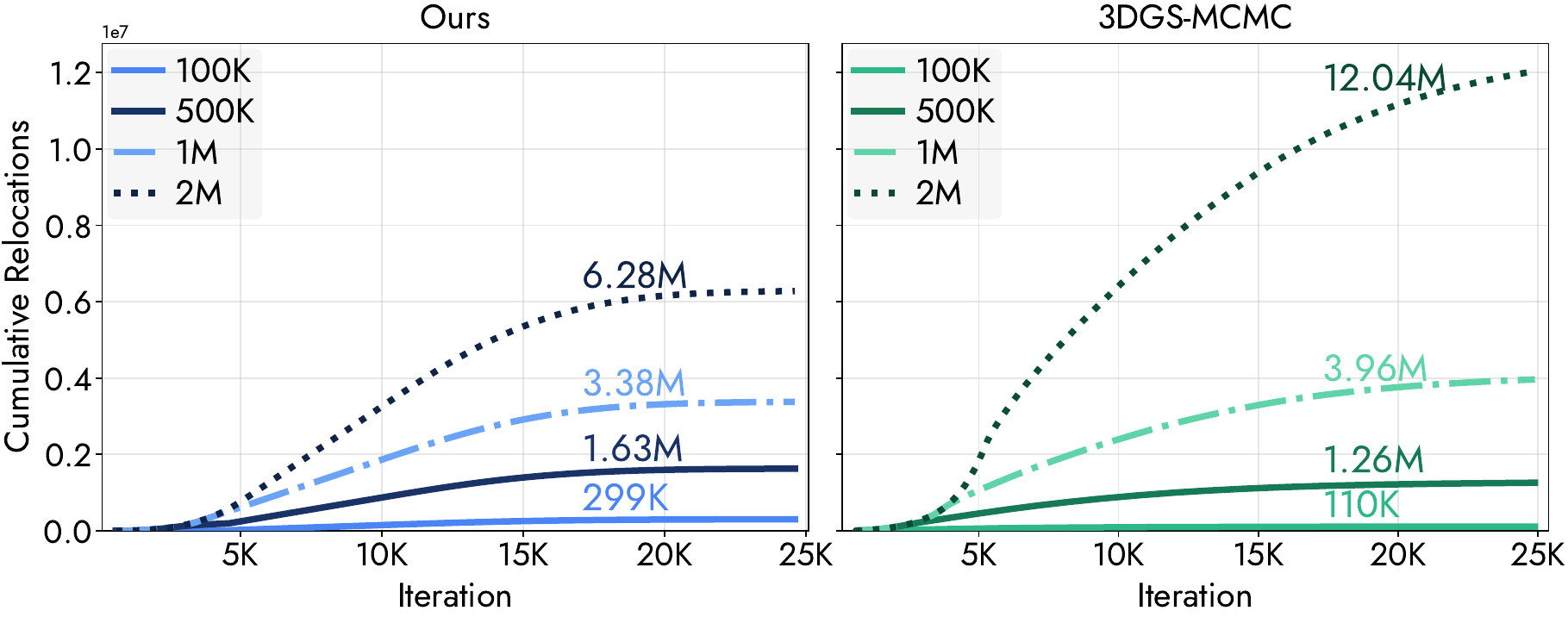}
    \caption{
    \textbf{Relocation Schedule}:
Cumulative number of relocated primitives during training at each budget, for our relocation every $400$ iterations (left) and for the opacity-threshold relocation of 3DGS-MCMC (right), measured for the \emph{Bonsai} scene~\cite{barron2022mipnerf360}.
We relocate more primitives than 3DGS-MCMC at $\mathrm{100k}$ and $\mathrm{500k}$, but fewer at $\mathrm{1M}$ and $\mathrm{2M}$.
    }
    \label{fig:budget}
\end{figure}

\paragraph{Foreground Bias of 3DGS-MCMC.}
\label{sec:bias}
Both opacity rules concentrate primitives on the well-reconstructed foreground, starting while the budget fills.
The initialization already places many primitives in the foreground, and because these regions are seen in many training views, the photometric loss keeps their opacities high against $\mathcal{L}_\text{reg}$.
While the budget fills, opacity-proportional sampling therefore draws most source primitives there, and each growth step adds primitives where many already exist (Figure~\ref{fig:sampling_bias}).
Once the budget is reached, only primitives whose opacity has decayed are moved, and they go to opacity-sampled sources again.
Relative to the budget, 3DGS-MCMC relocates far fewer primitives at $\mathrm{100k}$ than at $\mathrm{2M}$ (Figure~\ref{fig:budget}), so the distribution formed during growth is corrected least where primitives are scarcest, which is also where prior-based methods lead 3DGS-MCMC most.
Opacity-proportional sampling misses the high-error regions (Figure~\ref{fig:distribution}), and in the trained models, 3DGS-MCMC blends many primitives into foreground pixels and few into background pixels, at $\mathrm{100k}$ and $\mathrm{500k}$ (Figure~\ref{fig:num_blended}).
We measure the cost of each rule in Section~\ref{sec:analysis}.

{
\newcommand{\overlaylabel}[3]{%
    \begin{tikzpicture}
        \node[anchor=south west, inner sep=0] (image) at (0,0) {%
            \includegraphics[width=0.49\linewidth]{#1}%
        };
        \begin{scope}[x={(image.south east)},y={(image.north west)}]
            \node[anchor=south east, 
                  fill=black, fill opacity=0.45, 
                  inner sep=2.5pt, 
                  text opacity=0.8,
                  rounded corners=1.5pt,
                  align=right] 
            at (0.98, 0.02) {%
                {\sffamily\scriptsize\textbf{\textcolor{white}{#2}}}\\[-6pt]
                {\sffamily\fontsize{5.5pt}{6.5pt}\selectfont\color{white!70} #3}%
            };
        \end{scope}
    \end{tikzpicture}%
}
\begin{figure}[t]
    \centering
    {\sffamily\scriptsize\textbf{@ 100k Primitives}}\\[1pt]
    \overlaylabel{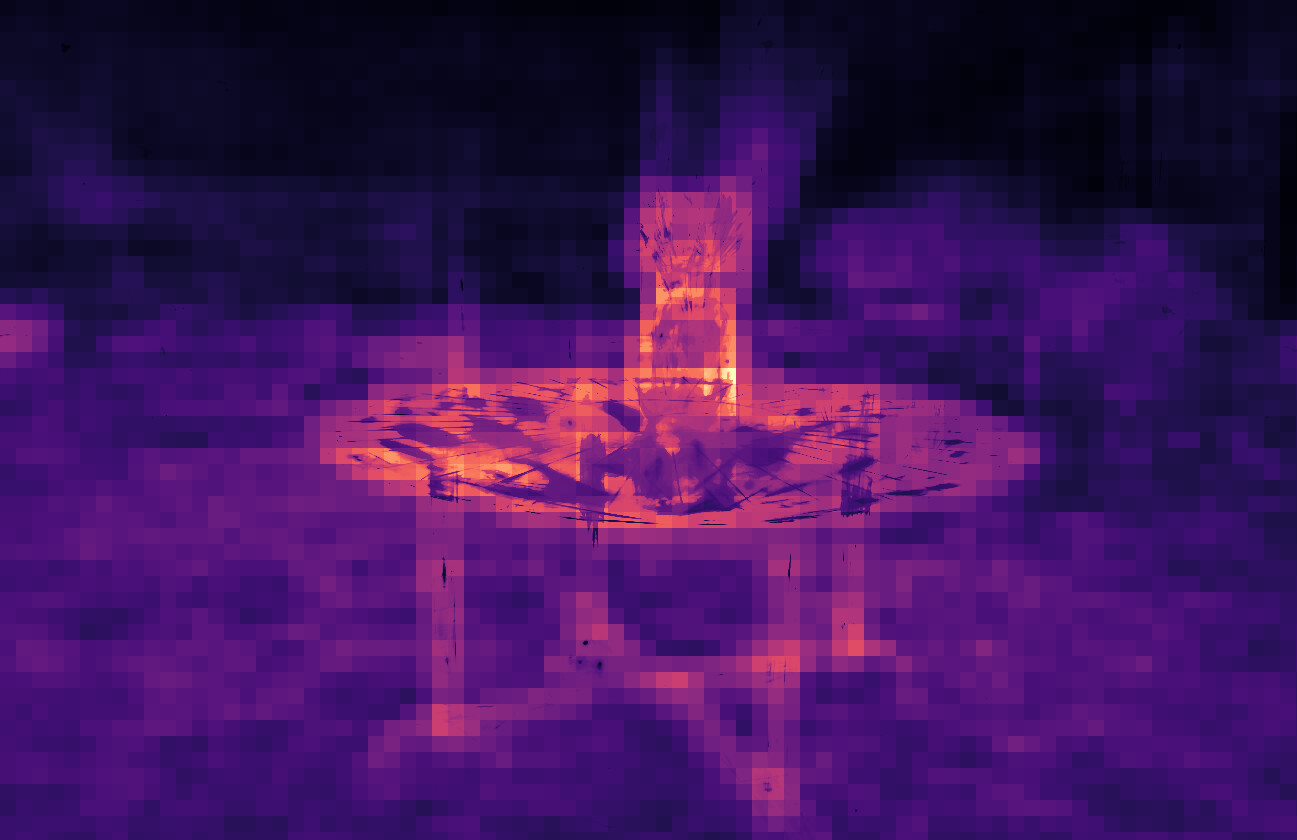}{3DGS-MCMC}{24.37dB}\hfill
    \overlaylabel{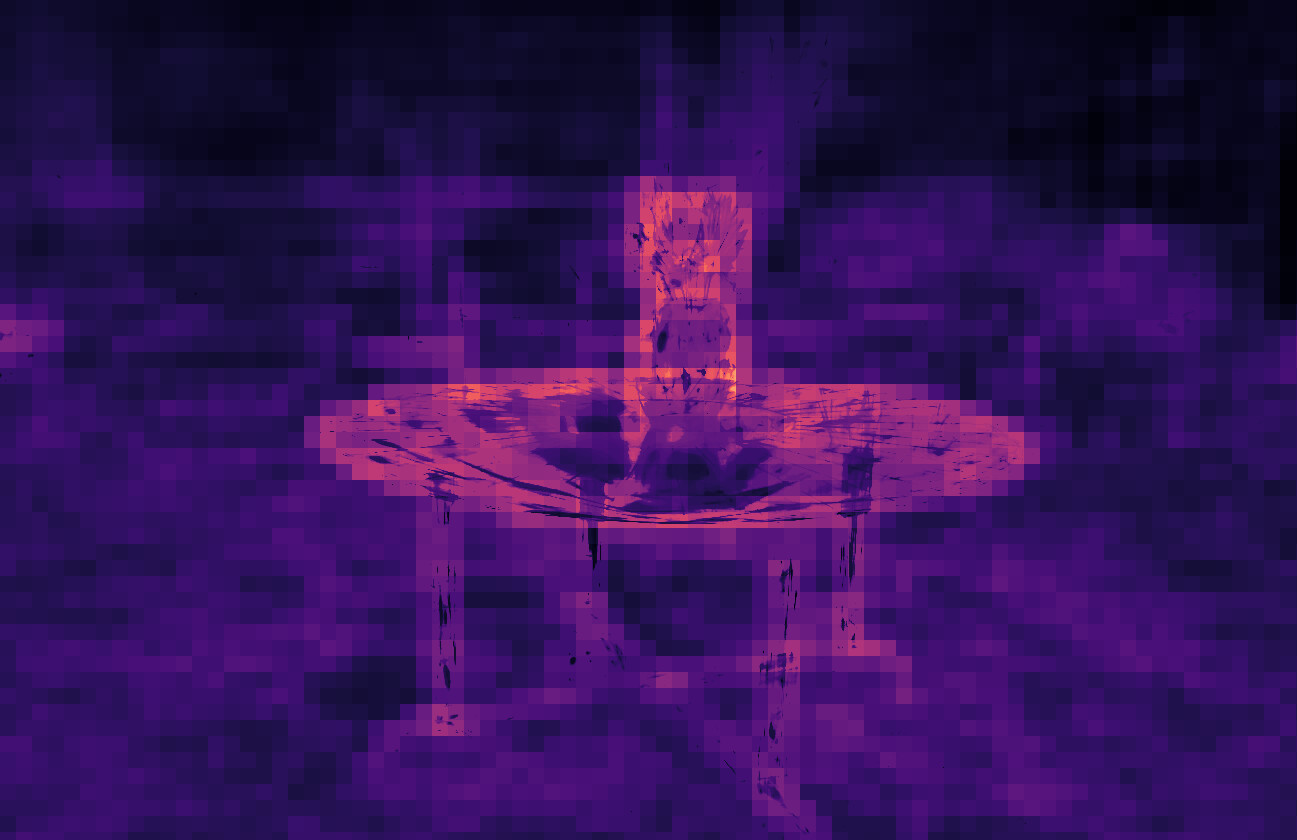}{Ours}{24.89dB}
    
    \vspace{-1pt} %

    {\sffamily\scriptsize\textbf{@ 500k Primitives}}\\[1pt]
    \overlaylabel{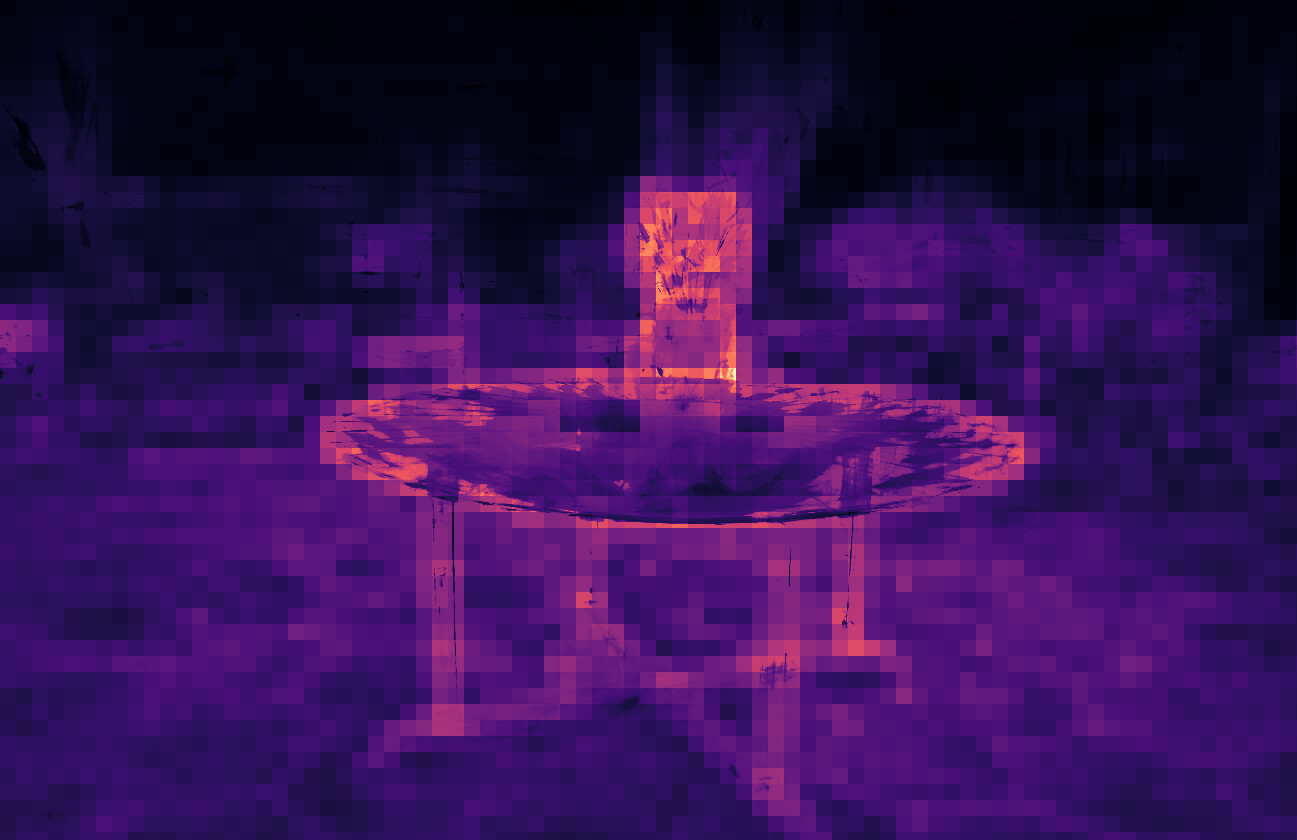}{3DGS-MCMC}{26.39dB}\hfill
    \overlaylabel{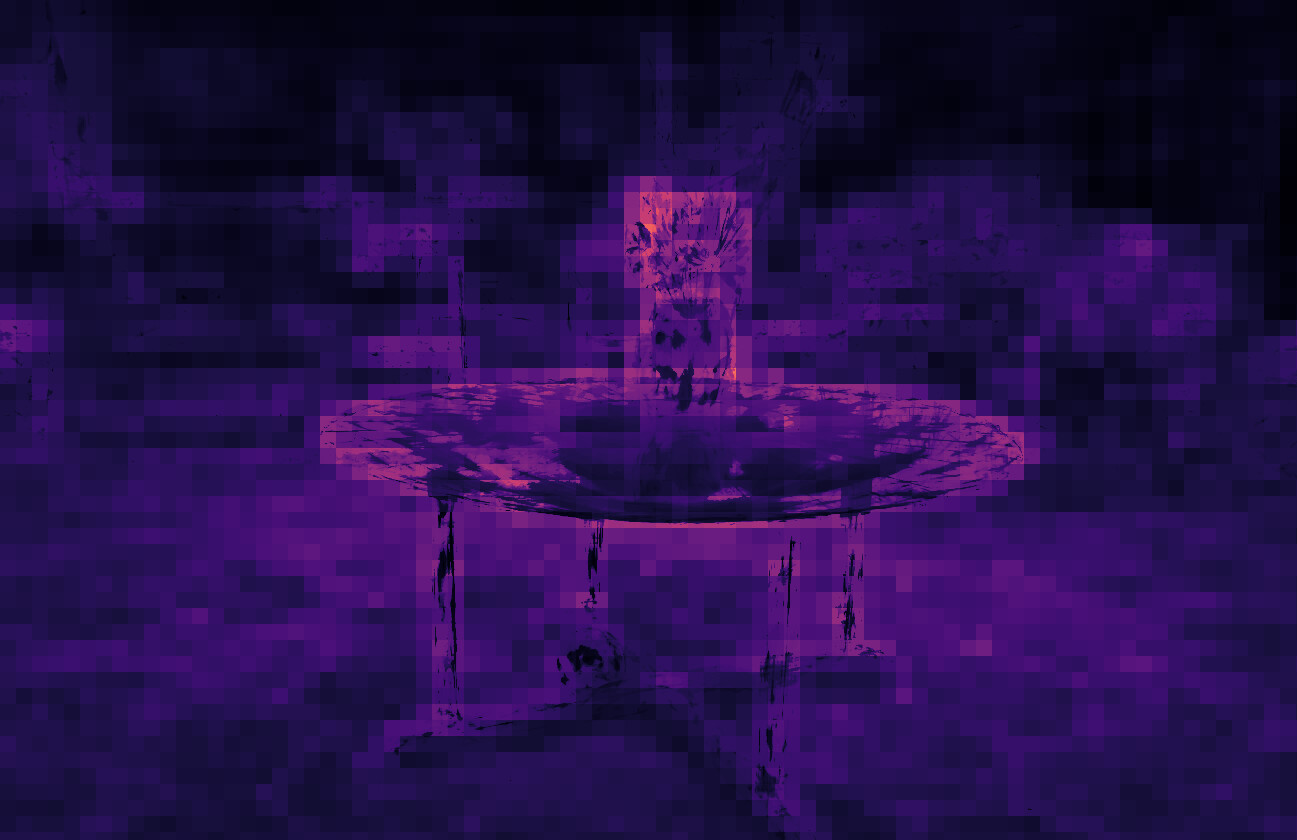}{Ours}{27.13dB}

    \caption{\textbf{3DGS-MCMC Foreground Bias}:
Number of primitives blended into each pixel for the Mip-NeRF 360 Garden scene (brighter means more) after training with budgets of $\mathrm{100k}$ (top) and $\mathrm{500k}$ (bottom).
In 3DGS-MCMC, most primitives end up on the foreground objects; ours are spread more evenly over foreground and background.
The labels give the mean PSNR of each model over the test views of this scene.}
    \label{fig:num_blended}
\end{figure}
}

{
\newcommand{\overlabel}[2]{%
    \begin{tikzpicture}
        \node[anchor=south west, inner sep=0] (image) at (0,0) {%
            \includegraphics[width=0.24\linewidth]{#1}%
        };
        \begin{scope}[x={(image.south east)},y={(image.north west)}]
            \node[anchor=south east, 
                  fill=black, fill opacity=0.576, 
                  inner sep=1.5pt, 
                  text opacity=0.8, 
                  rounded corners=0.7pt, 
                  align=right] 
            at (0.98, 0.02) {%
                {\sffamily\fontsize{4.2pt}{5.2pt}\selectfont{\textcolor{white}{#2}}}%
            };
        \end{scope}
    \end{tikzpicture}%
}
\begin{figure}
    \centering
    \overlabel{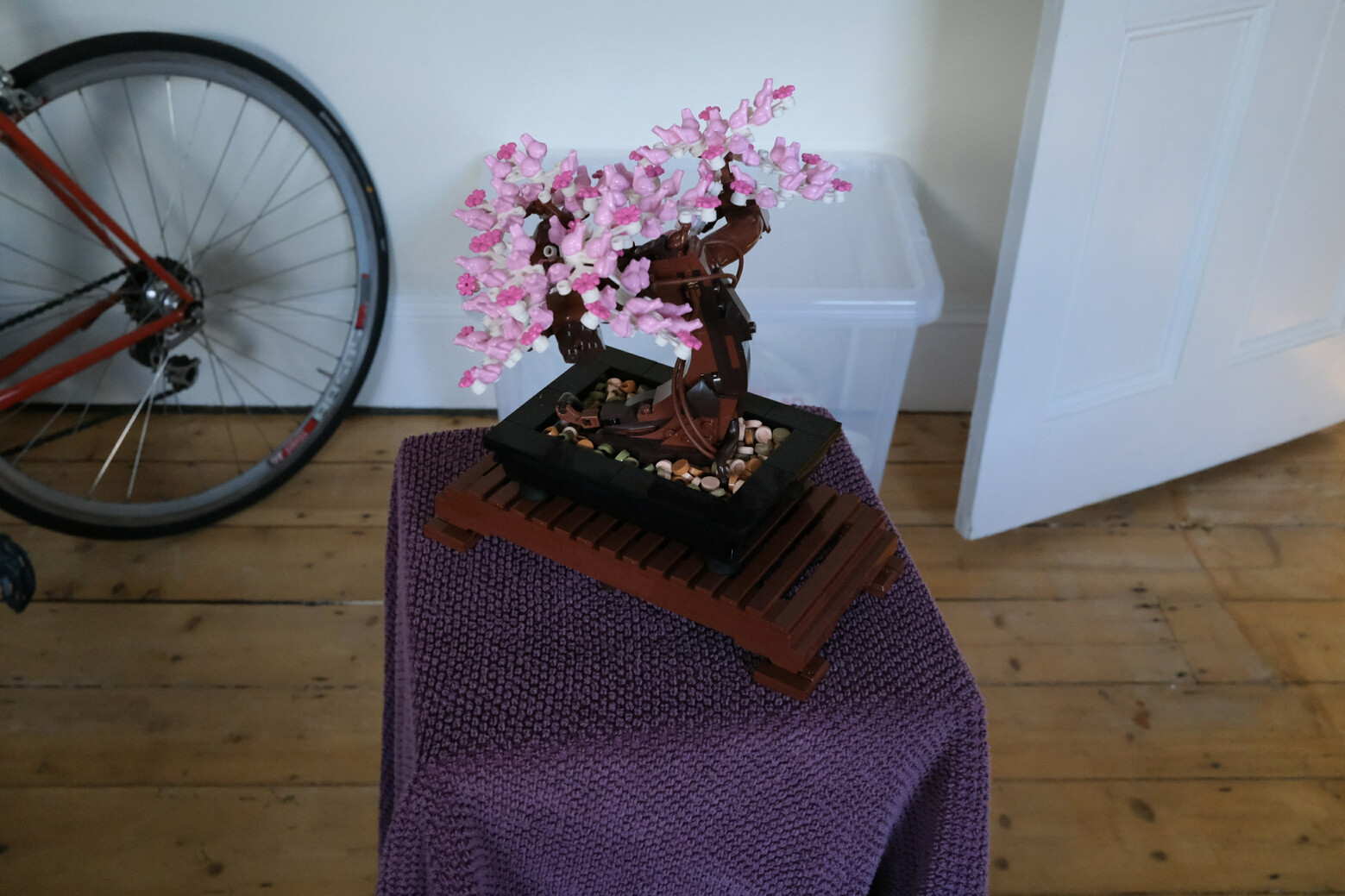}{GT}
    \overlabel{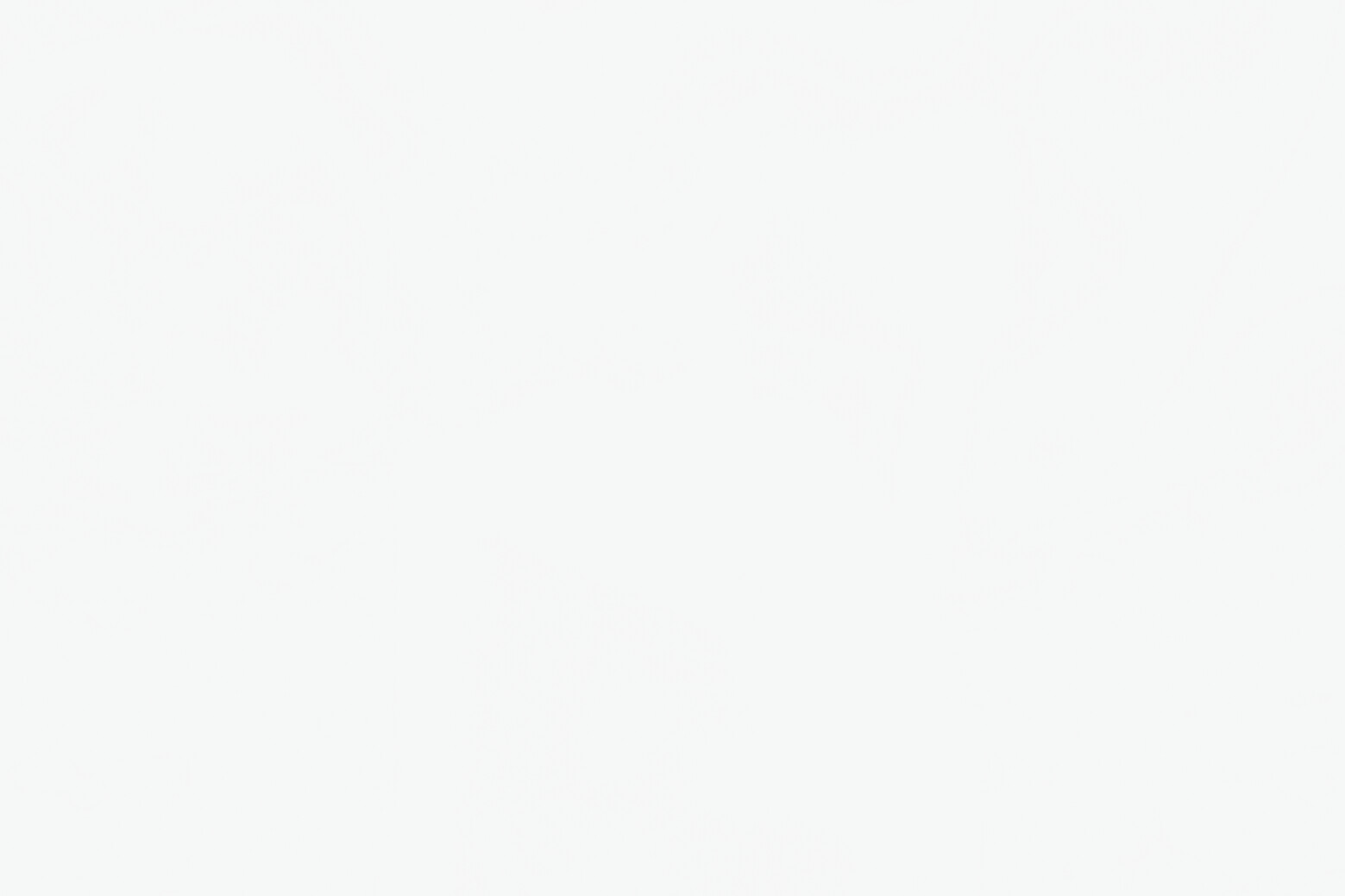}{iter 500}
    \overlabel{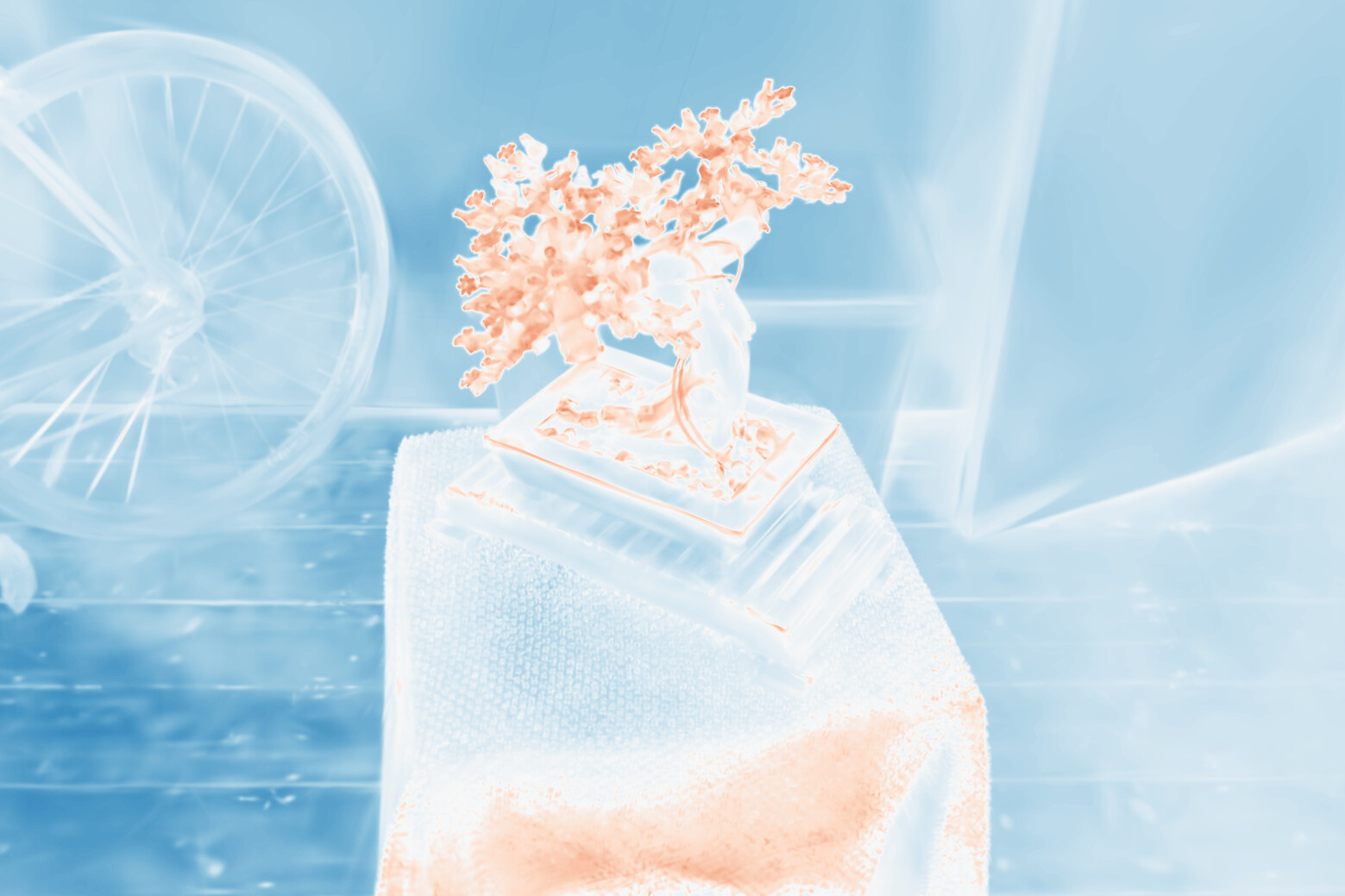}{iter 5k}
    \overlabel{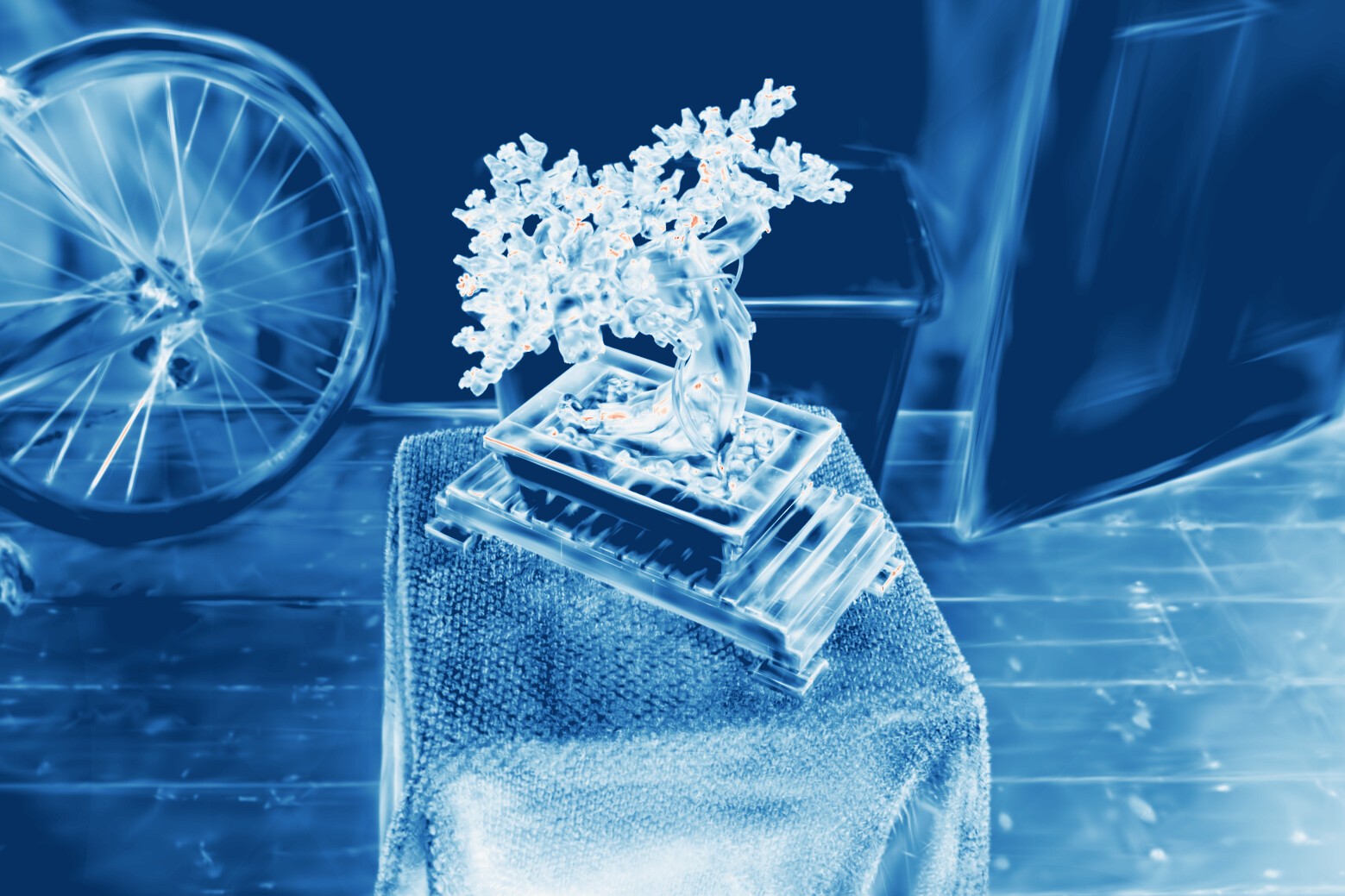}{iter 30k} \\
    \overlabel{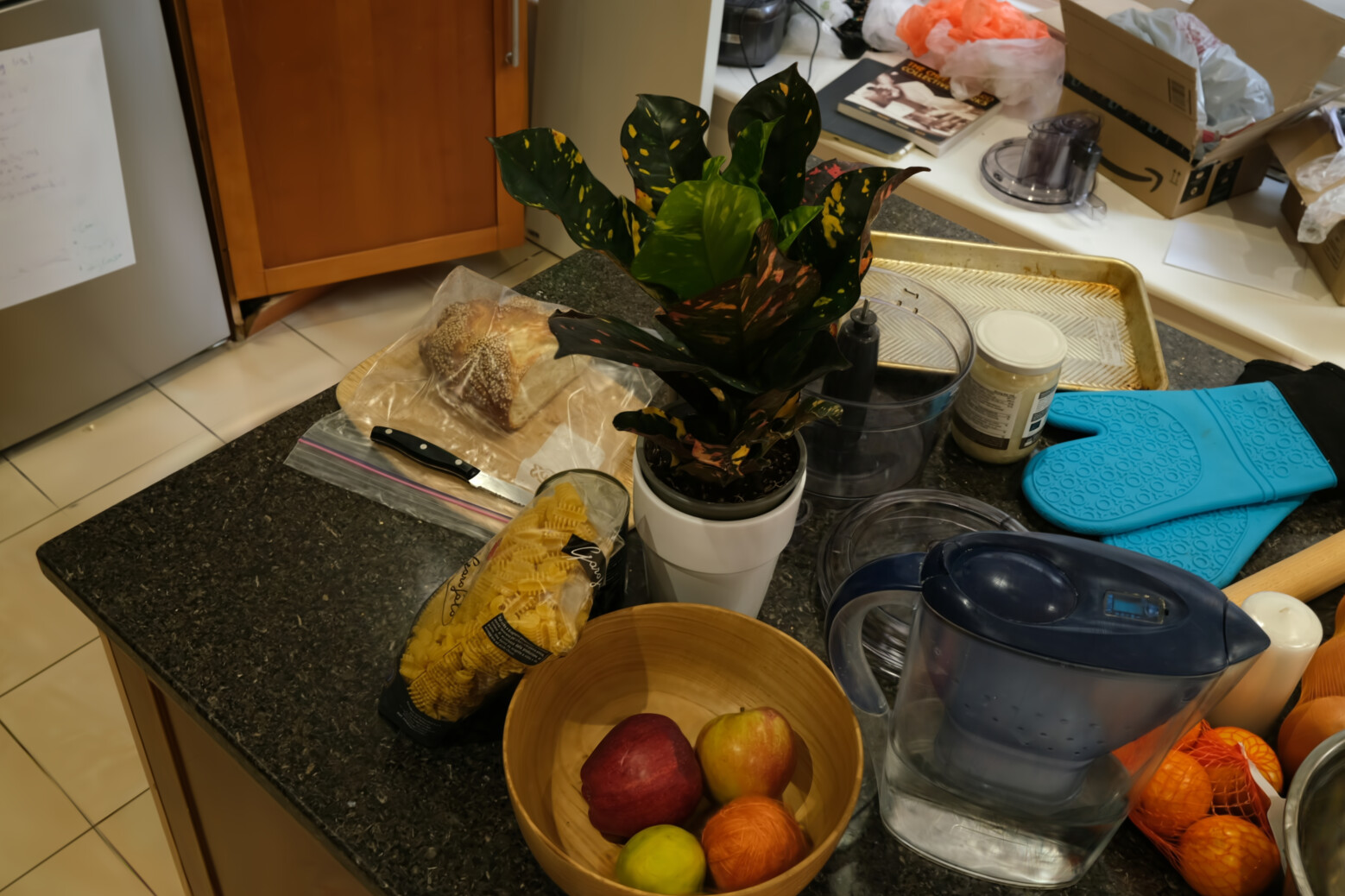}{GT}
    \overlabel{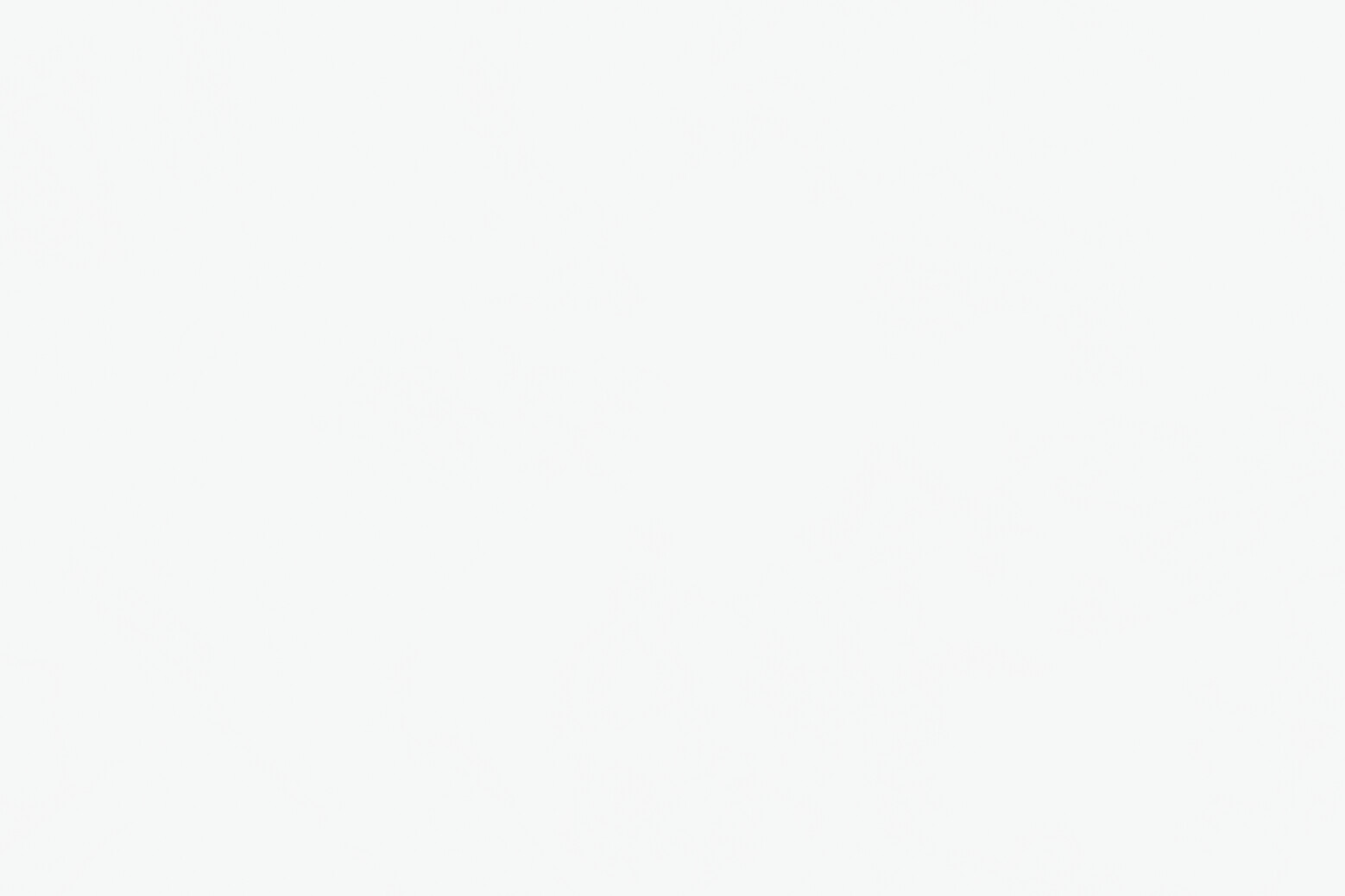}{iter 500}
    \overlabel{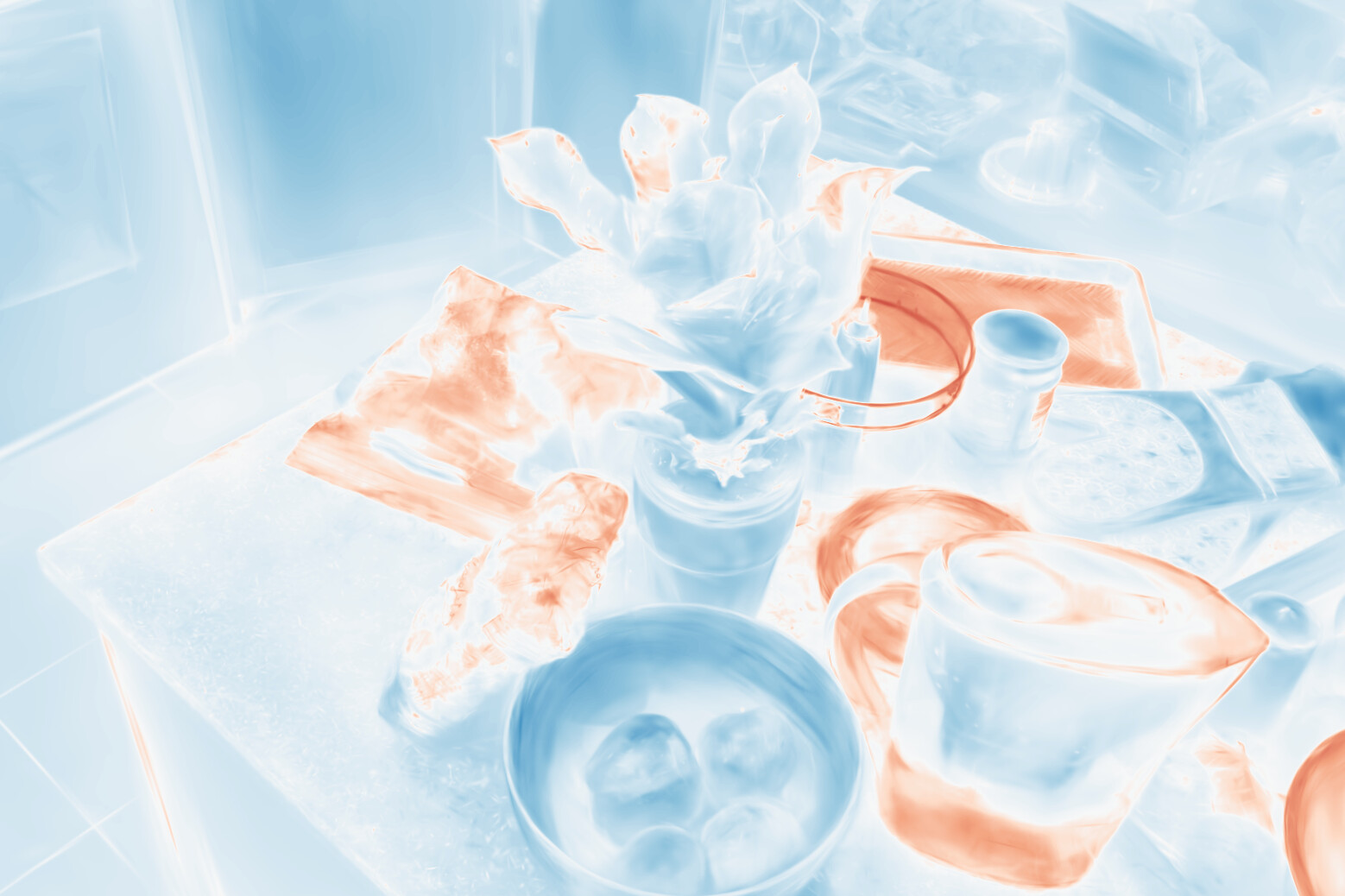}{iter 5k}
    \overlabel{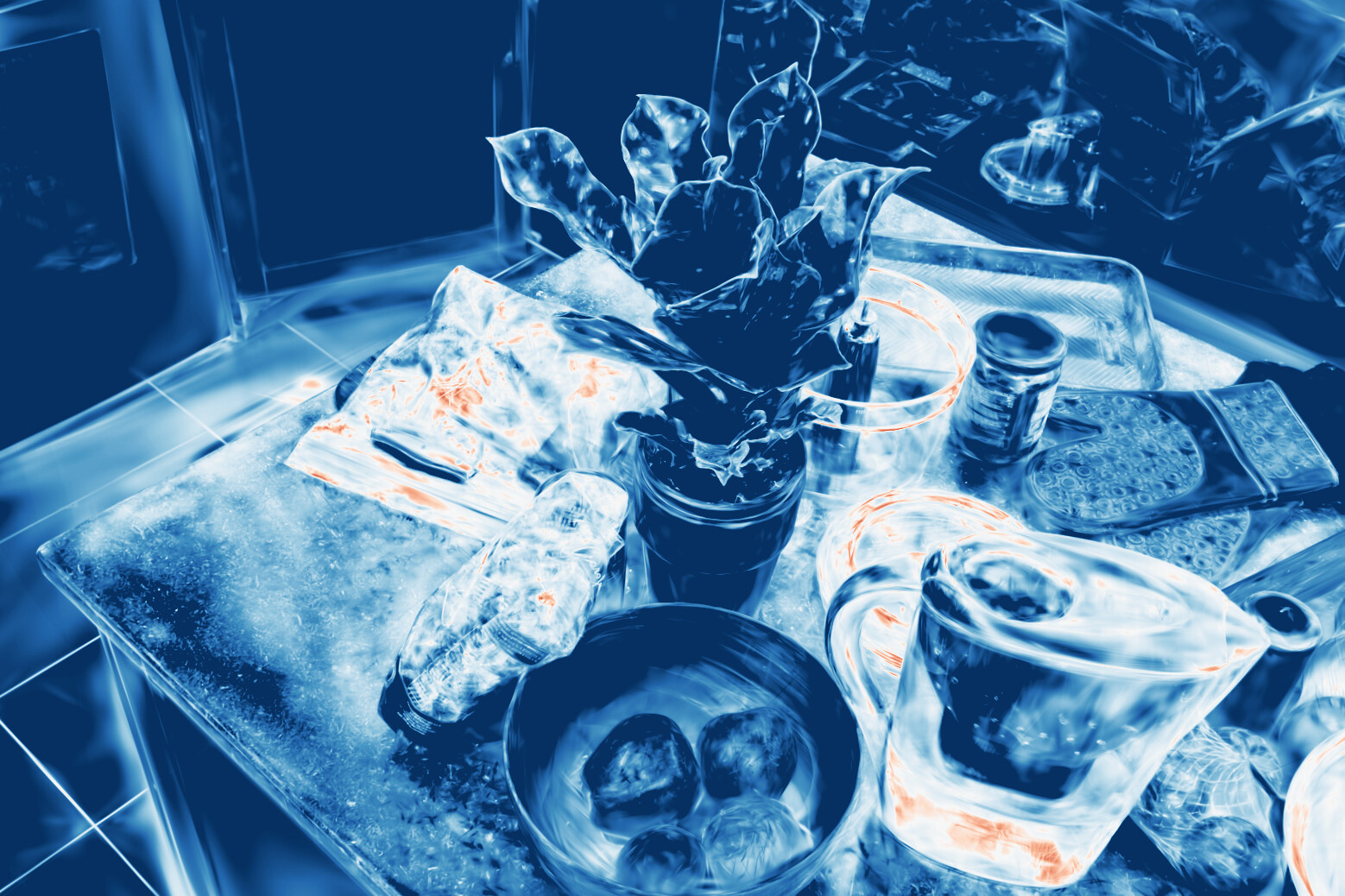}{iter 30k} \\
    \includegraphics[width=0.65\linewidth]{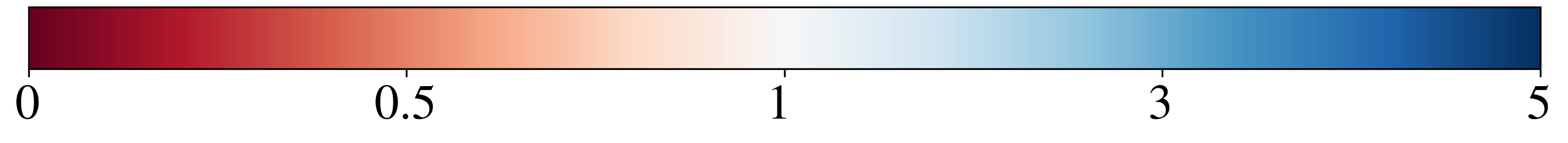}
    \caption{
\textbf{Convergence of Confidence}:
Rendered confidence maps~\cite{radl2026come} of Bonsai (top) and Counter (bottom) at iterations 500, 5k and 30k, next to the ground truth (GT); red marks confidences below 1 and blue above. At iteration 500, the maps are nearly white because all confidences start at 1.
We therefore start by sampling by absolute gradients alone and blend in the confidence over the first 5000 iterations (Section~\ref{sec:method:sampling}).
    }
    \label{fig:confidence}
\end{figure}
}

\section{Confidence-Aware Densification}
\label{sec:method}
We keep the fixed-budget framework of 3DGS-MCMC: 
within the 30k training iterations, densification steps every 100 iterations from iteration 500 to 25k, growth by 5\% per step until the budget is reached, and relocation alongside growth.
What we change is where this framework places primitives:
At each step, we draw source primitives from a distribution that follows the image error instead of the opacity (Section~\ref{sec:method:sampling}).
The same draw serves growth and relocation, so this distribution alone decides where the model densifies.
We relocate the primitives with the lowest maximum blending weight in the training views instead of those below an opacity threshold (Section~\ref{sec:method:population}).
Each new primitive receives a position drawn from the Gaussian of its source, which replaces the continuous position noise of 3DGS-MCMC, and a color taken from a training image (Section~\ref{sec:method:initialization}).
Compared with 3DGS-MCMC, we add one confidence per primitive and the loss of Equation~\eqref{eq:confidence}, need no initialization stage, and use one set of hyperparameters for all scenes and budgets.
Growth reaches the budget early in training, and relocation moves up to 10\% of the primitives per step until iteration 10k, so our sampling acts while the distribution of primitives forms.

\subsection{Sampling-Based Densification}
\label{sec:method:sampling}
The learned confidence is one error signal for this distribution, since $\bar{\gamma}_i < 1$ marks primitives in under-reconstructed regions~\cite{radl2026come}, but sampling by confidence alone has a drawback.
All confidences start at $\bar{\gamma}_i = 1$, so the confidence map is uniform at first and becomes a useful error proxy only after several thousand iterations (Figure~\ref{fig:confidence}).

We therefore combine the confidence with the absolute screen-space gradient $\nabla_\text{abs}$ of GOF~\cite{yu2024gof} and AbsGS~\cite{ye2024absgs}, which is usable from the first iteration on.
For each primitive, $\nabla_\text{abs}$ sums the magnitudes of the per-pixel gradients of its screen-space position, so gradients of opposite sign do not cancel as they do in ADC.
This matters most for large primitives that cover several structures: their per-pixel gradients cancel in the sum, so the signed gradient stays small although such a primitive blurs detail~\cite{ye2024absgs}.
We average it over the observations of the primitive since the previous densification step.
We use the ranks of both factors, so neither their scale nor a few very large gradients can dominate the distribution.
Let $r_i(x) \in [0,1]$ be the rank of $x_i$ among all primitives, normalized so that the smallest value has rank 0 and the largest rank 1.
We draw 80\% of the source primitives with probability proportional to
\begin{equation}
    \mathcal{P}_i = r_i(\nabla_\text{abs}) \left( (1 - \eta) + \eta \left(1.2 - r_i(\bar{\gamma})\right) \right), \label{eq:pdf}
\end{equation}
where $\eta = \min(t/5000, 1)$ at iteration $t$ shifts sampling from $\nabla_\text{abs}$ alone to the combined distribution over the first 5000 iterations.
At $\eta = 1$, the confidence factor $1.2 - r_i(\bar{\gamma})$ ranges from $1.2$ for the least to $0.2$ for the most confident primitive, so well-reconstructed primitives are still drawn, less often.
We visualize our combined distribution, compared to opacity-proportional sampling, in Figure~\ref{fig:distribution};
our error-informed distribution picks under-reconstructed regions for densification, which opacity-based sampling deprioritizes.

{
\newcommand{\overlaylabel}[3]{%
    \begin{tikzpicture}
        \node[anchor=south west, inner sep=0] (image) at (0,0) {%
            \includegraphics[width=0.325\linewidth]{#1}%
        };
\begin{scope}[x={(image.south east)},y={(image.north west)}]
            \node[anchor=north west,
                  fill=black, fill opacity=0.35, 
                  inner sep=2.5pt, 
                  text opacity=0.9,
                  rounded corners=0.1pt,
                  align=left] 
at (0.01, 0.99) {%
                {\sffamily\fontsize{5.5pt}{6.5pt}\selectfont\textbf{\textcolor{white}{#2}}}\\[-7pt]
                {\sffamily\fontsize{4.5pt}{5.5pt}\selectfont\color{white!70} #3}%
            };
        \end{scope}
    \end{tikzpicture}%
}
\newcommand{\overlabel}[2]{%
    \begin{tikzpicture}
        \node[anchor=south west, inner sep=0] (image) at (0,0) {%
            \includegraphics[width=0.49\linewidth]{#1}%
        };
        \begin{scope}[x={(image.south east)},y={(image.north west)}]
            \draw[red, line width=0.6pt, draw opacity=0.9] (0.5, 0.45) rectangle (0.9, 0.9);

            \node[anchor=south east, 
                  fill=black, fill opacity=0.25, 
                  inner sep=2.5pt, 
                  text opacity=0.8, 
                  rounded corners=1.5pt, 
                  align=right] 
            at (0.98, 0.02) {%
                {\sffamily\scriptsize\textbf{\textcolor{white}{#2}}}
            };
        \end{scope}
    \end{tikzpicture}%
}
\begin{figure}[t!]
    \centering
    \overlabel{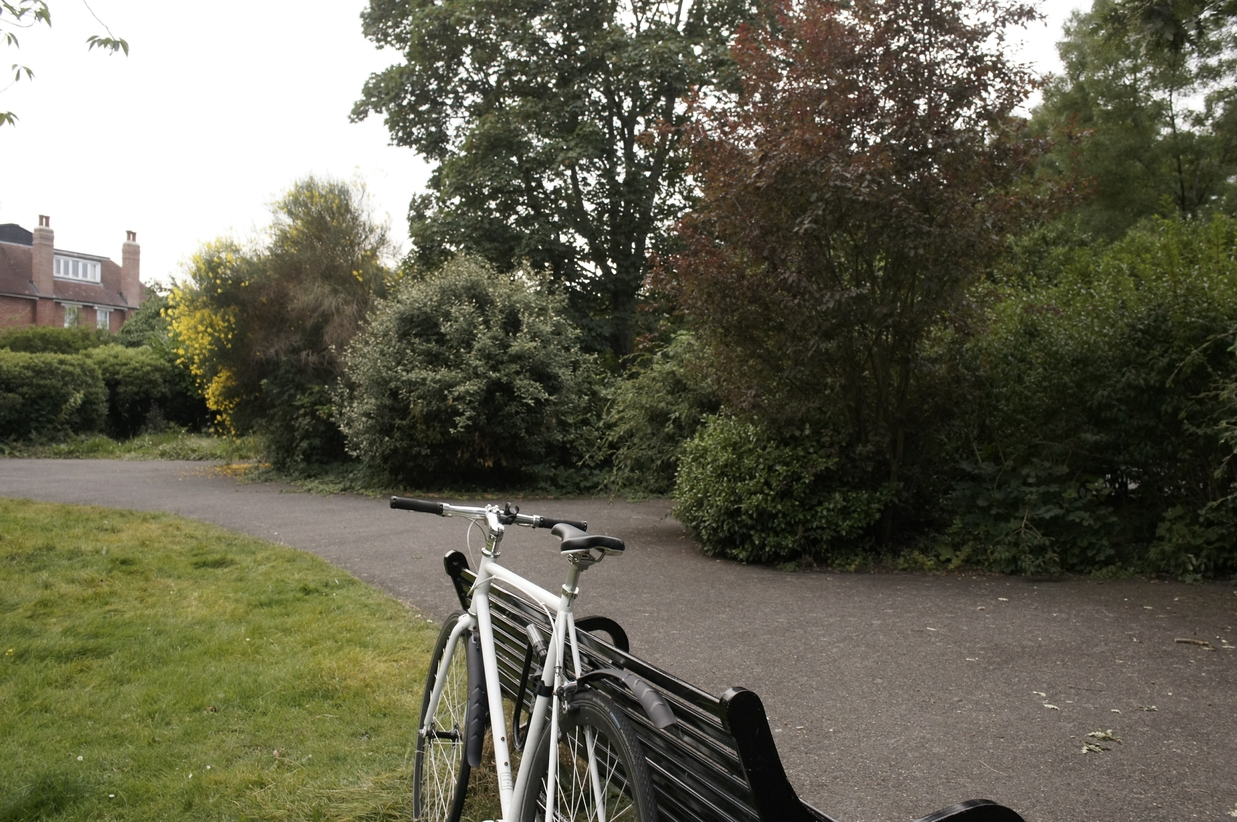}{Ground-Truth}
    \overlabel{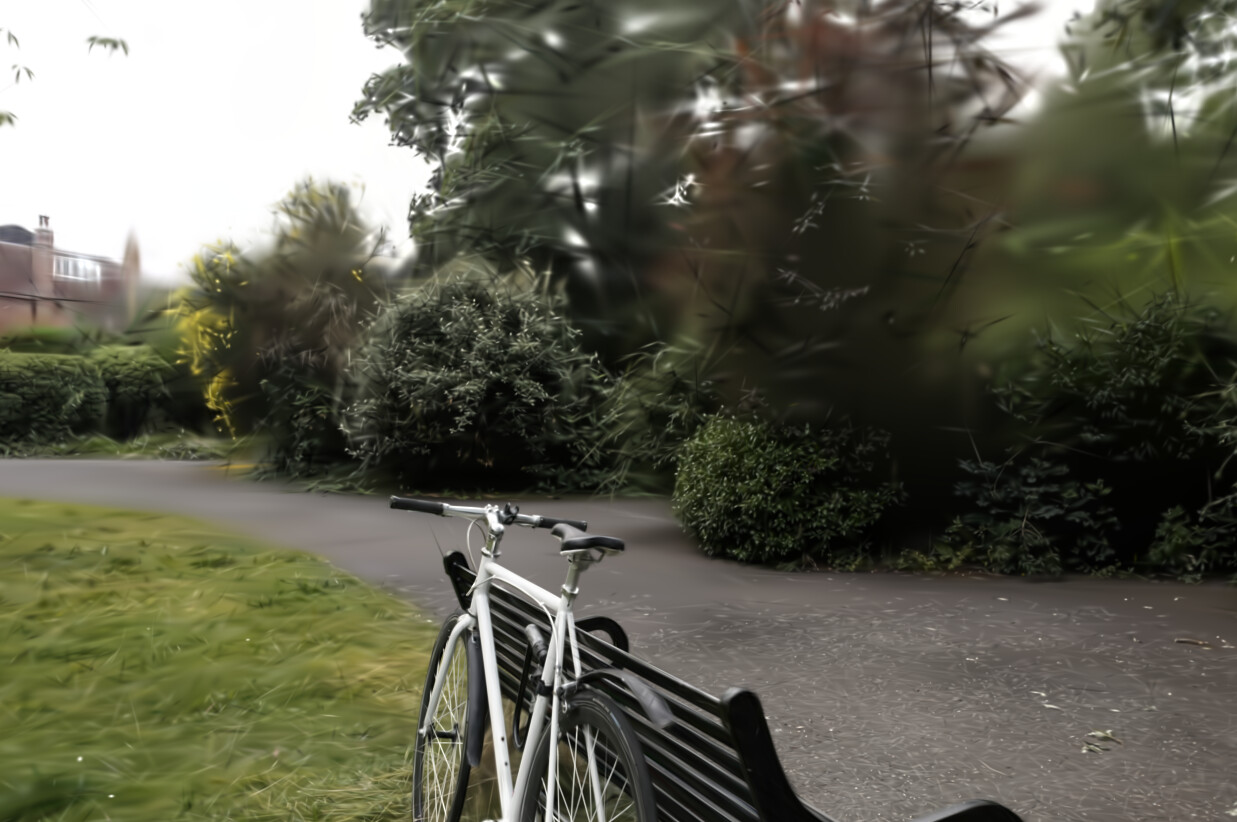}{3DGS Model}
    \hfill
    \overlaylabel{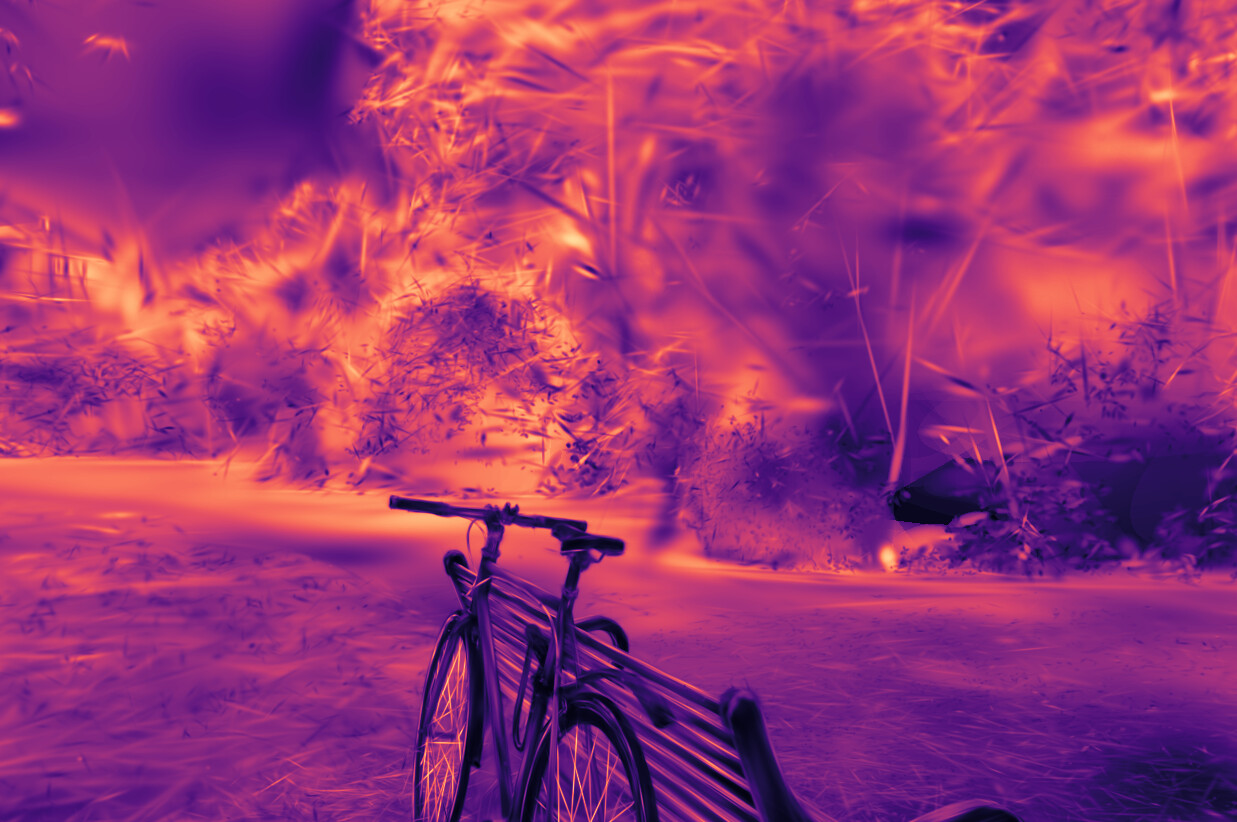}{Ours}{Gradient + Confidence}
    \overlaylabel{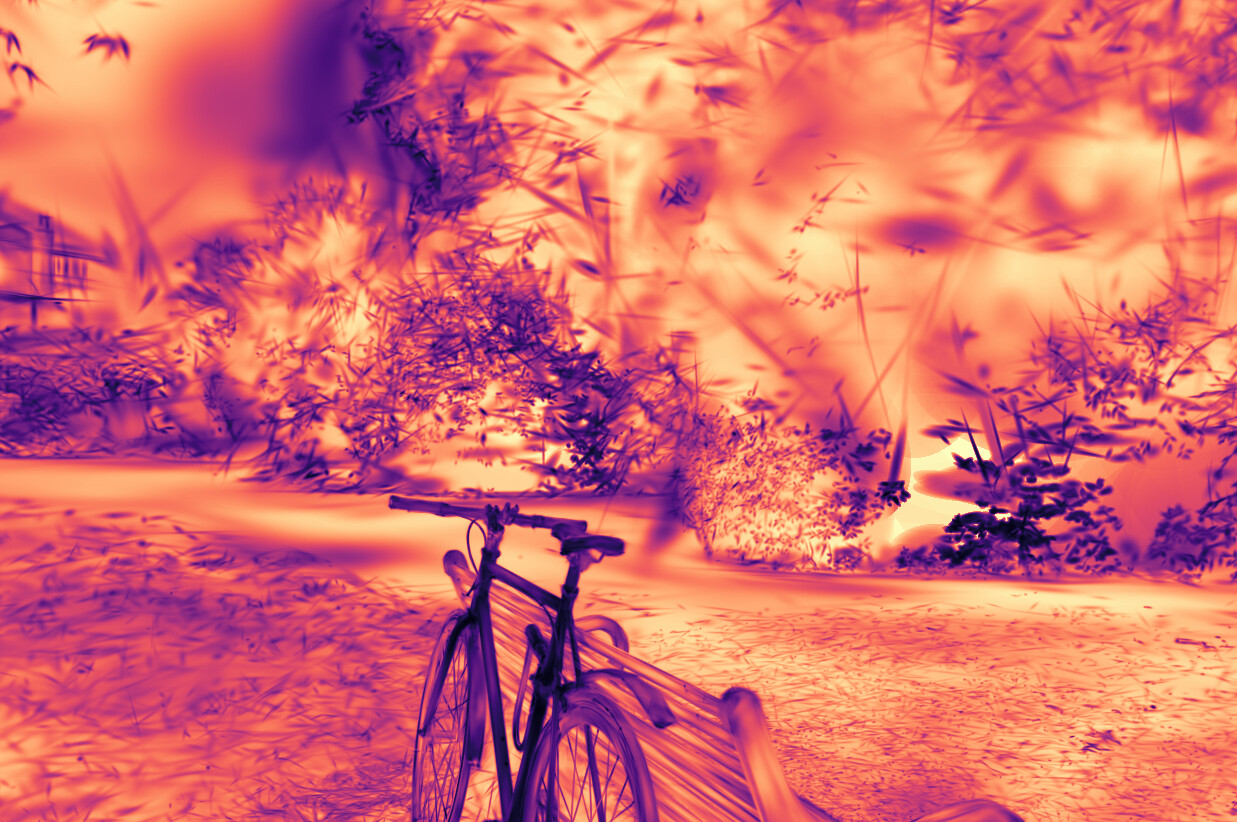}{Ours}{SSIM}
    \overlaylabel{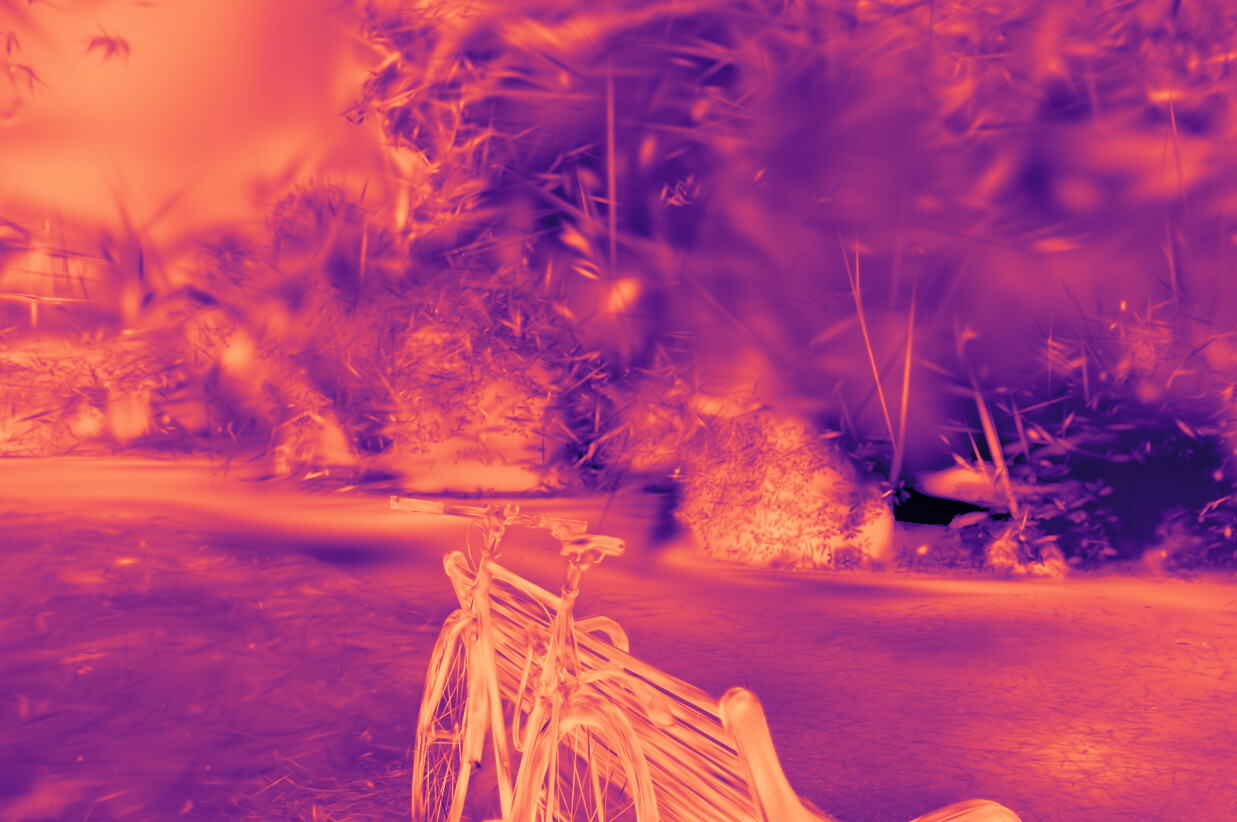}{3DGS-MCMC}{Opacity}
    
    \caption{
\textbf{Sampling Distributions}:
Top: ground truth and a rendering of our trained Bicycle model.
Bottom: our confidence-weighted absolute gradients (Equation~\eqref{eq:pdf}), the SSIM term and the opacity-proportional sampling of 3DGS-MCMC in image space (brighter is more likely).
Opacity-proportional sampling misses the high-error regions in the red box, which both our distributions favor.
    }
    \label{fig:distribution}
\end{figure}
}

We draw the remaining 20\% of the source primitives from a per-primitive structural error, which we call the SSIM term.
Sampling by Equation~\eqref{eq:pdf} alone occasionally misses regions with low $\mathcal{L}_1$ error but visible structural error, because the photometric loss in Equation~\eqref{eq:photoloss} weights $\mathcal{L}_1$ four times more strongly than its D-SSIM part.
Inspired by Rota Bul\`o~\emph{et al.}~\cite{bulo2024revising}, we compute the per-pixel structural error $1-\mathrm{SSIM}$, which we distribute to the contributing primitives proportional to their blending weight.
We do not weight the SSIM term by confidence, so it also reaches regions that the confidence factor in Equation~\eqref{eq:pdf} down-weights.

\subsection{Relocation by Maximum Blending Weight}
\label{sec:method:population}

3DGS-MCMC relocates the primitives whose opacity has fallen to $0.005$ or below, at every densification step (Section~\ref{sec:mcmc}).
We relocate only at every fourth step, \ie, every 400 iterations, so that the blending weights can adapt to the new primitives in between, and we select the primitives to relocate by their maximum blending weight
\begin{equation}
    w_i := \max_{k \in \mathcal{V},\, \bm{p}} b_{i}^{k}(\bm{p}),
\end{equation}
where $i$ indexes the primitives, $b_i^k(\bm{p})$ is the blending weight of primitive $i$ at pixel $\bm{p}$ of view $k$, $\mathcal{V} \subseteq \mathcal{K}$ is the set of training views rendered since the previous relocation and $\bm{p}$ runs over the pixels of view $k$.
Unlike the opacity, $w_i$ accounts for occlusion through the transmittance, and a primitive not visible in any of these views has $w_i = 0$.
Pruning methods use the same quantity~\cite{niemeyer2025radsplat, Fang2024minisplatting2}.

\begin{figure*}[t!]
    \centering
    \includegraphics[width=0.98\linewidth]{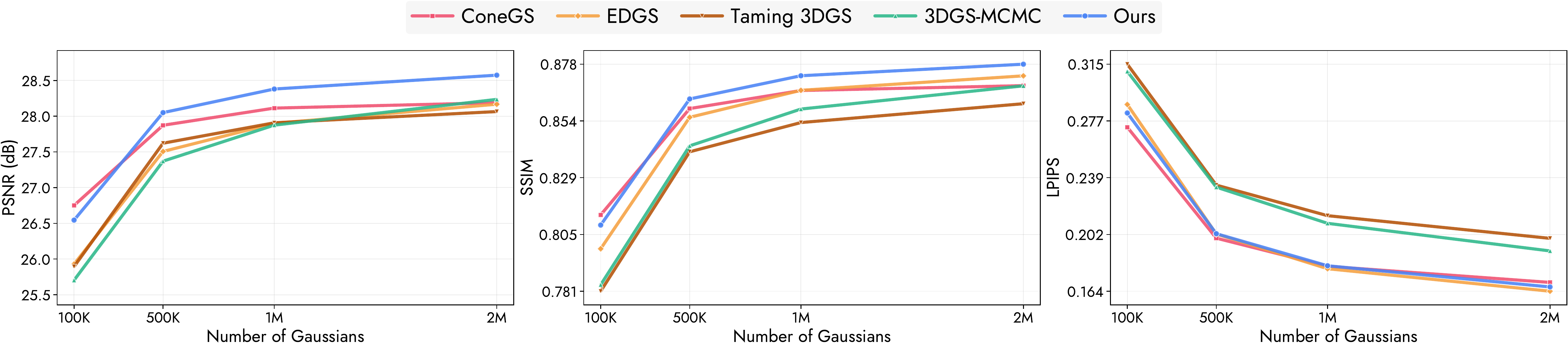}
    \caption{
\textbf{Quantitative Evaluation}:
Average PSNR, SSIM and LPIPS over the 24 scenes of four datasets~\cite{barron2022mipnerf360, Knapitsch2017tanks, ling2023dl3dv, hedman2018deepblending} at four primitive budgets.
Our method and its prior-free baseline 3DGS-MCMC~\cite{kheradmand2024mcmc} start from SfM initialization.
    }
    \label{fig:nvs_main}
\end{figure*}

Since we set no threshold on $w_i$, we need a schedule that sets how many primitives to relocate.
During the first 5000 iterations, we relocate only primitives with $w_i = 0$, at most 10\% of all primitives, to keep early training stable.
From iteration 5000 to 10k, we relocate the 10\% of primitives with the lowest $w_i$, and from 10k on, we scale this share by $\exp(-\frac{1}{2}((t - 10\text{k})/5\text{k})^2)$ until densification ends at iteration 25k.
This Gaussian fall-off lowers the share smoothly, to 61\% of its initial value at 15k, 14\% at 20k and about 1\% at 25k, so relocation fades out instead of ending abruptly.
We exempt primitives placed in a relocation from the next one.

\subsection{Position Sampling and Color Initialization}
\label{sec:method:initialization}
Each new primitive, added or relocated, starts as a copy of its source and then needs its own position and appearance.
We keep the adjustment of opacity $o_i$ and scale $\bm{s}_i$ of 3DGS-MCMC (Section~\ref{sec:mcmc}) and copy rotation, SH coefficients and confidence from the source.

For the position, 3DGS-MCMC relies on its continuous position noise (Section~\ref{sec:mcmc}).
In our experiments, this noise helps only primitives that are seen in many views and can disrupt optimization in background regions.
Windisch~\emph{et al.}~\cite{windisch2026ALoDOfGaussians} likewise report that it becomes problematic in large-scale reconstruction, where primitives are rarely observed.
We disable this noise.
In preliminary experiments, noise scaled by the learned confidence instead of the opacity also performed worse than no noise.
Instead, as suggested by GOF~\cite{yu2024gof}, we draw the mean $\bm{\mu}'$ of each new primitive from the Gaussian $\mathcal{N}(\bm{\mu}_i, \bm{\Sigma}'_i)$ of its source primitive $i$, whose covariance $\bm{\Sigma}'_i$ combines the rotation $\bm{R}_i$ with the adjusted scale.
Unlike the noise, this position sampling acts once and only on new primitives.

The color inherited from the source primitive may not match the new position, so we initialize it from a training image to speed up convergence.
We take the first training view after placement in which the new primitive is visible with a projected radius below 16 pixels, project $\bm{\mu}'$ into it and look up the ground-truth color $\bm{c}_\text{GT}$ by bilinear interpolation.
We then keep the higher-order SH coefficients of the source primitive, set the DC term, \ie the zeroth-order SH coefficient, so that the SH color evaluated in this view equals $\bm{c}_\text{GT}$, and reset the optimizer moments of the new primitive.
If no such view comes within 400 iterations, the new primitive keeps the inherited color.

\begin{table}[t]
\centering
\resizebox{\linewidth}{!}{%
\begin{tabular}{llcccc}
\toprule
 & & \multicolumn{4}{c}{$\Delta$ Ours $-$ 3DGS-MCMC} \\
\cmidrule(lr){3-6}
\textit{Metric} & \textit{Region} & $\mathrm{100k}$ & $\mathrm{500k}$ & $\mathrm{1M}$ & $\mathrm{2M}$ \\
\midrule
\multirow{2}{*}{PSNR $\uparrow$} & Background & \cellcolor[RGB]{255,211,211}$+0.69$ & \cellcolor[RGB]{255,179,179}$+1.20$ & \cellcolor[RGB]{255,190,190}$+1.03$ & \cellcolor[RGB]{255,195,195}$+0.94$ \\
 & Foreground & \cellcolor[RGB]{255,229,229}$+0.41$ & \cellcolor[RGB]{255,240,240}$+0.23$ & \cellcolor[RGB]{255,245,245}$+0.15$ & \cellcolor[RGB]{255,255,255}$-0.03$ \\
\midrule
\multirow{2}{*}{SSIM $\uparrow$} & Background & \cellcolor[RGB]{255,200,200}$+0.050$ & \cellcolor[RGB]{255,179,179}$+0.070$ & \cellcolor[RGB]{255,189,189}$+0.061$ & \cellcolor[RGB]{255,205,205}$+0.046$ \\
 & Foreground & \cellcolor[RGB]{255,233,233}$+0.021$ & \cellcolor[RGB]{255,234,234}$+0.019$ & \cellcolor[RGB]{255,241,241}$+0.013$ & \cellcolor[RGB]{255,247,247}$+0.007$ \\
\midrule
\multirow{2}{*}{LPIPS $\downarrow$} & Background & \cellcolor[RGB]{255,214,214}$-0.044$ & \cellcolor[RGB]{255,179,179}$-0.083$ & \cellcolor[RGB]{255,179,179}$-0.083$ & \cellcolor[RGB]{255,190,190}$-0.072$ \\
 & Foreground & \cellcolor[RGB]{255,244,244}$-0.012$ & \cellcolor[RGB]{255,237,237}$-0.020$ & \cellcolor[RGB]{255,234,234}$-0.023$ & \cellcolor[RGB]{255,233,233}$-0.024$ \\
\bottomrule
\end{tabular}}
\caption{\textbf{Foreground vs. background performance.} Mean metric differences between our method and 3DGS-MCMC across outdoor Mip-NeRF 360 scenes (SfM initialization), split by image region. Background is defined as the farthest 25\% of test pixels via COLMAP-aligned~\cite{schoenberger2016sfm,schoenberger2016mvs} Depth Anything V2~\cite{yang2024depthanythingv2}. Positive PSNR and SSIM and negative LPIPS differences favor our method. Gains are largest in the distant background.}
\label{tab:background}
\end{table}

\section{Evaluation}
\label{sec:eval}
\paragraph{Implementation Details.}
We implement our method in the 3DGS codebase~\cite{kerbl20233dgs} with the Faster-GS rasterizer and fused Adam optimizer~\cite{hahlbohm2026fastergs} and run 3DGS-MCMC in the same codebase; all other baselines run in their official codebases.
We train all our runs for 30k iterations.
If the initial SfM point cloud exceeds the budget, we subsample it by farthest point sampling.
We regularize scale and opacity with $\lambda_\text{reg} = 0.01$ for all scenes, and keep the regularizer active until densification ends at iteration 25k.
We use the confidence loss of Equation~\eqref{eq:confidence} throughout training, with $\beta_\text{conf}=0.075$ and the learning rate of CoMe~\cite{radl2026come}, and we clamp the confidence map $\hat{C}$ to $[10^{-3}, 5]$.

\paragraph{Datasets and Comparison Methods.}
We evaluate on the scenes used in 3DGS~\cite{kerbl20233dgs}, \ie, 9 scenes from Mip-NeRF 360~\cite{barron2022mipnerf360} and 2 scenes each from Tanks \& Temples~\cite{Knapitsch2017tanks} and Deep Blending~\cite{hedman2018deepblending}. Following Power Foam~\cite{govindarajan2026powerfoam}, we add 11 scenes from DL3DV~\cite{ling2023dl3dv}, for 24 real-world scenes in total.
We downsample the Mip-NeRF 360 outdoor scenes by 4 and the indoor scenes and DL3DV by 2, use Tanks \& Temples and Deep Blending at the provided resolution, and hold out every eighth image for testing.

The main baselines for our fixed-budget method areConeGS~\cite{baranowski2026conegs}, EDGS~\cite{Kotovenko2026edgs}, Taming 3DGS~\cite{mallick2024taming} and 3DGS-MCMC~\cite{kheradmand2024mcmc}, which also never exceed the budget during optimization.
For EDGS, we match the number of primitives at initialization, as done in ConeGS~\cite{baranowski2026conegs}.
We also compare against variable-budget methods, which determine the final primitive count dynamically during optimization: 3DGS~\cite{kerbl20233dgs}, DashGaussian~\cite{chen2025dashgaussian} and the simplification methods Mini-Splatting2~\cite{Fang2024minisplatting2} and GaussianSpa~\cite{zhang2025gaussianspa}.
We report PSNR, SSIM~\cite{Wang2004Structural} and LPIPS~\cite{zhang2018unreasonable} metrics, with images normalized to $[0,1]$ and using the VGG backbone for LPIPS.

\subsection{Novel View Synthesis}
\label{sec:results}
\begin{table*}[t]
\centering
\footnotesize
\begin{tabular}{llcccccccccccc}
\toprule
 & & \multicolumn{3}{c}{Mip-NeRF 360} & \multicolumn{3}{c}{Tanks \& Temples} & \multicolumn{3}{c}{Deep Blending} & \multicolumn{3}{c}{DL3DV} \\
\cmidrule(lr){3-5} \cmidrule(lr){6-8} \cmidrule(lr){9-11} \cmidrule(lr){12-14}
\textit{Budget} & \textit{Method} & PSNR & SSIM & LPIPS & PSNR & SSIM & LPIPS & PSNR & SSIM & LPIPS & PSNR & SSIM & LPIPS \\
\midrule
\multirow{5}{*}{$\mathrm{100k}$} & Ours & \cellcolor{tabsecond}26.32 & \cellcolor{tabsecond}0.748 & \cellcolor{tabthird}0.336 & \cellcolor{tabfirst}23.47 & \cellcolor{tabfirst}0.815 & \cellcolor{tabsecond}0.251 & \cellcolor{tabsecond}29.55 & \cellcolor{tabfirst}0.899 & \cellcolor{tabfirst}0.283 & \cellcolor{tabsecond}26.74 & \cellcolor{tabsecond}0.842 & \cellcolor{tabsecond}0.244 \\
 & ConeGS & \cellcolor{tabfirst}26.38 & \cellcolor{tabfirst}0.750 & \cellcolor{tabfirst}0.326 & \cellcolor{tabsecond}23.36 & \cellcolor{tabsecond}0.811 & \cellcolor{tabfirst}0.240 & \cellcolor{tabfirst}29.81 & \cellcolor{tabsecond}0.896 & \cellcolor{tabsecond}0.286 & \cellcolor{tabfirst}27.11 & \cellcolor{tabfirst}0.850 & \cellcolor{tabfirst}0.233 \\
 & EDGS & \cellcolor{tabthird}25.75 & \cellcolor{tabthird}0.741 & \cellcolor{tabsecond}0.333 & 22.69 & \cellcolor{tabthird}0.802 & \cellcolor{tabthird}0.254 & \cellcolor{tabthird}28.99 & \cellcolor{tabthird}0.890 & \cellcolor{tabthird}0.295 & 26.10 & \cellcolor{tabthird}0.829 & \cellcolor{tabthird}0.256 \\
 & Taming 3DGS & 25.57 & 0.715 & 0.367 & \cellcolor{tabthird}22.73 & 0.785 & 0.286 & 28.94 & 0.884 & 0.310 & \cellcolor{tabthird}26.18 & 0.814 & 0.278 \\
 & 3DGS-MCMC & \cellcolor{tabthird}25.75 & 0.726 & 0.354 & 22.24 & 0.782 & 0.282 & 28.86 & 0.887 & 0.308 & 25.72 & 0.811 & 0.279 \\
\midrule
\multirow{5}{*}{$\mathrm{500k}$} & Ours & \cellcolor{tabfirst}27.84 & \cellcolor{tabfirst}0.819 & \cellcolor{tabthird}0.231 & \cellcolor{tabfirst}24.30 & \cellcolor{tabfirst}0.860 & \cellcolor{tabthird}0.164 & \cellcolor{tabsecond}30.06 & \cellcolor{tabfirst}0.912 & \cellcolor{tabfirst}0.241 & \cellcolor{tabfirst}28.54 & \cellcolor{tabfirst}0.891 & \cellcolor{tabsecond}0.178 \\
 & ConeGS & \cellcolor{tabsecond}27.52 & \cellcolor{tabthird}0.811 & \cellcolor{tabsecond}0.227 & \cellcolor{tabthird}23.95 & \cellcolor{tabthird}0.851 & \cellcolor{tabsecond}0.160 & \cellcolor{tabfirst}30.21 & \cellcolor{tabsecond}0.907 & \cellcolor{tabsecond}0.243 & \cellcolor{tabsecond}28.46 & \cellcolor{tabfirst}0.891 & \cellcolor{tabfirst}0.175 \\
 & EDGS & 27.25 & \cellcolor{tabsecond}0.812 & \cellcolor{tabfirst}0.226 & 23.92 & \cellcolor{tabsecond}0.857 & \cellcolor{tabfirst}0.159 & \cellcolor{tabthird}29.83 & \cellcolor{tabthird}0.906 & \cellcolor{tabthird}0.244 & 27.94 & \cellcolor{tabthird}0.881 & \cellcolor{tabthird}0.183 \\
 & Taming 3DGS & \cellcolor{tabthird}27.36 & 0.787 & 0.273 & \cellcolor{tabsecond}24.07 & 0.842 & 0.196 & 29.81 & 0.900 & 0.263 & \cellcolor{tabthird}28.08 & 0.873 & 0.205 \\
 & 3DGS-MCMC & 27.30 & 0.797 & 0.260 & 23.81 & 0.843 & 0.198 & 29.07 & 0.900 & 0.277 & 27.76 & 0.871 & 0.209 \\
\bottomrule
\end{tabular}
\caption{
\textbf{Per-Dataset Results}:
Means over the scenes of each dataset for the fixed-budget methods at budgets of $\mathrm{100k}$ and $\mathrm{500k}$, with SfM initialization for our method.
The values are the dataset averages of supplementary Tables~S.1 and~S.2.
In each block, red, orange and yellow mark the best, second and third value of each column; equal values share a color.
}
\label{tab:per_dataset}
\end{table*}

\paragraph{Fixed-Budget Methods.}
Following ConeGS~\cite{baranowski2026conegs}, we evaluate a set of primitive budgets $\left\{\mathrm{100k}, \mathrm{500k}, \mathrm{1M}, \mathrm{2M}\right\}$ and average the metrics over the 24 scenes (Figure~\ref{fig:nvs_main}).
Table~\ref{tab:per_dataset} and supplementary Tables~\ref{tab:results-100000}--\ref{tab:results-2000000} break the results down by dataset and scene.
We treat differences below 0.05\,dB PSNR or 0.002 SSIM or LPIPS as on par.

Our error-driven densification helps most where primitives are scarce.
It is better than its prior-free baseline 3DGS-MCMC in all three metrics at every budget, by 0.84\,dB PSNR at $\mathrm{100k}$ and 0.34\,dB at $\mathrm{2M}$.
On the outdoor Mip-NeRF 360 scenes, most of this gain lies in the background (Table~\ref{tab:background}): from $\mathrm{500k}$ on, our PSNR there improves by about 1\,dB, against at most 0.23\,dB in the foreground.
When inspecting the distribution of blended primitives for low-budget models (\cf Figure~\ref{fig:num_blended}), we find our primitives are not as concentrated on the foreground as those in 3DGS-MCMC. 

Prior-based methods show clear benefits for smaller budgets, but their lead over 3DGS-MCMC shrinks as the budget grows.
ConeGS is 1.05\,dB ahead of 3DGS-MCMC in PSNR at $\mathrm{100k}$, whereas at $\mathrm{2M}$, 3DGS-MCMC is on par with ConeGS and ahead of EDGS.
Our method preserves the steady improvement of 3DGS-MCMC with higher budgets \emph{on top} of its established low-budget advantage.
From $\mathrm{500k}$ on, it has the highest PSNR and SSIM of all fixed-budget methods, and its PSNR lead over the second-best method grows from 0.18\,dB at $\mathrm{500k}$ to 0.34\,dB at $\mathrm{2M}$.
Intuitively, high-quality densification outweighs prior-based initialization as the number of primitives grows larger. 

At $\mathrm{100k}$, ConeGS is the only method ahead of ours, in all three metrics and by 0.21\,dB PSNR, mainly on DL3DV (Table~\ref{tab:per_dataset}).
Note that the priors used by ConeGS require a dedicated initialization stage and generate associated overhead: iNGP pre-training alone takes almost as long as our entire optimization (Table~\ref{tab:memory}).
Started from their initialization, our method matches ConeGS in PSNR and SSIM (Section~\ref{sec:analysis}).

LPIPS is our weakest metric, as it penalizes missing high-frequency detail more than PSNR does, and such detail profits most from a dense prior or a larger intermediate set of primitives.
From SfM points, our method never has the best 24-scene LPIPS, but from $\mathrm{500k}$ on, it trails ConeGS or EDGS by at most 0.003.
At equal budget, the gap is confined to outdoor scenes (Table~\ref{tab:fixed_prim}), and it disappears when our method starts from a dense prior initialization (Section~\ref{sec:analysis}) or is allowed to exceed the budget during training (Figure~\ref{fig:primitive_cap}).
Figure~\ref{fig:comparison} compares test views of seven scenes at $\mathrm{500k}$.

\def\cmpZoomGarden{0.51,0.73}
\def\cmpZoomBicycle{0.82,0.81}
\def\cmpZoomRoom{0.49,0.93}
\def\cmpZoomTrain{0.87,0.89}
\def\cmpZoomPlayroom{0.70,0.54}
\def\cmpZoomCar{0.30,0.30}
\def\cmpZoomCounter{0.08,0.92}
\newlength{\cmpW}\setlength{\cmpW}{0.192\textwidth}
\newlength{\cmpH}\setlength{\cmpH}{0.6667\cmpW}
\newcommand{\cmpImage}[4]{%
  \begin{tikzpicture}[baseline=(img.south), spy using outlines={rectangle, draw=red, line width=0.5pt,
      magnification=3, width=0.42\cmpW, height=0.42\cmpH}]
    \IfFileExists{figures/comparison/#1_#2.jpg}{%
      \node[anchor=south west, inner sep=0] (img) at (0,0)
        {\includegraphics[width=\cmpW]{figures/comparison/#1_#2.jpg}};}{%
      \node[anchor=south west, inner sep=0, fill=black!8, draw=black!25,
        minimum width=\cmpW, minimum height=\cmpH] (img) at (0,0) {};
      \node[anchor=north west, inner sep=2pt, font=\scriptsize\ttfamily, text=black!60] at (img.north west) {#1\_#2.jpg};}
    \node[anchor=north, inner sep=1pt, font=\scriptsize] at (img.south) {\strut #4};
    \begin{scope}[x={(img.south east)}, y={(img.north west)}]
      \coordinate (zoomcenter) at (#3);
    \end{scope}
    \spy on (zoomcenter) in node[anchor=south east] at (img.south east);
  \end{tikzpicture}}
\newcommand{\cmpRow}[1]{\rotatebox{90}{\makebox[\cmpH]{\small #1}}}
\newcommand{\cmpHead}[1]{\makebox[\cmpW]{\small #1}}

\begin{figure*}[p]
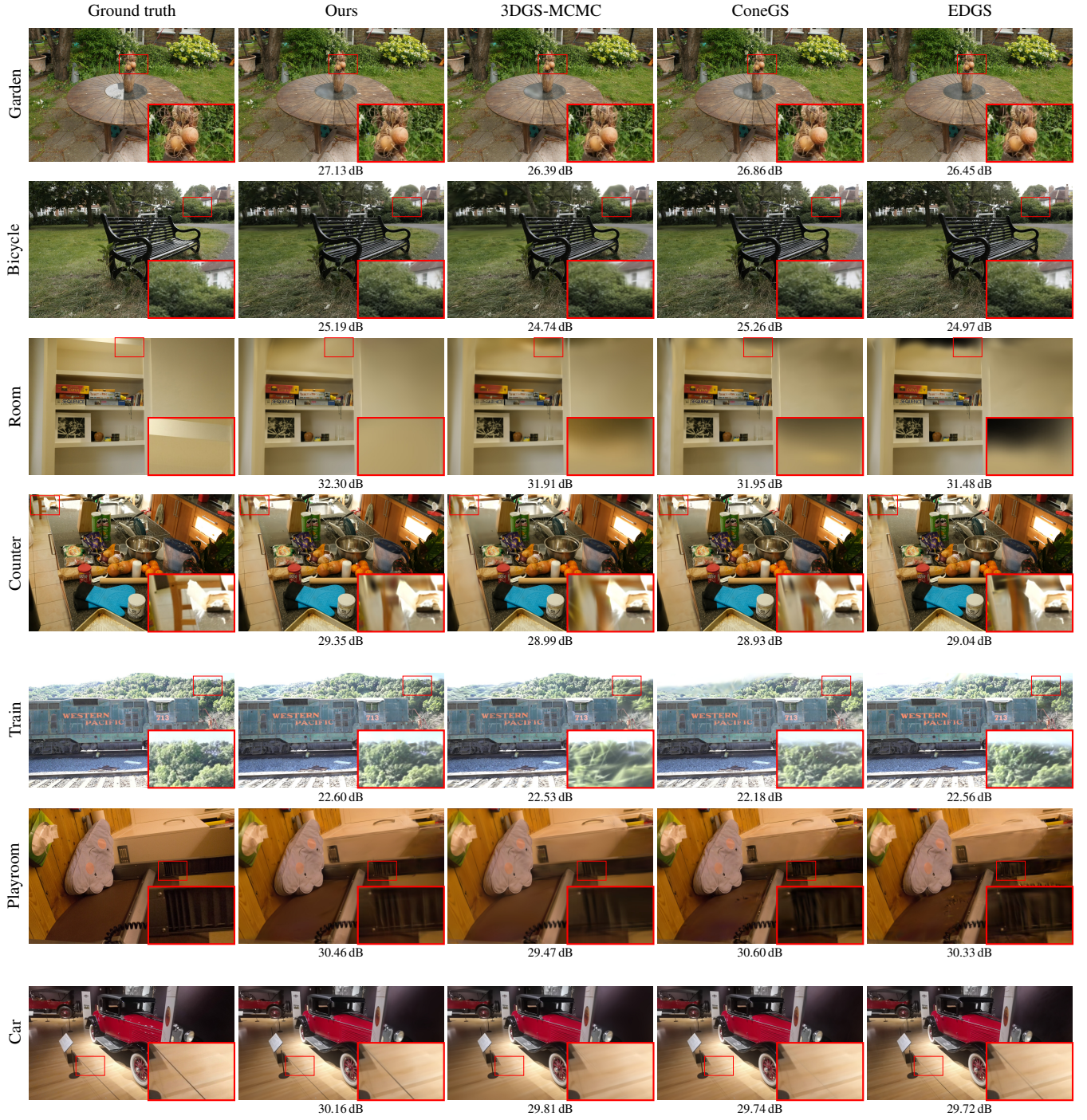

\centering
\setlength{\tabcolsep}{0pt}
\begin{tabular}{c@{\hspace{3pt}}c@{\hspace{1.5pt}}c@{\hspace{1.5pt}}c@{\hspace{1.5pt}}c@{\hspace{1.5pt}}c}
 & \cmpHead{Ground truth} & \cmpHead{Ours} & \cmpHead{3DGS-MCMC} & \cmpHead{ConeGS} & \cmpHead{EDGS} \\[1pt]
\cmpRow{Garden} &
  \cmpImage{garden}{gt}{\cmpZoomGarden}{} &
  \cmpImage{garden}{ours}{\cmpZoomGarden}{27.13\,dB} &
  \cmpImage{garden}{mcmc}{\cmpZoomGarden}{26.39\,dB} &
  \cmpImage{garden}{conegs}{\cmpZoomGarden}{26.86\,dB} &
  \cmpImage{garden}{edgs}{\cmpZoomGarden}{26.45\,dB} \\[2pt]
\cmpRow{Bicycle} &
  \cmpImage{bicycle}{gt}{\cmpZoomBicycle}{} &
  \cmpImage{bicycle}{ours}{\cmpZoomBicycle}{25.19\,dB} &
  \cmpImage{bicycle}{mcmc}{\cmpZoomBicycle}{24.74\,dB} &
  \cmpImage{bicycle}{conegs}{\cmpZoomBicycle}{25.26\,dB} &
  \cmpImage{bicycle}{edgs}{\cmpZoomBicycle}{24.97\,dB} \\[2pt]
\cmpRow{Room} &
  \cmpImage{room}{gt}{\cmpZoomRoom}{} &
  \cmpImage{room}{ours}{\cmpZoomRoom}{32.30\,dB} &
  \cmpImage{room}{mcmc}{\cmpZoomRoom}{31.91\,dB} &
  \cmpImage{room}{conegs}{\cmpZoomRoom}{31.95\,dB} &
  \cmpImage{room}{edgs}{\cmpZoomRoom}{31.48\,dB} \\[2pt]
\cmpRow{Counter} &
  \cmpImage{counter}{gt}{\cmpZoomCounter}{} &
  \cmpImage{counter}{ours}{\cmpZoomCounter}{29.35\,dB} &
  \cmpImage{counter}{mcmc}{\cmpZoomCounter}{28.99\,dB} &
  \cmpImage{counter}{conegs}{\cmpZoomCounter}{28.93\,dB} &
  \cmpImage{counter}{edgs}{\cmpZoomCounter}{29.04\,dB} \\[2pt]
\cmpRow{Train} &
  \cmpImage{train}{gt}{\cmpZoomTrain}{} &
  \cmpImage{train}{ours}{\cmpZoomTrain}{22.60\,dB} &
  \cmpImage{train}{mcmc}{\cmpZoomTrain}{22.53\,dB} &
  \cmpImage{train}{conegs}{\cmpZoomTrain}{22.18\,dB} &
  \cmpImage{train}{edgs}{\cmpZoomTrain}{22.56\,dB} \\[2pt]
\cmpRow{Playroom} &
  \cmpImage{playroom}{gt}{\cmpZoomPlayroom}{} &
  \cmpImage{playroom}{ours}{\cmpZoomPlayroom}{30.46\,dB} &
  \cmpImage{playroom}{mcmc}{\cmpZoomPlayroom}{29.47\,dB} &
  \cmpImage{playroom}{conegs}{\cmpZoomPlayroom}{30.60\,dB} &
  \cmpImage{playroom}{edgs}{\cmpZoomPlayroom}{30.33\,dB} \\[2pt]
\cmpRow{Car} &
  \cmpImage{car}{gt}{\cmpZoomCar}{} &
  \cmpImage{car}{ours}{\cmpZoomCar}{30.16\,dB} &
  \cmpImage{car}{mcmc}{\cmpZoomCar}{29.81\,dB} &
  \cmpImage{car}{conegs}{\cmpZoomCar}{29.74\,dB} &
  \cmpImage{car}{edgs}{\cmpZoomCar}{29.72\,dB} \\[2pt]
\end{tabular}
\caption{
\textbf{Qualitative Comparison at $\mathrm{500k}$ Primitives}:
Test views of seven scenes from Mip-NeRF 360 (Garden, Bicycle, Room, Counter), Tanks \& Temples (Train), Deep Blending (Playroom) and DL3DV (Car).
The columns show the ground truth, our method and 3DGS-MCMC~\cite{kheradmand2024mcmc}, both with SfM initialization, and the prior-based methods ConeGS~\cite{baranowski2026conegs} and EDGS~\cite{Kotovenko2026edgs}, all at a budget of $\mathrm{500k}$.
The label under each image is the per-scene test PSNR of the method, averaged over all test views of the scene (supplementary Table~S.2); it is not the PSNR of the view shown.
Each inset magnifies the marked background region by $3\times$, illustrating how much background detail each method retains at this low budget.
}
\label{fig:comparison}
\end{figure*}

\begin{table*}[t]
    \centering
    \small
    \begin{tabular}{lcccccccccccc@{\hspace{12pt}}r}
    \toprule
     & \multicolumn{3}{c}{Mip-NeRF 360 Outdoor} & \multicolumn{3}{c}{Mip-NeRF 360 Indoor} & \multicolumn{3}{c}{T\&T \& DB} & \multicolumn{3}{c}{DL3DV} \\
    \cmidrule(lr){2-4} \cmidrule(lr){5-7} \cmidrule(lr){8-10} \cmidrule(lr){11-13}
     \textit{Method} & PSNR & SSIM & LPIPS & PSNR & SSIM & LPIPS & PSNR & SSIM & LPIPS & PSNR & SSIM & LPIPS & \#G\\
    \midrule
    Ours & \cellcolor{tabfirst}25.04 & \cellcolor{tabfirst}0.754 & \cellcolor{tabfirst}0.189 & \cellcolor{tabfirst}31.92 & \cellcolor{tabfirst}0.933 & \cellcolor{tabfirst}0.165 & \cellcolor{tabfirst}27.33 &  \cellcolor{tabfirst}0.888 &  \cellcolor{tabfirst}0.179 &    \cellcolor{tabfirst}29.13 & \cellcolor{tabfirst}0.906 & \cellcolor{tabfirst}0.155 \\
    3DGS-MCMC & 24.94 & 0.744 & 0.220 & 31.61 & 0.931 & 0.177 &  26.41 & 0.882 & 0.213 & 
    28.90 & 0.901 & 0.168 &
1.94M\\
    3DGS & 24.71 & 0.730 & 0.236 & 30.78 & 0.923 & 0.188 & 26.69 & 0.880 & 0.203 & 27.85 & 0.884 & 0.187\\
    \midrule
    Ours & \cellcolor{tabfirst}25.07 &  \cellcolor{tabfirst}0.754 &  \cellcolor{tabfirst}0.189 &  \cellcolor{tabfirst}31.81 &  \cellcolor{tabfirst}0.932 &  \cellcolor{tabfirst}0.168 &  \cellcolor{tabfirst}27.25 &  \cellcolor{tabfirst}0.887 &  \cellcolor{tabfirst}0.182 &  \cellcolor{tabfirst}28.84 &  \cellcolor{tabfirst}0.898 &  \cellcolor{tabfirst}0.168\\
    3DGS-MCMC & 24.93 & 0.743 & 0.221 & 31.45 & 0.929 & 0.181 & 26.63 & 0.881 & 0.214 & 28.54 & 0.894 & 0.180 & 1.55M \\
    DashGaussian %
    & 25.04 & 0.741 & 0.227 & 31.04 & 0.921 & 0.193 & 26.81 & 0.879 & 0.210 & 28.34 & 0.889 & 0.182\\
    \midrule
    Ours & \cellcolor{tabfirst}25.05 & \cellcolor{tabfirst}0.754 & \cellcolor{tabfirst}0.190 & \cellcolor{tabfirst}31.75 & \cellcolor{tabfirst}0.931 & \cellcolor{tabfirst}0.170 & \cellcolor{tabfirst}27.40 & \cellcolor{tabfirst}0.888 & \cellcolor{tabfirst}0.184 &
    \cellcolor{tabfirst}29.00 & \cellcolor{tabfirst}0.904 & \cellcolor{tabfirst}0.160\\
    3DGS-MCMC & 24.94 & 0.743 & 0.222 & 31.31 & 0.928 & 0.185 & 26.58 & 0.879 & 0.217 & 
    28.68 & 0.898 & 0.173
    & 1.51M\\
    3DGS + GOF & 24.67 & 0.746 & 0.206 & 30.72 & 0.922 & 0.187 & 26.64 & 0.883 & 0.200 & 28.08 & 0.894 & 0.175 \\
    \midrule
    Ours &  \cellcolor{tabfirst}25.02 &  \cellcolor{tabfirst}0.738 & 0.256 &  \cellcolor{tabfirst}31.59 &  \cellcolor{tabfirst}0.929 &  \cellcolor{tabfirst}0.176 &  \cellcolor{tabfirst}27.06 &  \cellcolor{tabfirst}0.880 &  \cellcolor{tabfirst}0.209 &  \cellcolor{tabfirst}28.18 &  \cellcolor{tabfirst}0.889 &  \cellcolor{tabfirst}0.186\\
    3DGS-MCMC & 24.50 & 0.705 & 0.298 & 31.09 & 0.922 & 0.194 & 26.09 & 0.861 & 0.247 &  27.78 & 0.880 & 0.199 & 583k\\
    Mini-Splatting2%
    & 24.63 & 0.736 &  \cellcolor{tabfirst}0.244 & 30.76 & 0.926 & 0.178 & 26.58 & 0.876 & 0.213 & 27.29 & 0.878 & 0.194 \\
    \midrule
    Ours &  \cellcolor{tabfirst}24.97 & 0.736 & 0.259 &  \cellcolor{tabfirst}31.38 &  \cellcolor{tabfirst}0.927 &  \cellcolor{tabfirst}0.183 &  \cellcolor{tabfirst}27.19 &  \cellcolor{tabfirst}0.883 & 0.205 & \cellcolor{tabfirst}28.44 & \cellcolor{tabfirst}0.891 & \cellcolor{tabfirst}0.180\\
    3DGS-MCMC & 24.47 & 0.703 & 0.301 & 30.81 & 0.919 & 0.201 & 26.43 & 0.866 & 0.243 & 28.04 & 0.882 & 0.194 & 515k\\
    GaussianSpa%
    & 24.80 &  \cellcolor{tabfirst}0.741 &  \cellcolor{tabfirst}0.245 & 30.54 & 0.924 & 0.187 & 26.67 & 0.882 &  \cellcolor{tabfirst}0.204 & 27.69 & 0.884 & 0.185 \\
    \bottomrule
    \end{tabular}
    \caption{
\textbf{Equal-Budget Comparison}:
In each block, we compare our method and 3DGS-MCMC with one variable-budget method at its per-scene primitive count: 3DGS, 3DGS + GOF, DashGaussian~\cite{kerbl20233dgs, yu2024gof, chen2025dashgaussian} and the simplification methods Mini-Splatting2 and GaussianSpa~\cite{Fang2024minisplatting2, zhang2025gaussianspa}.
\#G is the mean primitive budget of each block.
Columns average the Mip-NeRF 360 outdoor and indoor scenes, the Tanks \& Temples (T\&T) and Deep Blending (DB) scenes, and the DL3DV scenes.
Red shading marks the best value in each block.
}
    \label{tab:fixed_prim}
    \end{table*}

\paragraph{Variable-Budget Methods.}
Our method also holds up against methods that do not explicitly limit their own primitive count.
For each comparison, we set the budget of our method and 3DGS-MCMC to the final per-scene primitive count of the compared variable-budget method.
We also include 3DGS with absolute gradients for densification (3DGS + GOF)~\cite{yu2024gof}.
Against 3DGS, DashGaussian and 3DGS + GOF, our method is better in all three metrics in every group of scenes, including DL3DV, and only tied by DashGaussian in outdoor PSNR (Table~\ref{tab:fixed_prim}, supplementary Figure~\ref{fig:supp-variable-budget}).
It also beats 3DGS-MCMC in all metrics in every block.

The simplification methods Mini-Splatting2 and GaussianSpa are ahead of us only in outdoor scenes.
There, both show better LPIPS, and GaussianSpa also higher SSIM, while our method keeps the lead in PSNR.
Indoors and on Tanks \& Temples and Deep Blending, our method is ahead of both or on par, and on DL3DV, it leads both in all three metrics.
Both methods first optimize a much larger set of primitives, up to 6M for the Bicycle scene, and then prune it to the final count.
Outdoor scenes hold the most fine, distant texture, which such a large intermediate set can cover before pruning keeps the most useful primitives.
When we similarly let our method exceed the budget during training, its LPIPS reaches that of both methods (Section~\ref{sec:analysis}, Figure~\ref{fig:primitive_cap}).
However, at the matched budget on Mip-NeRF 360, Mini-Splatting2 and GaussianSpa need $1.7\times$ and $2.4\times$ our peak memory, and GaussianSpa trains $8\times$ as long (Table~\ref{tab:memory}).
Only Mini-Splatting2 trains faster, as it stops after 18k iterations.

\begin{table}[t]
    \centering
    \setlength{\tabcolsep}{4.5pt}
    \resizebox{\linewidth}{!}{%
    \begin{tabular}{lcccccc}
    \toprule
    & \multicolumn{3}{c}{$\mathrm{100k}$} & \multicolumn{3}{c}{$\mathrm{500k}$} \\
    \cmidrule(lr){2-4} \cmidrule(lr){5-7}
    \textit{Configuration} & PSNR & SSIM & LPIPS & PSNR & SSIM & LPIPS \\
    \midrule
    Ours (full) & \cellcolor[RGB]{255,179,179}26.32 & \cellcolor[RGB]{255,179,179}0.748 & \cellcolor[RGB]{255,183,179}0.336 & \cellcolor[RGB]{255,179,179}27.84 & \cellcolor[RGB]{255,179,179}0.819 & \cellcolor[RGB]{255,179,179}0.231 \\
    \midrule
    \quad w/ MCMC noise & \cellcolor[RGB]{255,194,179}26.17 & \cellcolor[RGB]{255,200,179}0.741 & \cellcolor[RGB]{255,201,179}0.341 & \cellcolor[RGB]{255,198,179}27.62 & \cellcolor[RGB]{255,206,179}0.808 & \cellcolor[RGB]{255,210,179}0.244 \\
    \quad w/o color initialization & \cellcolor[RGB]{255,189,179}26.22 & \cellcolor[RGB]{255,191,179}0.744 & \cellcolor[RGB]{255,201,179}0.341 & \cellcolor[RGB]{255,193,179}27.68 & \cellcolor[RGB]{255,186,179}0.816 & \cellcolor[RGB]{255,191,179}0.236 \\
    \quad w/o position sampling & \cellcolor[RGB]{255,225,180}25.87 & \cellcolor[RGB]{255,200,179}0.741 & \cellcolor[RGB]{255,204,179}0.342 & \cellcolor[RGB]{255,253,180}26.96 & \cellcolor[RGB]{255,255,180}0.788 & \cellcolor[RGB]{255,253,180}0.262 \\
    \midrule
    \quad w/ MCMC sampling & \cellcolor[RGB]{255,197,179}26.14 & \cellcolor[RGB]{255,191,179}0.744 & \cellcolor[RGB]{255,190,179}0.338 & \cellcolor[RGB]{255,187,179}27.74 & \cellcolor[RGB]{255,191,179}0.814 & \cellcolor[RGB]{255,198,179}0.239 \\
    \quad w/ MCMC relocation & \cellcolor[RGB]{255,237,180}25.75 & \cellcolor[RGB]{255,231,180}0.731 & \cellcolor[RGB]{255,215,179}0.345 & \cellcolor[RGB]{255,231,180}27.23 & \cellcolor[RGB]{255,216,179}0.804 & \cellcolor[RGB]{255,222,180}0.249 \\
    \quad w/ $\beta_\text{conf}$ = 0.05 & \cellcolor[RGB]{255,180,179}26.31 & \cellcolor[RGB]{255,179,179}0.748 & \cellcolor[RGB]{255,179,179}0.335 & \cellcolor[RGB]{255,186,179}27.76 & \cellcolor[RGB]{255,184,179}0.817 & \cellcolor[RGB]{255,181,179}0.232 \\
    \quad w/ $\beta_\text{conf}$ = 0.1 & \cellcolor[RGB]{255,180,179}26.31 & \cellcolor[RGB]{255,179,179}0.748 & \cellcolor[RGB]{255,183,179}0.336 & \cellcolor[RGB]{255,181,179}27.82 & \cellcolor[RGB]{255,184,179}0.817 & \cellcolor[RGB]{255,184,179}0.233 \\
    \midrule
    3DGS-MCMC & \cellcolor[RGB]{255,237,180}25.75 & \cellcolor[RGB]{255,246,180}0.726 & \cellcolor[RGB]{255,248,180}0.354 & \cellcolor[RGB]{255,225,180}27.30 & \cellcolor[RGB]{255,233,180}0.797 & \cellcolor[RGB]{255,248,180}0.260 \\
    \quad w/o noise & \cellcolor[RGB]{255,255,180}25.57 & \cellcolor[RGB]{255,255,180}0.723 & \cellcolor[RGB]{255,251,180}0.355 & \cellcolor[RGB]{255,255,180}26.94 & \cellcolor[RGB]{255,255,180}0.788 & \cellcolor[RGB]{255,255,180}0.263 \\
    \quad w/ our relocation & \cellcolor[RGB]{255,228,180}25.84 & \cellcolor[RGB]{255,231,180}0.731 & \cellcolor[RGB]{255,255,180}0.356 & \cellcolor[RGB]{255,220,180}27.35 & \cellcolor[RGB]{255,226,180}0.800 & \cellcolor[RGB]{255,243,180}0.258 \\
    \quad w/ our sampling & \cellcolor[RGB]{255,234,180}25.78 & \cellcolor[RGB]{255,246,180}0.726 & \cellcolor[RGB]{255,251,180}0.355 & \cellcolor[RGB]{255,224,180}27.31 & \cellcolor[RGB]{255,233,180}0.797 & \cellcolor[RGB]{255,255,180}0.263 \\
    \bottomrule
    \end{tabular}%
    }
    \caption{
\textbf{Ablation Study}:
Means over the 9 Mip-NeRF 360 scenes~\cite{barron2022mipnerf360} at budgets of $\mathrm{100k}$ and $\mathrm{500k}$.
In each row of the middle blocks, we remove, replace or add one component of our method, whose default is $\beta_\text{conf} = 0.075$.
In \emph{w/o position sampling}, we remove all positional randomness.
The bottom block shows 3DGS-MCMC with and without its continuous position noise, and with either our relocation (\emph{w/ our relocation}) or our sampling distribution (\emph{w/ our sampling}) in place of its opacity rule.
Our full configuration is on par or best in every column. 
Shading runs from red for the best to yellow for the worst value in each column.
}
    \label{tab:ablation}
\end{table}

\subsection{Analysis}
\label{sec:analysis}
\paragraph{Ablation Study.}
The gain of our method comes from its components acting together, at the two budgets of $\mathrm{100k}$ and $\mathrm{500k}$ that we test (Table~\ref{tab:ablation}), and the relocation rule shows this most clearly.
With the opacity threshold of 3DGS-MCMC in its place (\emph{w/ 3DGS-MCMC relocation}), our method loses about 0.6\,dB at both budgets and falls back to the PSNR of 3DGS-MCMC, although its SSIM and LPIPS stay better.
Conversely, neither half helps 3DGS-MCMC on its own: our relocation by blending weight (\emph{w/ our relocation}) and our sampling distribution (\emph{w/ our sampling}) each add less than 0.1\,dB, while the full method is more than 0.5\,dB ahead.
Relocation decides which primitives move, and sampling and placement decide where they go.
In 3DGS-MCMC, the freed primitives are still sent to opaque sources, mostly in the foreground.
Each of the other components adds 0.1 to 0.2\,dB.
Our method is also robust to the confidence parameter: setting $\beta_\text{conf}$ to 0.05 or 0.1 instead of 0.075 changes PSNR by less than 0.1\,dB at both budgets.
Likewise, our relocation share of 10\% is within 0.06\,dB PSNR of the best share at every budget from $\mathrm{100k}$ to $\mathrm{2M}$ (supplementary Table~\ref{tab:explore}).

For exploration from prior-free initialization, some positional randomness is needed, and sampling the position of each new primitive once provides enough of it.
Without any (\emph{w/o position sampling}), our PSNR falls by up to 0.88\,dB, and 3DGS-MCMC loses up to 0.36\,dB without its noise (\emph{w/o noise}), since each new primitive then sits exactly on its source and the optimizer struggles to separate the two.
Re-adding the continuous noise of 3DGS-MCMC on top of position sampling costs up to 0.22\,dB (\emph{w/ 3DGS-MCMC noise}).

\begin{table}[t]
\centering
\small
\setlength{\tabcolsep}{4pt}
\begin{tabular}{lccrrr}
    \toprule
    Distribution & w/ conf. & w/ 20\% SSIM & PSNR & SSIM & LPIPS \\
    \midrule
    \multirow{4}{*}{Abs.\ gradients} & \cmark & \cmark & \cellcolor[RGB]{255,179,179}27.84 & \cellcolor[RGB]{255,179,179}0.819 & \cellcolor[RGB]{255,185,179}0.231\\
     & \cmark & - & \cellcolor[RGB]{255,218,179}27.70 & \cellcolor[RGB]{255,189,179}0.818 & \cellcolor[RGB]{255,185,179}0.231\\
     & - & \cmark & \cellcolor[RGB]{255,207,179}27.74 & \cellcolor[RGB]{255,211,179}0.816 & \cellcolor[RGB]{255,179,179}0.230 \\
     & - & - & \cellcolor[RGB]{255,243,179}27.61 & \cellcolor[RGB]{255,211,179}0.816 & \cellcolor[RGB]{255,179,179}0.230\\
    \midrule
    \multirow{2}{*}{SSIM} & \cmark &  - & \cellcolor[RGB]{255,212,179}27.72 & \cellcolor[RGB]{255,189,179}0.818 & \cellcolor[RGB]{255,185,179}0.231 \\
     & - & - & \cellcolor[RGB]{255,218,179}27.70 & \cellcolor[RGB]{255,255,180}0.812 & \cellcolor[RGB]{255,255,180}0.241 \\
    \midrule
    \multirow{2}{*}{Confidence} & - & \cmark & \cellcolor[RGB]{255,255,180}27.57 & \cellcolor[RGB]{255,233,179}0.814 & \cellcolor[RGB]{255,206,179}0.234\\
     & - & - & \cellcolor[RGB]{255,249,179}27.59 & \cellcolor[RGB]{255,233,179}0.814 & \cellcolor[RGB]{255,199,179}0.233\\
     \midrule
    {Opacity} & - & - & \cellcolor[RGB]{255,207,179}27.74 & \cellcolor[RGB]{255,233,179}0.814 & \cellcolor[RGB]{255,243,179}0.239\\
    \bottomrule
\end{tabular}
\caption{
    \textbf{Sampling Ablation Study}:
    Distributions for drawing source primitives on the 9 Mip-NeRF 360 scenes~\cite{barron2022mipnerf360} at a budget of $\mathrm{500k}$.
    Each block starts from absolute gradients, the SSIM term or the confidence alone.
    With the first check mark, we weight this distribution by the confidence factor, and with the second, we draw 20\% of the source primitives from the SSIM term.
    The first row is our full distribution.
    Shading as in Table~\ref{tab:ablation}.
}
\label{tab:ablation_sampling}
\end{table}

\begin{table}[]
    \centering
    \small
    \setlength{\tabcolsep}{4pt}
    \begin{tabular}{lrrr}
    \toprule
        Method & Mem. [GB] & Init.  [min] & Train  [min] \\%
        \midrule
        \multicolumn{4}{l}{\qquad\emph{\textcolor{gray}{fixed-budget (@500k)}}} \\
        EDGS & 7.3 & 1.8 & 7.2\\%
        ConeGS & 4.9 & 2.2 & 11.8 \\%
        Ours & \textbf{4.6} & -- & \textbf{2.8}  \\ %
        \midrule
        \multicolumn{4}{l}{\qquad\emph{\textcolor{gray}{variable-budget}}} \\
        Mini-Splatting2\textsuperscript{$\dagger$} & 8.2 & -- & \textbf{1.9} \\ %
        GaussianSpa & 11.3 & -- & 24.6 \\ %
        Ours & \textbf{4.8} & -- & 3.0 \\ %
         \bottomrule
    \end{tabular} 
    
    {\scriptsize \textsuperscript{$\dagger$} Mini-Splatting2 only trains for 18k iterations.}
    \caption{
\textbf{Memory and Runtime} during optimization, as means over the 9 Mip-NeRF 360 scenes on a single NVIDIA RTX 5090.
Peak memory (Mem.) includes initialization (Init.), \ie the dense matching of EDGS and the iNGP pre-training of ConeGS.
Train is the optimization time w/o initialization.
Bold marks the best value in each block.
    }
    \label{tab:memory}
\end{table}

\paragraph{Sampling Distribution.}
With a single exception, all of our tested error signals beat opacity-proportional sampling within our method at $\mathrm{500k}$ primitives (Table~\ref{tab:ablation_sampling}) --- though for PSNR, only our full method outperforms it.
Confidence weighting and the SSIM term each improve gradient-only sampling, and their respective gains compound.
With the SSIM term, background detail is reconstructed more faithfully (Figure~\ref{fig:ssim_sampling_analysis}).
In SSIM and LPIPS, the gradient-based variants are close, and confidence weighting pays off mainly for the SSIM term alone, whose LPIPS it improves by 0.01 at equal PSNR.

\def\cmpZoomGardenSSIM{0.20,0.85}
\def\cmpZoomBicycleSSIM{0.10,0.60}

\newcommand{\cmpImageSSIM}[5][]{%
    \begin{tikzpicture}[baseline=(image.south), spy using outlines={rectangle, draw=red, 
        line width=0.5pt, magnification=3, width=0.25\linewidth, height=0.147\linewidth}]
        
        \IfFileExists{figures/ssim_comparison/#2_#3.jpg}{%
            \node[anchor=south west, inner sep=0] (image) at (0,0) {%
                \includegraphics[width=0.494\linewidth]{figures/ssim_comparison/#2_#3.jpg}%
            };
        }{%
            \node[anchor=south west, inner sep=0, fill=black!8, draw=black!25,
                minimum width=0.49\linewidth, minimum height=0.35\linewidth] (image) at (0,0) {};
            \node[anchor=center, font=\ttfamily\tiny, text=gray] at (image.center) {#2\_#3};
        }
        
        \begin{scope}[x={(image.south east)}, y={(image.north west)}]
            \ifx\relax#1\relax\else
                \node[anchor=south west, 
                      fill=black, fill opacity=0.45, 
                      inner sep=2.5pt, 
                      text opacity=0.8,
                      rounded corners=1.5pt,
                      align=left] 
                at (0.02, 0.02) {%
                    {\sffamily\scriptsize\textbf{\textcolor{white}{#1}}}%
                };
            \fi
            
            \coordinate (zoomcenter) at (#4);
        \end{scope}
        
        \spy on (zoomcenter) in node[anchor=south east] at (image.south east);
    \end{tikzpicture}%
}

\begin{figure}[t]
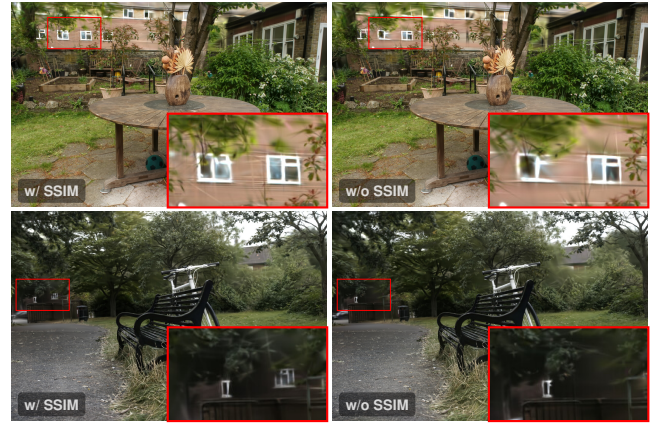

    \centering
    
    \cmpImageSSIM[w/ SSIM]{garden}{ssim}{\cmpZoomGardenSSIM}{}\hfill
    \cmpImageSSIM[w/o SSIM]{garden}{no\_ssim}{\cmpZoomGardenSSIM}{}
    \cmpImageSSIM[w/ SSIM]{bicycle}{ssim}{\cmpZoomBicycleSSIM}{}\hfill
    \cmpImageSSIM[w/o SSIM]{bicycle}{no\_ssim}{\cmpZoomBicycleSSIM}{}
    
    \caption{
\textbf{Effect of the SSIM Term}: Our method on Garden and Bicycle with and without drawing 20\% of the source primitives from the SSIM term (Section~\ref{sec:method:sampling}). 
Including the SSIM term leads to more detailed background reconstruction.
}
    \label{fig:ssim_sampling_analysis}
\end{figure}

\paragraph{Matched Initialization.}
Our densification can also start from a prior: we run it from the dense matches of EDGS~\cite{Kotovenko2026edgs} and the iNGP initialization of ConeGS~\cite{baranowski2026conegs} on all 24 scenes (Table~\ref{tab:init_comparison}).
At $\mathrm{100k}$, either initialization brings our method on par with ConeGS in PSNR and SSIM, while ConeGS keeps the better LPIPS, and at $\mathrm{500k}$, our method is best in all three metrics.
From both initializations, our results are nearly identical and better than from SfM points, whereas EDGS and ConeGS, which also differ in their densification, are clearly apart.
A good initialization thus helps our method, and the gain hardly depends on which prior provides it.

\paragraph{Random Initialization.}
Our method also keeps the robustness of 3DGS-MCMC~\cite{kheradmand2024mcmc} to random initialization.
On Mip-NeRF 360 at $\mathrm{100k}$, it reaches 1.9\,dB higher PSNR than 3DGS-MCMC from random points (supplementary Table~\ref{tab:results-random}), and adding the noise of 3DGS-MCMC removes most of this lead.
At the primitive counts of the original 3DGS models, ranging from $\mathrm{1.2M}$ to $\mathrm{6.1M}$, both methods reach the same PSNR from random points, but ours has improved LPIPS metrics (0.188 against 0.210, supplementary Table~\ref{tab:results-random}).
Exploration from random points thus does not need to be driven by noise, and adding noise can even hurt.

\paragraph{Memory and Runtime.}
Without an initialization stage, our method is faster than both prior-based methods and needs less peak memory (Table~\ref{tab:memory}).
As measured at $\mathrm{500k}$, ConeGS and EDGS take about $5\times$ and $3\times$ as long, and part of this gap comes from implementation, since the baselines run in their official codebases and ours uses the Faster-GS rasterizer and optimizer~\cite{hahlbohm2026fastergs}.
The initialization stage alone shows that the gap is not only an implementation effect: for ConeGS, it takes 2.2 minutes, nearly as long as our entire run of 2.8 minutes; hence, even with a training stage as fast as ours, ConeGS would need almost twice as long.
Against the simplification methods, which pass through a much larger set of primitives, it saves even more memory (Section~\ref{sec:results}).

\begin{figure}
    \centering
    \includegraphics[width=0.995\linewidth]{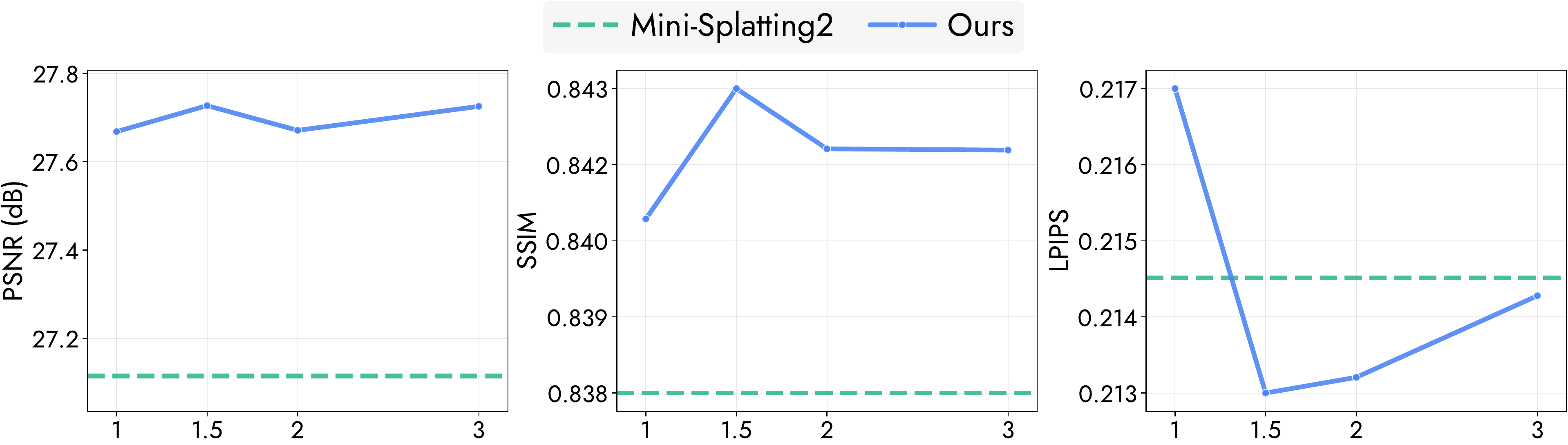}
    \includegraphics[width=0.995\linewidth]{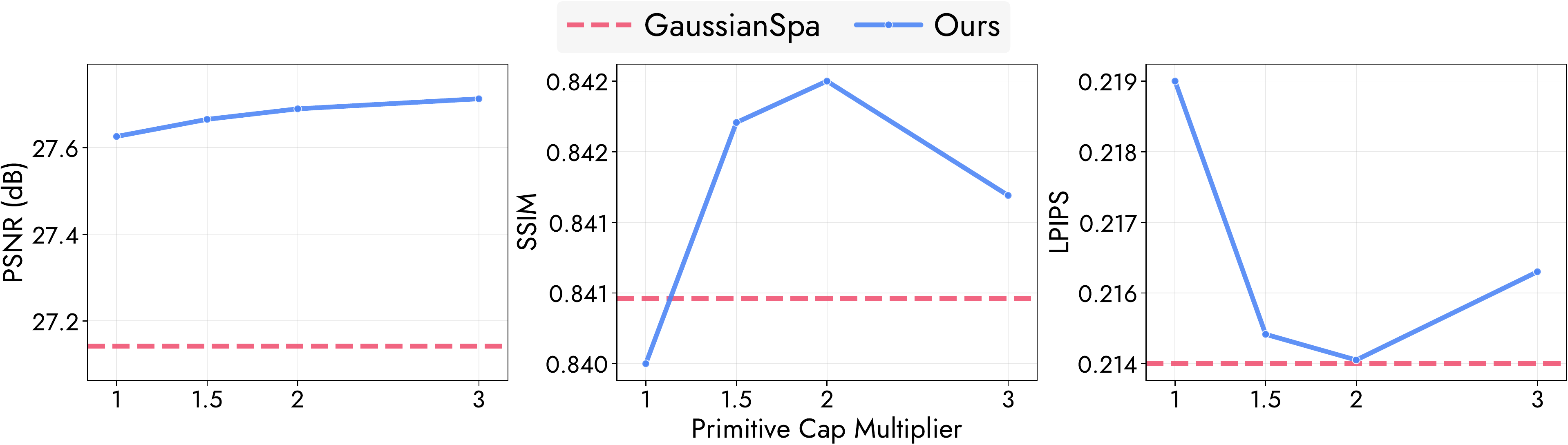}
    \caption{\textbf{Growth Beyond the Budget}:
PSNR, SSIM and LPIPS of our method when it may grow to $\lambda_\text{grow}$ times the budget during training, as means over the 13 Mip-NeRF 360, Tanks \& Temples and Deep Blending scenes of Table~\ref{tab:fixed_prim}.
In each row, the budget is the per-scene primitive count of the simplification method shown dashed, Mini-Splatting2 (top) or GaussianSpa (bottom), so $\lambda_\text{grow} = 1$ repeats the corresponding block of Table~\ref{tab:fixed_prim}.
At $\lambda_\text{grow}$ of 1.5 or 2, our LPIPS is on par with both methods.
}
    \label{fig:primitive_cap}
\end{figure}

\paragraph{Growth Beyond the Budget.}
The LPIPS advantage of the simplification methods in Table~\ref{tab:fixed_prim} most likely comes from their large intermediate set of primitives.
To test this, we let our method grow to $\lambda_\text{grow}$ times the budget by half of training and then reduce the count to the budget by iteration 25k, removing the primitives with the lowest $w_i$ (Figure~\ref{fig:primitive_cap}).
At $\lambda_\text{grow}$ of 1.5 or 2, our LPIPS is on par with both methods, and our PSNR stays 0.5 to 0.6\,dB ahead at every $\lambda_\text{grow}$.
The larger intermediate set thus captures the fine detail that is covered late from SfM points (Section~\ref{sec:results}), and the reduction keeps the primitives that carry it.
At $\lambda_\text{grow} = 3$, LPIPS gets worse again, most likely because the maximum in $w_i$ favors small primitives that dominate a few pixels but contribute little to the image as a whole.
Mini-Splatting2 and GaussianSpa pass through far larger intermediate sets, whereas 1.5 to 2 times the budget suffices for our method.
Since bounding peak memory is one of our goals, we keep $\lambda_\text{grow} = 1$ by default, but where memory allows, we recommend $\lambda_\text{grow}$ of 1.5 to 2 for the best perceptual quality.

\begin{table}
\centering
\small
\setlength{\tabcolsep}{4pt}
\begin{tabular}{llcccccc}
\toprule
 & & \multicolumn{3}{c}{100k} & \multicolumn{3}{c}{500k} \\
\cmidrule(lr){3-5} \cmidrule(lr){6-8}
Method & Init. & PSNR & SSIM & LPIPS & PSNR & SSIM & LPIPS \\
\midrule
Ours & SfM & 26.54 & 0.809 & 0.282 & 28.05 & 0.863 & 0.202 \\
\midrule
Ours & EDGS & \textbf{26.72} & \textbf{0.813} & \textbf{0.277} & \textbf{28.20} & \textbf{0.867} & \textbf{0.195} \\
EDGS & EDGS & 25.93 & 0.799 & 0.288 & 27.51 & 0.855 & 0.202 \\
\midrule
Ours & iNGP & \textbf{26.79} & \textbf{0.813} & 0.278 & \textbf{28.22} & \textbf{0.866} & \textbf{0.197} \\
ConeGS & iNGP & 26.75 & \textbf{0.813} & \textbf{0.273} & 27.87 & 0.859 & 0.199 \\
\bottomrule
\end{tabular}
\caption{\textbf{Matched Initialization}:
Our method started from the dense initialization of EDGS~\cite{Kotovenko2026edgs} or ConeGS~\cite{baranowski2026conegs}, compared to the respective prior-based method and to our method with SfM initialization; means over all 24 scenes.
Bold marks the better value of each pair of rows with shared initialization.
}
\label{tab:init_comparison}
\end{table}

\section{Limitations and Future Work}
\label{sec:limitations}
At the lowest budget, our densification cannot fully make up for a better initialization.
From SfM points, ConeGS stays ahead of our method at $\mathrm{100k}$, and even when we start from its initialization, ConeGS keeps the better LPIPS (Table~\ref{tab:init_comparison}).
A better starting point thus still pays off under a strict budget.
It remains open whether such a start can be obtained more cheaply than by fitting Gaussians directly: the iNGP initialization of ConeGS alone takes nearly as long as our complete training at $\mathrm{500k}$ (Table~\ref{tab:memory}).

Without a dense initialization, recovering fine detail under a strict budget remains challenging: outdoors, methods that simplify a denser reconstruction reach the better LPIPS at equal budget (Table~\ref{tab:fixed_prim}).
Growth beyond the budget closes this gap at the cost of peak memory, but only up to about twice the budget, since the maximum blending weight keeps primitives that matter for single pixels (Section~\ref{sec:analysis}).
Richer appearance models~\cite{condor2026nht, liu2025beta, tojo2026diffsoup} could recover this detail at a lower peak count.

Our schedules for growth and relocation are set by hand and shared by all budgets.
Relocating a larger share of primitives slightly improves PSNR at $\mathrm{1M}$ and $\mathrm{2M}$, but not at $\mathrm{100k}$ (supplementary Table~\ref{tab:explore}), so the relocation share could adapt to the budget, which is also user-defined.
Since the confidence marks regions that the reconstruction still explains poorly, it could indicate when further primitives stop paying off and thus set the budget of each scene.

\section{Conclusion}
Error-driven densification belongs in the fixed-budget framework of 3DGS-MCMC.
We keep what makes 3DGS-MCMC robust, its fixed budget and gradual growth, and replace the two decisions it bases on opacity: the image error, weighted by a learned confidence, decides where new primitives go, and their contribution to the rendered images decides which primitives make room for them.
This simple change improves on 3DGS-MCMC at every budget we test.
Without any prior, our method produces the most accurate reconstructions of all fixed-budget methods from $\mathrm{500k}$ primitives on, ahead of methods that rely on a pre-trained radiance field or dense image matches, and it outperforms 3DGS, DashGaussian and 3DGS + GOF at their own primitive counts.

The same densification, with one configuration for all scenes and budgets, adapts to its starting point and goal.
It trains from random points as robustly as 3DGS-MCMC, and from $\mathrm{500k}$ on, it turns the initialization of a prior-based method into a better reconstruction than that method itself.
Allowed to grow beyond the budget, it matches the perceptual quality of the simplification methods from a much smaller intermediate set and remains more accurate.
It needs no pre-trained network, radiance field or image matcher, only one confidence per primitive and one loss term, and on the optimized Faster-GS pipeline~\cite{hahlbohm2026fastergs}, it trains a scene with $\mathrm{500k}$ primitives in under three minutes.
We therefore see it as a replacement for the densification of 3DGS-MCMC wherever the number of primitives is bounded.

\bibliographystyle{eg-alpha-doi} 
\bibliography{egbibsample}       

\setcounter{section}{0}
\setcounter{figure}{0}
\setcounter{table}{0}
\setcounter{equation}{0}

\renewcommand{\thesection}{S.\arabic{section}}
\renewcommand{\thefigure}{S.\arabic{figure}}
\renewcommand{\thetable}{S.\arabic{table}}
\renewcommand{\theequation}{S.\arabic{equation}}

\newpage 
\clearpage

\section{Supplementary Material}
The supplementary material contains additional details and experiments, which we omitted from the main material for brevity.

\subsection{Per-Scene Results}

\paragraph{Fixed Budget Results.}
We provide per-scene image quality metrics for all fixed-budget methods (Ours, ConeGS~\cite{baranowski2026conegs}, EDGS~\cite{Kotovenko2026edgs}, Taming 3DGS~\cite{mallick2024taming}, 3DGS-MCMC~\cite{kheradmand2024mcmc}) in Tables~\ref{tab:results-100000}--\ref{tab:results-2000000}.
As previously described in the main material, we attain state-of-the-art results starting from a primitive count of $\mathrm{500k}$ primitives;
at $\mathrm{100k}$ primitives, the prior-based method ConeGS attains better results, which we attribute to initialization.

\paragraph{Random Initialization.}
We provide per-scene results with random initialization in Table~\ref{tab:results-random}, where we report results at both $\mathrm{100k}$ primitives, as well as the setting of 3DGS-MCMC~\cite{kheradmand2024mcmc};
each scene follows the primitive count from 3DGS~\cite{kerbl20233dgs}.
Our method attains the strongest results across all tested scenes for $\mathrm{100k}$ primitives, while attaining better perceptual quality than 3DGS-MCMC at high primitive counts.

\paragraph{Matched Initialization.}
In Tables~\ref{tab:results-edgsinit-100000}--\ref{tab:conegs-comparison-500k}, we provide per-scene results for our evaluation with matched initialization (reported in Table~\ref{tab:init_comparison} in the main material).
Here, we compared against both ConeGS~\cite{baranowski2026conegs} and EDGS~\cite{Kotovenko2026edgs}.
Our method benefits from both of these initialization types, and achieves strong results across all tested scenes.

\subsection{Additional Results and Details}

\paragraph{Comparison with Revising-3DGS.}
We also compared our method to the method of Rota Bul\`o~\emph{et al.}~\cite{bulo2024revising} in Table~\ref{tab:revising_comparison}, which is also a controllable fixed-budget method.
We omitted these results from the main paper for brevity.
Note that the authors did not release their source code; 
hence, we re-implemented their approach following the details given in the original manuscript.

\paragraph{Qualitative Comparison with Variable-Budget Methods.}
We provide qualitative comparisons with our tested variable-budget methods (3DGS~\cite{kerbl20233dgs}, 3DGS w/ GOF densification~\cite{yu2024gof, ye2024absgs}, DashGaussian~\cite{chen2025dashgaussian}, Mini-Splatting2~\cite{Fang2024minisplatting2}, GaussianSpa~\cite{zhang2025gaussianspa});
see Figure~\ref{fig:supp-variable-budget}.

\paragraph{Relocation Share.}
Across all primitive budgets, we also evaluate the impact of our chosen primitive share, \ie the maximum percentage of low-contribution primitives to relocate --- increasing this value leads to more exploration.
We evaluated this for all 9 scenes from the Mip-NeRF 360 dataset~\cite{barron2022mipnerf360}, with relocation shares from 5\% up to 15\%.
Modifying this value has a minor impact on performance, with large primitive counts in particular benefiting from higher values.
Conversely, for lower primitive counts (\ie $\mathrm{100k}$) the image quality metrics decrease slightly for higher relocation shares --- ultimately, our chosen values of 10\% attains state-of-the-art results against all tested comparison methods, demonstrating its effectiveness. 

\paragraph{Background Reconstruction for Indoor Scenes.}
Complementary to Table~\ref{tab:background} in the main manuscript, we report foreground and background reconstruction metrics for the indoor scenes of Mip-NeRF 360~\cite{barron2022mipnerf360} in Table~\ref{tab:background-indoor}.
Our improvements in PSNR are similarly strong, whereas our SSIM/LPIPS improvements are less pronounced.
For the tested dataset, the indoor backgrounds are easier to reconstruct as they lack intricate details such as fine foliage, which induces an upper bound on the quantitative deltas for SSIM/LPIPS.
Still, we attain improved results for all tested metrics/primitive counts.

\paragraph{On Reproducibility.}
For the fixed-budget comparisons, we strictly follow the setting of ConeGS~\cite{baranowski2026conegs}; 
if the primitive count of the initialization exceeds the budget, we retain a correctly-sized set for initialization via subsampling.
For EDGS~\cite{Kotovenko2026edgs}, we followed this exact procedure, and train with pruning enabled; we replaced the pruned primitives with new primitives via the standard densification in the codebase, which ensures a fair comparisons between methods.

\begin{table*}
\centering
\resizebox{0.87\linewidth}{!}{%
\begin{tabular}{cl|ccc|ccc|ccc|ccc|ccc}
\toprule
  & & \multicolumn{3}{c}{Ours} & \multicolumn{3}{c}{ConeGS} & \multicolumn{3}{c}{EDGS} & \multicolumn{3}{c}{Taming 3DGS} & \multicolumn{3}{c}{3DGS-MCMC} \\
\cmidrule(lr){3-5} \cmidrule(lr){6-8} \cmidrule(lr){9-11} \cmidrule(lr){12-14} \cmidrule(lr){15-17}
  & \textit{Scene} & PSNR & SSIM & LPIPS & PSNR & SSIM & LPIPS & PSNR & SSIM & LPIPS & PSNR & SSIM & LPIPS & PSNR & SSIM & LPIPS \\
\midrule
 \multirow{10}{*}{\rotatebox{90}{\parbox{1.15cm}{\centering\scriptsize Mip-NeRF360}}} & Bicycle & \cellcolor{tabsecond}23.82 & \cellcolor{tabthird}0.636 & \cellcolor{tabthird}0.396 & \cellcolor{tabfirst}24.08 & \cellcolor{tabfirst}0.655 & \cellcolor{tabfirst}0.378 & \cellcolor{tabthird}23.71 & \cellcolor{tabsecond}0.636 & \cellcolor{tabsecond}0.392 & 23.30 & 0.588 & 0.436 & 23.34 & 0.603 & 0.421 \\
  & Flowers & \cellcolor{tabsecond}20.48 & \cellcolor{tabsecond}0.503 & \cellcolor{tabthird}0.471 & \cellcolor{tabfirst}20.59 & \cellcolor{tabfirst}0.509 & \cellcolor{tabfirst}0.465 & 20.12 & \cellcolor{tabthird}0.496 & \cellcolor{tabsecond}0.467 & 19.98 & 0.462 & 0.507 & \cellcolor{tabthird}20.15 & 0.474 & 0.491 \\
  & Garden & \cellcolor{tabfirst}24.89 & \cellcolor{tabthird}0.703 & \cellcolor{tabthird}0.357 & \cellcolor{tabsecond}24.68 & \cellcolor{tabfirst}0.709 & \cellcolor{tabfirst}0.336 & \cellcolor{tabthird}24.45 & \cellcolor{tabsecond}0.705 & \cellcolor{tabsecond}0.341 & 24.20 & 0.665 & 0.383 & 24.37 & 0.682 & 0.366 \\
  & Stump & \cellcolor{tabsecond}25.68 & \cellcolor{tabsecond}0.706 & \cellcolor{tabsecond}0.354 & \cellcolor{tabfirst}25.79 & \cellcolor{tabfirst}0.709 & \cellcolor{tabfirst}0.349 & \cellcolor{tabthird}25.13 & \cellcolor{tabthird}0.673 & \cellcolor{tabthird}0.377 & 24.19 & 0.626 & 0.423 & 24.78 & 0.657 & 0.394 \\
  & Treehill & \cellcolor{tabsecond}22.60 & \cellcolor{tabsecond}0.577 & \cellcolor{tabsecond}0.474 & \cellcolor{tabfirst}22.79 & \cellcolor{tabthird}0.569 & \cellcolor{tabthird}0.476 & 22.16 & \cellcolor{tabfirst}0.591 & \cellcolor{tabfirst}0.449 & \cellcolor{tabthird}22.28 & 0.560 & 0.494 & 22.28 & 0.563 & 0.486 \\
  & Room & \cellcolor{tabfirst}31.06 & \cellcolor{tabfirst}0.904 & \cellcolor{tabsecond}0.260 & \cellcolor{tabsecond}30.95 & \cellcolor{tabsecond}0.903 & \cellcolor{tabfirst}0.254 & 29.89 & 0.894 & \cellcolor{tabthird}0.267 & 30.03 & 0.888& 0.292 & \cellcolor{tabthird}30.54 & \cellcolor{tabthird}0.896 & 0.275 \\
  & Counter & \cellcolor{tabfirst}28.13 & \cellcolor{tabfirst}0.884 & \cellcolor{tabsecond}0.252 & \cellcolor{tabsecond}28.11 & \cellcolor{tabsecond}0.879 & \cellcolor{tabfirst}0.251 & 27.78 & \cellcolor{tabthird}0.876 & \cellcolor{tabthird}0.253 & \cellcolor{tabthird}27.89 & 0.870 & 0.271 & 27.77 & 0.874 & 0.266 \\
  & Kitchen & \cellcolor{tabsecond}29.73 & \cellcolor{tabsecond}0.893 & \cellcolor{tabthird}0.203 & \cellcolor{tabfirst}29.82 & \cellcolor{tabfirst}0.897 & \cellcolor{tabfirst}0.183 & \cellcolor{tabthird}28.92 & \cellcolor{tabthird}0.890 &\cellcolor{tabsecond}0.189 & 28.52 & 0.880 & 0.213 & 28.76 & 0.884 & 0.209 \\
  & Bonsai & \cellcolor{tabsecond}30.50 & \cellcolor{tabsecond}0.918 & \cellcolor{tabsecond}0.256 & \cellcolor{tabfirst}30.63 & \cellcolor{tabfirst}0.920 & \cellcolor{tabfirst}0.241 & 29.63 & \cellcolor{tabthird}0.905 & \cellcolor{tabthird}0.258 & 29.71 & 0.902 & 0.280 & \cellcolor{tabthird}29.76 & 0.903 & 0.276 \\
 & \textbf{Average} & \cellcolor{tabsecond}26.32 & \cellcolor{tabsecond}0.748 &\cellcolor{tabthird}0.336 & \cellcolor{tabfirst}26.38 & \cellcolor{tabfirst}0.750 & \cellcolor{tabfirst}0.326 & \cellcolor{tabthird}25.75 & \cellcolor{tabthird}0.741 & \cellcolor{tabsecond}0.333 & 25.57 & 0.715 & 0.367 & 25.75 & 0.726 & 0.354 \\
 \midrule
 \multirow{3}{*}{\rotatebox{90}{\parbox{1.15cm}{\centering\scriptsize Tanks \& Temples}}} & Truck & \cellcolor{tabsecond}24.74 & \cellcolor{tabsecond}0.848 & \cellcolor{tabthird}0.223 & \cellcolor{tabfirst}24.88 & \cellcolor{tabfirst}0.850& \cellcolor{tabfirst}0.201 & 23.80 & \cellcolor{tabthird}0.830 & \cellcolor{tabsecond}0.222 & \cellcolor{tabthird}24.02 & 0.825 & 0.246 & 22.98 & 0.805 & 0.261 \\
  & Train & \cellcolor{tabfirst}22.19 & \cellcolor{tabfirst}0.781 & \cellcolor{tabsecond}0.279 & \cellcolor{tabsecond}21.85 & \cellcolor{tabthird}0.772 & \cellcolor{tabfirst}0.279 & \cellcolor{tabthird}21.59 & \cellcolor{tabsecond}0.774 & \cellcolor{tabthird}0.285 & 21.43 & 0.745 & 0.327 & 21.50 & 0.759 & 0.304 \\
 & \textbf{Average} & \cellcolor{tabfirst}23.47 & \cellcolor{tabfirst}0.815 & \cellcolor{tabsecond}0.251 & \cellcolor{tabsecond}23.36 & \cellcolor{tabsecond}0.811 & \cellcolor{tabfirst}0.240 & 22.69 & \cellcolor{tabthird}0.802 & \cellcolor{tabthird}0.254 & \cellcolor{tabthird}22.73 & 0.785 & 0.286 & 22.24 & 0.782 & 0.282 \\
 \midrule
 \multirow{3}{*}{\rotatebox{90}{\parbox{1.15cm}{\centering\scriptsize Deep Blending}}} & Dr Johnson & \cellcolor{tabsecond}29.05 & \cellcolor{tabfirst}0.893 & \cellcolor{tabfirst}0.290 & \cellcolor{tabfirst}29.35 & \cellcolor{tabsecond}0.891 & \cellcolor{tabsecond}0.293 & 28.49 & \cellcolor{tabthird}0.885 & \cellcolor{tabthird}0.299 & 28.33 & 0.876 & 0.323 & \cellcolor{tabthird}28.68 & 0.883 & 0.311 \\
  & Playroom & \cellcolor{tabsecond}30.05 & \cellcolor{tabfirst}0.905 & \cellcolor{tabfirst}0.276 & \cellcolor{tabfirst}30.26 & \cellcolor{tabsecond}0.901 & \cellcolor{tabsecond}0.278 & 29.49 & \cellcolor{tabthird}0.894 & \cellcolor{tabthird}0.291 & \cellcolor{tabthird}29.55 & 0.893 & 0.296 & 29.05 & 0.891 & 0.305 \\
 & \textbf{Average} & \cellcolor{tabsecond}29.55 & \cellcolor{tabfirst}0.899 & \cellcolor{tabfirst}0.283 & \cellcolor{tabfirst}29.81 & \cellcolor{tabsecond}0.896 & \cellcolor{tabsecond}0.286 & \cellcolor{tabthird}28.99 & \cellcolor{tabthird}0.890 & \cellcolor{tabthird}0.295 & 28.94 & 0.884 & 0.310 & 28.86 & 0.887 & 0.308 \\
 \midrule
 \multirow{12}{*}{\rotatebox{90}{\parbox{1.15cm}{\centering\scriptsize DL3DV}}}& 140 & \cellcolor{tabsecond}26.22 & \cellcolor{tabsecond}0.859 & \cellcolor{tabsecond}0.253 & \cellcolor{tabfirst}26.60 & \cellcolor{tabfirst}0.867 & \cellcolor{tabfirst}0.239 & 25.93 & \cellcolor{tabthird}0.846 & \cellcolor{tabthird}0.267 & \cellcolor{tabthird}25.95 & 0.836 & 0.284 & 25.64 & 0.837 & 0.284 \\
  & Car & \cellcolor{tabsecond}29.28 & \cellcolor{tabfirst}0.934 & \cellcolor{tabfirst}0.203 & \cellcolor{tabthird}28.99 & \cellcolor{tabthird}0.931 & \cellcolor{tabsecond}0.208 & 28.77 & 0.928 & \cellcolor{tabthird}0.209 & \cellcolor{tabfirst}29.37 & \cellcolor{tabsecond}0.932 & 0.211 & 28.80 & 0.928 & 0.218 \\
  & Garden & \cellcolor{tabthird}22.78 & \cellcolor{tabthird}0.652 & \cellcolor{tabthird}0.360 & \cellcolor{tabfirst}23.47 & \cellcolor{tabfirst}0.682 & \cellcolor{tabfirst}0.333 & \cellcolor{tabsecond}23.04 & \cellcolor{tabsecond}0.662 & \cellcolor{tabsecond}0.346 & 22.37 & 0.598 & 0.401 & 21.91 & 0.595 & 0.408 \\
  & Greenhouse & \cellcolor{tabsecond}20.06 & \cellcolor{tabsecond}0.623 & \cellcolor{tabsecond}0.381 & \cellcolor{tabfirst}20.39 & \cellcolor{tabfirst}0.635 &\cellcolor{tabfirst}0.373 & \cellcolor{tabthird}19.33 & \cellcolor{tabthird}0.580 & \cellcolor{tabthird}0.418 & 19.27 & 0.557 & 0.445 & 18.91 & 0.560 & 0.433 \\
  & Grills & \cellcolor{tabsecond}25.68 & \cellcolor{tabsecond}0.891 & \cellcolor{tabsecond}0.270 & \cellcolor{tabfirst}25.84 & \cellcolor{tabfirst}0.898 & \cellcolor{tabfirst}0.259 & 24.72 & \cellcolor{tabthird}0.876 & \cellcolor{tabthird}0.283 & \cellcolor{tabthird}25.32 & 0.875 & 0.298 & 24.77 & 0.871 & 0.298 \\
  & Herbary & \cellcolor{tabsecond}30.41 & \cellcolor{tabsecond}0.935 & \cellcolor{tabsecond}0.144 & \cellcolor{tabfirst}30.69 & \cellcolor{tabfirst}0.936 & \cellcolor{tabfirst}0.138 & 29.35 & 0.919 & \cellcolor{tabthird}0.159 & \cellcolor{tabthird}29.91 & \cellcolor{tabthird}0.922 & 0.161 & 29.50 & 0.920 & 0.167 \\
  & Highrise & \cellcolor{tabsecond}25.24 & \cellcolor{tabsecond}0.848 & \cellcolor{tabsecond}0.220 & \cellcolor{tabfirst}25.80 & \cellcolor{tabfirst}0.864 & \cellcolor{tabfirst}0.203 & \cellcolor{tabthird}24.83 & \cellcolor{tabthird}0.842& \cellcolor{tabthird}0.231 & 23.99 & 0.802 & 0.280 & 24.05 & 0.810 & 0.268 \\
  & Roomset & \cellcolor{tabfirst}30.67 & \cellcolor{tabsecond}0.936 & \cellcolor{tabsecond}0.108 & \cellcolor{tabsecond}30.60 & \cellcolor{tabfirst}0.937 & \cellcolor{tabfirst}0.106 & 29.99 & 0.926 & 0.127 & \cellcolor{tabthird}30.59 & \cellcolor{tabthird}0.929 & \cellcolor{tabthird}0.123 & 30.07 & 0.926 & 0.126 \\
  & Statue & \cellcolor{tabsecond}29.52 & \cellcolor{tabsecond}0.909 & \cellcolor{tabsecond}0.203 & \cellcolor{tabfirst}30.81 & \cellcolor{tabfirst}0.919 & \cellcolor{tabfirst}0.191 & 28.46 & \cellcolor{tabthird}0.899 & \cellcolor{tabthird}0.209 & \cellcolor{tabthird}28.94 & 0.889 & 0.235 & 28.06 & 0.882 & 0.242 \\
  & Supermarket & \cellcolor{tabsecond}30.17 & \cellcolor{tabsecond}0.897 & \cellcolor{tabsecond}0.244 & \cellcolor{tabfirst}30.45 & \cellcolor{tabfirst}0.900 & \cellcolor{tabfirst}0.234 & \cellcolor{tabthird}29.11 & \cellcolor{tabthird}0.884 & \cellcolor{tabthird}0.255 & 28.96 & 0.872 & 0.275 & 28.59 & 0.873 & 0.272 \\
  & Vasary & \cellcolor{tabsecond}24.13 & \cellcolor{tabsecond}0.779 & \cellcolor{tabsecond}0.296 & \cellcolor{tabfirst}24.60 & \cellcolor{tabfirst}0.786 & \cellcolor{tabfirst}0.279 & \cellcolor{tabthird}23.60 & \cellcolor{tabthird}0.759 &\cellcolor{tabthird}0.314 & 23.36 & 0.741 & 0.345 & 22.62 & 0.723 & 0.353 \\
 & \textbf{Average} & \cellcolor{tabsecond}26.74 & \cellcolor{tabsecond}0.842 &\cellcolor{tabsecond}0.244 & \cellcolor{tabfirst}27.11 & \cellcolor{tabfirst}0.850 & \cellcolor{tabfirst}0.233 & 26.10 & \cellcolor{tabthird}0.829 & \cellcolor{tabthird}0.256 & \cellcolor{tabthird}26.18 & 0.814 & 0.278 & 25.72 & 0.811 & 0.279 \\
\bottomrule
\end{tabular}}
\caption{Per-scene results at a budget of $\mathrm{100k}$ primitives, SfM initialization.}
\label{tab:results-100000}
\end{table*}
\begin{table*}
\centering
\resizebox{0.87\linewidth}{!}{%
\begin{tabular}{cl|ccc|ccc|ccc|ccc|ccc}
\toprule
  & & \multicolumn{3}{c}{Ours} & \multicolumn{3}{c}{ConeGS} & \multicolumn{3}{c}{EDGS} & \multicolumn{3}{c}{Taming 3DGS} & \multicolumn{3}{c}{3DGS-MCMC} \\
\cmidrule(lr){3-5} \cmidrule(lr){6-8} \cmidrule(lr){9-11} \cmidrule(lr){12-14} \cmidrule(lr){15-17}
  & \textit{Scene} & PSNR & SSIM & LPIPS & PSNR & SSIM & LPIPS & PSNR & SSIM & LPIPS & PSNR & SSIM & LPIPS & PSNR & SSIM & LPIPS \\
\midrule
 \multirow{10}{*}{\rotatebox{90}{\parbox{1.15cm}{\centering\scriptsize Mip-NeRF360}}} & Bicycle & \cellcolor{tabsecond}25.19 & \cellcolor{tabsecond}0.753 & \cellcolor{tabthird}0.261 & \cellcolor{tabfirst}25.26 & \cellcolor{tabfirst}0.758 &\cellcolor{tabfirst}0.242 & \cellcolor{tabthird}24.97 & \cellcolor{tabthird}0.752 & \cellcolor{tabsecond}0.251 & 24.53 & 0.685 & 0.339 & 24.74 & 0.715 & 0.307 \\
  & Flowers & \cellcolor{tabfirst}21.91 & \cellcolor{tabfirst}0.610 & \cellcolor{tabthird}0.360 & \cellcolor{tabsecond}21.46 & \cellcolor{tabsecond}0.604 & \cellcolor{tabfirst}0.347 & 21.08 & \cellcolor{tabthird}0.598 & \cellcolor{tabsecond}0.351 & 21.14 & 0.560 & 0.406 & \cellcolor{tabthird}21.41 & 0.570 & 0.393 \\
  & Garden & \cellcolor{tabfirst}27.13 & \cellcolor{tabfirst}0.842 & \cellcolor{tabthird}0.171 & \cellcolor{tabsecond}26.86 & \cellcolor{tabsecond}0.837 & \cellcolor{tabfirst}0.157 & 26.45 & \cellcolor{tabthird}0.831 & \cellcolor{tabsecond}0.164 & \cellcolor{tabthird}26.54 & 0.812 & 0.206 & 26.39 & 0.816 & 0.195 \\
  & Stump & \cellcolor{tabfirst}27.15 & \cellcolor{tabfirst}0.797 & \cellcolor{tabsecond}0.220 & \cellcolor{tabsecond}26.96 & \cellcolor{tabsecond}0.794 & \cellcolor{tabfirst}0.214 & 26.52 & \cellcolor{tabthird}0.769 & \cellcolor{tabthird}0.238 & 26.07 & 0.731 & 0.299 & \cellcolor{tabthird}26.53 & 0.754 & 0.273 \\
  & Treehill & \cellcolor{tabfirst}23.09 & \cellcolor{tabsecond}0.644 & \cellcolor{tabsecond}0.360 & 22.79 & 0.608 & \cellcolor{tabthird}0.372 & 22.57 & \cellcolor{tabfirst}0.653 & \cellcolor{tabfirst}0.319 & \cellcolor{tabsecond}22.98 & 0.615 & 0.406 & \cellcolor{tabthird}22.80 & \cellcolor{tabthird}0.622 & 0.394 \\
  & Room & \cellcolor{tabfirst}32.30 & \cellcolor{tabfirst}0.928 & \cellcolor{tabfirst}0.200 & \cellcolor{tabsecond}31.95 & \cellcolor{tabsecond}0.923 & \cellcolor{tabthird}0.206 & 31.48 & \cellcolor{tabthird}0.923 & \cellcolor{tabsecond}0.205 & 31.78 & 0.914 & 0.234 & \cellcolor{tabthird}31.91 & 0.922 & 0.219 \\
  & Counter & \cellcolor{tabfirst}29.35 & \cellcolor{tabfirst}0.916 & \cellcolor{tabsecond}0.187 & 28.93 & 0.906 & \cellcolor{tabthird}0.194 & \cellcolor{tabthird}29.04 & \cellcolor{tabsecond}0.911 & \cellcolor{tabfirst}0.185 & \cellcolor{tabsecond}29.23 & 0.904 & 0.210 & 28.99 & \cellcolor{tabthird}0.908 & 0.203 \\
  & Kitchen & \cellcolor{tabfirst}31.89 & \cellcolor{tabfirst}0.931 & \cellcolor{tabthird}0.127 & \cellcolor{tabthird}31.41 & \cellcolor{tabsecond}0.929 & \cellcolor{tabsecond}0.124 & 31.29 & \cellcolor{tabthird}0.927 & \cellcolor{tabfirst}0.123 & \cellcolor{tabsecond}31.65 & 0.923 & 0.138 & 30.93 & 0.922 & 0.144 \\
  & Bonsai & \cellcolor{tabfirst}32.55 & \cellcolor{tabfirst}0.947 & \cellcolor{tabthird}0.194 & \cellcolor{tabthird}32.02 & \cellcolor{tabsecond}0.942 & \cellcolor{tabfirst}0.191 & 31.89 & \cellcolor{tabthird}0.941 & \cellcolor{tabsecond}0.194 & \cellcolor{tabsecond}32.36 & 0.938 & 0.215 & 32.00 & 0.940 & 0.215 \\
 & \textbf{Average} & \cellcolor{tabfirst}27.84 & \cellcolor{tabfirst}0.819 & \cellcolor{tabthird}0.231 & \cellcolor{tabsecond}27.52 & \cellcolor{tabthird}0.811 & \cellcolor{tabsecond}0.227 & 27.25 & \cellcolor{tabsecond}0.812 & \cellcolor{tabfirst}0.226 & \cellcolor{tabthird}27.36 & 0.787 & 0.273 & 27.30 & 0.797 & 0.260 \\
 \midrule
 \multirow{3}{*}{\rotatebox{90}{\parbox{1.15cm}{\centering\scriptsize Tanks \& Temples}}} & Truck & \cellcolor{tabfirst}26.00 & \cellcolor{tabfirst}0.890 & \cellcolor{tabthird}0.133 & \cellcolor{tabsecond}25.71 & \cellcolor{tabsecond}0.883& \cellcolor{tabfirst}0.124 & 25.29 & \cellcolor{tabthird}0.880 & \cellcolor{tabsecond}0.128 & \cellcolor{tabthird}25.68 & 0.877 & 0.165 & 25.10 & 0.866 & 0.181 \\
  & Train & \cellcolor{tabfirst}22.60 & \cellcolor{tabsecond}0.830 & \cellcolor{tabsecond}0.194 & 22.18 & 0.819 & \cellcolor{tabthird}0.197 & \cellcolor{tabsecond}22.56 & \cellcolor{tabfirst}0.835 & \cellcolor{tabfirst}0.190 & 22.46 & 0.808 & 0.227 & \cellcolor{tabthird}22.53 & \cellcolor{tabthird}0.820 & 0.215 \\
 & \textbf{Average} & \cellcolor{tabfirst}24.30 & \cellcolor{tabfirst}0.860 & \cellcolor{tabthird}0.164 & \cellcolor{tabthird}23.95 & \cellcolor{tabthird}0.851& \cellcolor{tabsecond}0.160 & 23.92 & \cellcolor{tabsecond}0.857 & \cellcolor{tabfirst}0.159 & \cellcolor{tabsecond}24.07 & 0.842 & 0.196 & 23.81 & 0.843 & 0.198 \\
 \midrule
 \multirow{3}{*}{\rotatebox{90}{\parbox{1.15cm}{\centering\scriptsize Deep Blending}}} & Dr Johnson & \cellcolor{tabsecond}29.65 & \cellcolor{tabfirst}0.909 & \cellcolor{tabfirst}0.245 & \cellcolor{tabfirst}29.81 & \cellcolor{tabsecond}0.905 & \cellcolor{tabthird}0.245 & 29.34 & \cellcolor{tabthird}0.905 & \cellcolor{tabsecond}0.245 & \cellcolor{tabthird}29.40 & 0.899 & 0.266 & 28.68 & 0.896 & 0.281 \\
  & Playroom & \cellcolor{tabsecond}30.46 & \cellcolor{tabfirst}0.916 & \cellcolor{tabfirst}0.238 & \cellcolor{tabfirst}30.60 & \cellcolor{tabsecond}0.908 & \cellcolor{tabsecond}0.241 & \cellcolor{tabthird}30.33 & \cellcolor{tabthird}0.907& \cellcolor{tabthird}0.244 & 30.21 & 0.900 & 0.260 & 29.47 & 0.904 & 0.272 \\
 & \textbf{Average} & \cellcolor{tabsecond}30.06 & \cellcolor{tabfirst}0.912 & \cellcolor{tabfirst}0.241 & \cellcolor{tabfirst}30.21 & \cellcolor{tabsecond}0.907 & \cellcolor{tabsecond}0.243 & \cellcolor{tabthird}29.83 & \cellcolor{tabthird}0.906 & \cellcolor{tabthird}0.244 & 29.81 & 0.900 & 0.263 & 29.07 & 0.900 & 0.277 \\
 \midrule
 \multirow{12}{*}{\rotatebox{90}{\parbox{1.15cm}{\centering\scriptsize DL3DV}}}& 140 & \cellcolor{tabfirst}27.57 & \cellcolor{tabfirst}0.892 & \cellcolor{tabthird}0.197 & \cellcolor{tabsecond}27.49 & \cellcolor{tabsecond}0.891 & \cellcolor{tabfirst}0.192 & 27.24 & \cellcolor{tabthird}0.884 & \cellcolor{tabsecond}0.195 & \cellcolor{tabthird}27.24 & 0.878 & 0.220 & 27.08 & 0.879 & 0.225 \\
  & Car & \cellcolor{tabfirst}30.16 & \cellcolor{tabfirst}0.944 & \cellcolor{tabsecond}0.182 & 29.74 & \cellcolor{tabthird}0.940 & \cellcolor{tabthird}0.187 & 29.72 & 0.939 & \cellcolor{tabfirst}0.180 & \cellcolor{tabsecond}30.05 & \cellcolor{tabsecond}0.941 & 0.192 & \cellcolor{tabthird}29.81 & 0.939 & 0.196 \\
  & Garden & \cellcolor{tabthird}24.97 & \cellcolor{tabsecond}0.783 & \cellcolor{tabthird}0.225 & \cellcolor{tabfirst}25.36 & \cellcolor{tabfirst}0.799 & \cellcolor{tabfirst}0.200 & \cellcolor{tabsecond}24.98 & \cellcolor{tabthird}0.782 & \cellcolor{tabsecond}0.217 & 24.39 & 0.733 & 0.272 & 24.06 & 0.730 & 0.281 \\
  & Greenhouse & \cellcolor{tabsecond}22.32 & \cellcolor{tabfirst}0.759 & \cellcolor{tabsecond}0.249 & \cellcolor{tabfirst}22.33 & \cellcolor{tabsecond}0.756 &\cellcolor{tabfirst}0.249 & 21.32 & 0.709 & \cellcolor{tabthird}0.291 & \cellcolor{tabthird}21.62 & \cellcolor{tabthird}0.717 & 0.293 & 21.17 & 0.705 & 0.296 \\
  & Grills & \cellcolor{tabfirst}27.42 & \cellcolor{tabfirst}0.919 & \cellcolor{tabfirst}0.216 & 26.56 & \cellcolor{tabsecond}0.915 & \cellcolor{tabsecond}0.220 & 26.09 & 0.909 & \cellcolor{tabthird}0.226 & \cellcolor{tabthird}26.70 & \cellcolor{tabthird}0.911 & 0.242 & \cellcolor{tabsecond}26.90 & 0.910 & 0.243 \\
  & Herbary & \cellcolor{tabfirst}32.33 & \cellcolor{tabfirst}0.954 & \cellcolor{tabsecond}0.110 & \cellcolor{tabsecond}32.20 & \cellcolor{tabsecond}0.953 & \cellcolor{tabfirst}0.109 & 31.65 & \cellcolor{tabthird}0.949 & \cellcolor{tabthird}0.111 & \cellcolor{tabthird}31.93 & 0.949 & 0.120 & 31.57 & 0.948 & 0.124 \\
  & Highrise & \cellcolor{tabfirst}27.29 & \cellcolor{tabsecond}0.901 & \cellcolor{tabsecond}0.147 & \cellcolor{tabsecond}27.24 & \cellcolor{tabfirst}0.903 & \cellcolor{tabfirst}0.141 & \cellcolor{tabthird}26.92 & \cellcolor{tabthird}0.897& \cellcolor{tabthird}0.149 & 26.44 & 0.874 & 0.187 & 26.31 & 0.878 & 0.184 \\
  & Roomset & \cellcolor{tabsecond}31.56 & \cellcolor{tabfirst}0.948 & \cellcolor{tabfirst}0.084 & \cellcolor{tabthird}31.38 & \cellcolor{tabsecond}0.947 & \cellcolor{tabsecond}0.085 & 31.35 & 0.945 & \cellcolor{tabthird}0.089 & \cellcolor{tabfirst}31.63 & \cellcolor{tabthird}0.945 & 0.093 & 31.20 & 0.943 & 0.096 \\
  & Statue & \cellcolor{tabsecond}32.21 & \cellcolor{tabsecond}0.945 & \cellcolor{tabthird}0.139 & \cellcolor{tabfirst}32.96 & \cellcolor{tabfirst}0.947 & \cellcolor{tabsecond}0.136 & 31.36 & \cellcolor{tabthird}0.942 & \cellcolor{tabfirst}0.135 & \cellcolor{tabthird}31.75 & 0.930 & 0.170 & 31.16 & 0.926 & 0.180 \\
  & Supermarket & \cellcolor{tabfirst}32.37 & \cellcolor{tabfirst}0.926 & \cellcolor{tabthird}0.191 & \cellcolor{tabsecond}32.12 & \cellcolor{tabsecond}0.924 &\cellcolor{tabsecond}0.190 & 31.71 & \cellcolor{tabthird}0.922 & \cellcolor{tabfirst}0.186 & \cellcolor{tabthird}32.11 & 0.919 & 0.203 & 31.23 & 0.915 & 0.213 \\
  & Vasary & \cellcolor{tabfirst}25.73 & \cellcolor{tabfirst}0.829 & \cellcolor{tabsecond}0.222 & \cellcolor{tabsecond}25.63 & \cellcolor{tabsecond}0.823 & \cellcolor{tabfirst}0.218 & 25.00 & \cellcolor{tabthird}0.814 & \cellcolor{tabthird}0.231 & \cellcolor{tabthird}25.03 & 0.806 & 0.262 & 24.92 & 0.802 & 0.263 \\
 & \textbf{Average} & \cellcolor{tabfirst}28.54 & \cellcolor{tabfirst}0.891 & \cellcolor{tabsecond}0.178 & \cellcolor{tabsecond}28.46 & \cellcolor{tabsecond}0.891 & \cellcolor{tabfirst}0.175 & 27.94 & \cellcolor{tabthird}0.881 & \cellcolor{tabthird}0.183 & \cellcolor{tabthird}28.08 & 0.873 & 0.205 & 27.76 & 0.871 & 0.209 \\
\bottomrule
\end{tabular}}
\caption{Per-scene results at a budget of $\mathrm{500k}$ primitives, SfM initialization.}
\label{tab:results-500000}
\end{table*}
\begin{table*}
\centering
\resizebox{0.87\linewidth}{!}{%
\begin{tabular}{cl|ccc|ccc|ccc|ccc|ccc}
\toprule
  & & \multicolumn{3}{c}{Ours} & \multicolumn{3}{c}{ConeGS} & \multicolumn{3}{c}{EDGS} & \multicolumn{3}{c}{Taming 3DGS} & \multicolumn{3}{c}{3DGS-MCMC} \\
\cmidrule(lr){3-5} \cmidrule(lr){6-8} \cmidrule(lr){9-11} \cmidrule(lr){12-14} \cmidrule(lr){15-17}
  & \textit{Scene} & PSNR & SSIM & LPIPS & PSNR & SSIM & LPIPS & PSNR & SSIM & LPIPS & PSNR & SSIM & LPIPS & PSNR & SSIM & LPIPS \\
\midrule
 \multirow{10}{*}{\rotatebox{90}{\parbox{1.15cm}{\centering\scriptsize Mip-NeRF360}}} & Bicycle & \cellcolor{tabsecond}25.44 & \cellcolor{tabthird}0.778 & \cellcolor{tabthird}0.214 & \cellcolor{tabfirst}25.45 & \cellcolor{tabfirst}0.779 & \cellcolor{tabfirst}0.196 & \cellcolor{tabthird}25.27 & \cellcolor{tabsecond}0.778 & \cellcolor{tabsecond}0.201 & 24.96 & 0.722 & 0.291 & 25.10 & 0.749 & 0.259 \\
  & Flowers & \cellcolor{tabfirst}22.07 & \cellcolor{tabfirst}0.633 & \cellcolor{tabthird}0.313 & \cellcolor{tabthird}21.52 & \cellcolor{tabthird}0.624 & \cellcolor{tabfirst}0.303 & 21.32 & \cellcolor{tabsecond}0.624 & \cellcolor{tabsecond}0.306 & 21.47 & 0.586 & 0.373 & \cellcolor{tabsecond}21.75 & 0.597 & 0.358 \\
  & Garden & \cellcolor{tabfirst}27.68 & \cellcolor{tabfirst}0.867 & \cellcolor{tabsecond}0.120 & \cellcolor{tabsecond}27.45 & \cellcolor{tabsecond}0.861 & \cellcolor{tabfirst}0.115 & 27.08 & \cellcolor{tabthird}0.856 & \cellcolor{tabthird}0.121 & \cellcolor{tabthird}27.19 & 0.843 & 0.153 & 26.86 & 0.846 & 0.147 \\
  & Stump & \cellcolor{tabfirst}27.30 & \cellcolor{tabfirst}0.808 & \cellcolor{tabsecond}0.189 & \cellcolor{tabsecond}27.23 & \cellcolor{tabsecond}0.802 & \cellcolor{tabfirst}0.185 & \cellcolor{tabthird}26.79 & \cellcolor{tabthird}0.784 & \cellcolor{tabthird}0.204 & 26.36 & 0.753 & 0.262 & 26.78 & 0.776 & 0.235 \\
  & Treehill & \cellcolor{tabfirst}22.96 & \cellcolor{tabsecond}0.654 & \cellcolor{tabsecond}0.317 & 22.47 & 0.604 & \cellcolor{tabthird}0.334 & 22.61 & \cellcolor{tabfirst}0.659 & \cellcolor{tabfirst}0.282 & \cellcolor{tabsecond}22.94 & 0.629 & 0.372 & \cellcolor{tabthird}22.88 & \cellcolor{tabthird}0.639 & 0.352 \\
  & Room & \cellcolor{tabfirst}32.45 & \cellcolor{tabfirst}0.931 & \cellcolor{tabfirst}0.191 & \cellcolor{tabsecond}32.21 & 0.927 & \cellcolor{tabthird}0.195 &31.91 & \cellcolor{tabsecond}0.929 & \cellcolor{tabsecond}0.192 & 32.12 & 0.919& 0.221 & \cellcolor{tabthird}32.17 & \cellcolor{tabthird}0.928 & 0.206 \\
  & Counter & \cellcolor{tabfirst}29.64 & \cellcolor{tabfirst}0.920 & \cellcolor{tabsecond}0.174 & 29.03 & 0.910 & \cellcolor{tabthird}0.182 & 29.26 & \cellcolor{tabsecond}0.916 & \cellcolor{tabfirst}0.173 & \cellcolor{tabsecond}29.37 & 0.909 & 0.199 & \cellcolor{tabthird}29.29 & \cellcolor{tabthird}0.915 & 0.189 \\
  & Kitchen & \cellcolor{tabfirst}32.39 & \cellcolor{tabfirst}0.935 & \cellcolor{tabthird}0.117 & 31.67 & \cellcolor{tabthird}0.932 & \cellcolor{tabsecond}0.117 & \cellcolor{tabsecond}31.88 & \cellcolor{tabsecond}0.932 & \cellcolor{tabfirst}0.115 & \cellcolor{tabthird}31.86 & 0.928 & 0.127 & 31.63 & 0.929 & 0.130 \\
  & Bonsai & \cellcolor{tabfirst}32.92 & \cellcolor{tabfirst}0.951 & \cellcolor{tabfirst}0.182 & \cellcolor{tabthird}32.54 & \cellcolor{tabsecond}0.947 & \cellcolor{tabsecond}0.182 & 32.41 & \cellcolor{tabthird}0.946 & \cellcolor{tabthird}0.182 & \cellcolor{tabsecond}32.63 & 0.942 & 0.204 & 32.53 & 0.945 & 0.200 \\
 & \textbf{Average} & \cellcolor{tabfirst}28.10 & \cellcolor{tabfirst}0.831 & \cellcolor{tabthird}0.202 & \cellcolor{tabsecond}27.73 & \cellcolor{tabthird}0.821 & \cellcolor{tabsecond}0.201 & 27.61 & \cellcolor{tabsecond}0.825 & \cellcolor{tabfirst}0.197 & 27.65 & 0.803 & 0.245 & \cellcolor{tabthird}27.67 & 0.814 & 0.231 \\
 \midrule
 \multirow{3}{*}{\rotatebox{90}{\parbox{1.15cm}{\centering\scriptsize Tanks \& Temples}}} & Truck & \cellcolor{tabfirst}26.23 & \cellcolor{tabfirst}0.896 & \cellcolor{tabthird}0.112 & 25.72 & \cellcolor{tabthird}0.888 & \cellcolor{tabsecond}0.109 & 25.62 & \cellcolor{tabsecond}0.889 & \cellcolor{tabfirst}0.105 & \cellcolor{tabsecond}25.87 & 0.883 & 0.150 & \cellcolor{tabthird}25.77 & 0.881 & 0.156 \\
  & Train & \cellcolor{tabfirst}23.02 & \cellcolor{tabsecond}0.840 & \cellcolor{tabsecond}0.172 & 22.21 & 0.824 & \cellcolor{tabthird}0.176 & 22.52 & \cellcolor{tabfirst}0.846 & \cellcolor{tabfirst}0.164 & \cellcolor{tabthird}22.74 & 0.822& 0.205 & \cellcolor{tabsecond}22.78 & \cellcolor{tabthird}0.836 & 0.187 \\
 & \textbf{Average} & \cellcolor{tabfirst}24.62 & \cellcolor{tabfirst}0.868 & \cellcolor{tabsecond}0.142 & 23.97 & 0.856 & \cellcolor{tabthird}0.143 & 24.07 & \cellcolor{tabsecond}0.867 & \cellcolor{tabfirst}0.135 & \cellcolor{tabsecond}24.30 & 0.853 & 0.177 & \cellcolor{tabthird}24.28 & \cellcolor{tabthird}0.858 & 0.171 \\
 \midrule
 \multirow{3}{*}{\rotatebox{90}{\parbox{1.15cm}{\centering\scriptsize Deep Blending}}} & Dr Johnson & \cellcolor{tabfirst}29.88 & \cellcolor{tabfirst}0.910 & \cellcolor{tabsecond}0.234 & \cellcolor{tabsecond}29.84 & \cellcolor{tabthird}0.905 & \cellcolor{tabthird}0.236 & 29.45 & \cellcolor{tabsecond}0.907 & \cellcolor{tabfirst}0.231 & \cellcolor{tabthird}29.66 & 0.903 & 0.253 & 28.58 & 0.898 & 0.273 \\
  & Playroom & \cellcolor{tabthird}30.18 & \cellcolor{tabfirst}0.912 & \cellcolor{tabfirst}0.227 & \cellcolor{tabfirst}30.69 & \cellcolor{tabthird}0.907 & \cellcolor{tabsecond}0.227 & \cellcolor{tabsecond}30.29 & 0.905 & \cellcolor{tabthird}0.230 & 29.83 & 0.898 & 0.251 & 29.97 & \cellcolor{tabsecond}0.911 & 0.258 \\
 & \textbf{Average} & \cellcolor{tabsecond}30.03 & \cellcolor{tabfirst}0.911 & \cellcolor{tabfirst}0.230 & \cellcolor{tabfirst}30.27 & \cellcolor{tabthird}0.906 & \cellcolor{tabthird}0.232 & \cellcolor{tabthird}29.87 & \cellcolor{tabsecond}0.906 & \cellcolor{tabsecond}0.231 & 29.74 & 0.901 & 0.252 & 29.27 & 0.905 & 0.266 \\
 \midrule
 \multirow{12}{*}{\rotatebox{90}{\parbox{1.15cm}{\centering\scriptsize DL3DV}}}& 140 & \cellcolor{tabfirst}27.93 & \cellcolor{tabfirst}0.899 & \cellcolor{tabthird}0.182 & \cellcolor{tabsecond}27.76 & \cellcolor{tabsecond}0.896 & \cellcolor{tabsecond}0.181 & \cellcolor{tabthird}27.63 & \cellcolor{tabthird}0.893 & \cellcolor{tabfirst}0.174 & 27.56 & 0.886 & 0.206 & 27.52 & 0.890 & 0.207 \\
  & Car & \cellcolor{tabfirst}30.35 & \cellcolor{tabfirst}0.945 & \cellcolor{tabsecond}0.177 & 29.93 & 0.942 & \cellcolor{tabthird}0.182 & 29.94 & 0.941 & \cellcolor{tabfirst}0.172 & \cellcolor{tabsecond}30.25 & \cellcolor{tabsecond}0.943 & 0.187 & \cellcolor{tabthird}30.11 & \cellcolor{tabthird}0.942 & 0.189 \\
  & Garden & \cellcolor{tabthird}25.58 & \cellcolor{tabthird}0.813 & \cellcolor{tabthird}0.188 & \cellcolor{tabfirst}25.89 & \cellcolor{tabfirst}0.825 & \cellcolor{tabfirst}0.167 & \cellcolor{tabsecond}25.61 & \cellcolor{tabsecond}0.814 & \cellcolor{tabsecond}0.178 & 25.08 & 0.776 & 0.225 & 24.92 & 0.775 & 0.231 \\
  & Greenhouse & \cellcolor{tabfirst}23.08 & \cellcolor{tabfirst}0.793 & \cellcolor{tabfirst}0.211 & \cellcolor{tabsecond}22.88 & \cellcolor{tabsecond}0.785 & \cellcolor{tabsecond}0.214 & 22.06 & \cellcolor{tabthird}0.750 & \cellcolor{tabthird}0.246 & \cellcolor{tabthird}22.21 & 0.750 & 0.258 & 22.07 & 0.750 & 0.252 \\
  & Grills & \cellcolor{tabfirst}27.84 & \cellcolor{tabfirst}0.924 & \cellcolor{tabfirst}0.205 & 26.69 & \cellcolor{tabthird}0.918 & \cellcolor{tabthird}0.214 & 26.49 & 0.916 & \cellcolor{tabsecond}0.211 & \cellcolor{tabthird}26.91 & 0.915& 0.235 & \cellcolor{tabsecond}27.46 & \cellcolor{tabsecond}0.920 & 0.227 \\
  & Herbary & \cellcolor{tabfirst}32.74 & \cellcolor{tabfirst}0.958 & \cellcolor{tabsecond}0.102 & \cellcolor{tabsecond}32.58 & \cellcolor{tabsecond}0.956 & \cellcolor{tabthird}0.103 & 32.29 & \cellcolor{tabthird}0.954 & \cellcolor{tabfirst}0.101 & \cellcolor{tabthird}32.31 & 0.952 & 0.113 & 32.24 & 0.954 & 0.115 \\
  & Highrise & \cellcolor{tabfirst}27.76 & \cellcolor{tabfirst}0.910 & \cellcolor{tabthird}0.131 & \cellcolor{tabthird}27.46 & \cellcolor{tabthird}0.909 & \cellcolor{tabsecond}0.129 & \cellcolor{tabsecond}27.46 & \cellcolor{tabsecond}0.909& \cellcolor{tabfirst}0.128 & 26.75 & 0.887 & 0.168 & 27.13 & 0.896 & 0.158 \\
  & Roomset & \cellcolor{tabsecond}31.73 & \cellcolor{tabfirst}0.950 & \cellcolor{tabfirst}0.080 & 31.56 & \cellcolor{tabsecond}0.949 & \cellcolor{tabsecond}0.081 & \cellcolor{tabthird}31.66 & \cellcolor{tabthird}0.948 & \cellcolor{tabthird}0.082 & \cellcolor{tabfirst}31.78 & 0.947 & 0.088 & 31.41 & 0.947 & 0.090 \\
  & Statue & \cellcolor{tabsecond}33.04 & \cellcolor{tabsecond}0.952 & \cellcolor{tabthird}0.123 & \cellcolor{tabfirst}33.61 & \cellcolor{tabfirst}0.952 & \cellcolor{tabsecond}0.122 & 32.42 & \cellcolor{tabthird}0.951 & \cellcolor{tabfirst}0.117 & 32.23 & 0.936 & 0.158 & \cellcolor{tabthird}32.46 & 0.940 & 0.160 \\
  & Supermarket & \cellcolor{tabfirst}32.78 & \cellcolor{tabfirst}0.931 & \cellcolor{tabsecond}0.181 & \cellcolor{tabsecond}32.48 & \cellcolor{tabsecond}0.928 & \cellcolor{tabthird}0.181 & 32.23 & \cellcolor{tabthird}0.928 & \cellcolor{tabfirst}0.172 & \cellcolor{tabthird}32.40 & 0.924 & 0.195 & 32.01 & 0.923 & 0.200 \\
  & Vasary & \cellcolor{tabfirst}26.14 & \cellcolor{tabfirst}0.842 & \cellcolor{tabfirst}0.200 & \cellcolor{tabsecond}25.85 & \cellcolor{tabsecond}0.830 & \cellcolor{tabsecond}0.203 & 25.38 & \cellcolor{tabthird}0.827 & \cellcolor{tabthird}0.208 & 25.30 & 0.817 & 0.245 & \cellcolor{tabthird}25.57 & 0.822 & 0.238 \\
 & \textbf{Average} & \cellcolor{tabfirst}29.00 & \cellcolor{tabfirst}0.901 & \cellcolor{tabsecond}0.162 & \cellcolor{tabsecond}28.79 & \cellcolor{tabsecond}0.899 & \cellcolor{tabfirst}0.161 & \cellcolor{tabthird}28.47 & \cellcolor{tabthird}0.894 & \cellcolor{tabthird}0.163 & 28.43 & 0.885 & 0.189 & 28.45 & 0.887 & 0.188 \\
\bottomrule
\end{tabular}}
\caption{Per-scene results at a budget of $\mathrm{1M}$ primitives, SfM initialization.}
\label{tab:results-1000000}
\end{table*}
\begin{table*}
\centering
\resizebox{0.87\linewidth}{!}{%
\begin{tabular}{cl|ccc|ccc|ccc|ccc|ccc}
\toprule
  & & \multicolumn{3}{c}{Ours} & \multicolumn{3}{c}{ConeGS} & \multicolumn{3}{c}{EDGS} & \multicolumn{3}{c}{Taming 3DGS} & \multicolumn{3}{c}{3DGS-MCMC} \\
\cmidrule(lr){3-5} \cmidrule(lr){6-8} \cmidrule(lr){9-11} \cmidrule(lr){12-14} \cmidrule(lr){15-17}
  & \textit{Scene} & PSNR & SSIM & LPIPS & PSNR & SSIM & LPIPS & PSNR & SSIM & LPIPS & PSNR & SSIM & LPIPS & PSNR & SSIM & LPIPS \\
\midrule
 \multirow{10}{*}{\rotatebox{90}{\parbox{1.15cm}{\centering\scriptsize Mip-NeRF360}}} & Bicycle & \cellcolor{tabfirst}25.61 & \cellcolor{tabsecond}0.791 & \cellcolor{tabthird}0.181 & \cellcolor{tabsecond}25.49 & \cellcolor{tabthird}0.783 &\cellcolor{tabsecond}0.174 & \cellcolor{tabthird}25.41 & \cellcolor{tabfirst}0.791 & \cellcolor{tabfirst}0.170 & 25.21 & 0.750 & 0.246 & 25.27 & 0.768 & 0.224 \\
  & Flowers & \cellcolor{tabfirst}21.92 & \cellcolor{tabfirst}0.642 & \cellcolor{tabsecond}0.274 & 21.41 & \cellcolor{tabthird}0.628 & \cellcolor{tabthird}0.276 & 21.48 & \cellcolor{tabsecond}0.639 & \cellcolor{tabfirst}0.272 & \cellcolor{tabthird}21.65 & 0.602 & 0.349 & \cellcolor{tabsecond}21.91 & 0.614 & 0.328 \\
  & Garden & \cellcolor{tabfirst}27.98 & \cellcolor{tabfirst}0.879 & \cellcolor{tabfirst}0.096 & \cellcolor{tabsecond}27.64 & \cellcolor{tabsecond}0.870 & \cellcolor{tabsecond}0.098 & 27.42 & \cellcolor{tabthird}0.869 & \cellcolor{tabthird}0.100 & \cellcolor{tabthird}27.56 & 0.860 & 0.122 & 27.44 & 0.864 & 0.118 \\
  & Stump & \cellcolor{tabfirst}27.24 & \cellcolor{tabfirst}0.808 & \cellcolor{tabfirst}0.174 & \cellcolor{tabsecond}27.01 & \cellcolor{tabsecond}0.795 & \cellcolor{tabsecond}0.178 & 26.91 & \cellcolor{tabthird}0.790 & \cellcolor{tabthird}0.185 & 26.55 & 0.766 & 0.231 & \cellcolor{tabthird}26.99 & 0.789 & 0.208 \\
  & Treehill & \cellcolor{tabthird}22.78 & \cellcolor{tabsecond}0.653 & \cellcolor{tabsecond}0.283 & 22.21 & 0.592 & \cellcolor{tabthird}0.302 & 22.52 & \cellcolor{tabfirst}0.656 & \cellcolor{tabfirst}0.254 & \cellcolor{tabfirst}23.09 & 0.639 & 0.339 & \cellcolor{tabsecond}22.81 & \cellcolor{tabthird}0.647 & 0.316 \\
  & Room & \cellcolor{tabfirst}32.78 & \cellcolor{tabfirst}0.933 & \cellcolor{tabfirst}0.184 & 32.27 & 0.929 & \cellcolor{tabthird}0.191 & 32.27 & \cellcolor{tabsecond}0.932 & \cellcolor{tabsecond}0.184 & \cellcolor{tabthird}32.30 & 0.924 & 0.206 & \cellcolor{tabsecond}32.46 & \cellcolor{tabthird}0.931 & 0.196 \\
  & Counter & \cellcolor{tabfirst}29.75 & \cellcolor{tabfirst}0.922 & \cellcolor{tabsecond}0.167 & 29.27 & 0.912 & \cellcolor{tabthird}0.175 & 29.44 & \cellcolor{tabthird}0.919 & \cellcolor{tabfirst}0.165 & \cellcolor{tabsecond}29.54 & 0.913 & 0.190 & \cellcolor{tabthird}29.53 & \cellcolor{tabsecond}0.919 & 0.178 \\
  & Kitchen & \cellcolor{tabfirst}32.57 & \cellcolor{tabfirst}0.937 & \cellcolor{tabsecond}0.113 & 31.82 & \cellcolor{tabthird}0.934 & \cellcolor{tabthird}0.114 & \cellcolor{tabthird}32.13 & \cellcolor{tabsecond}0.935 & \cellcolor{tabfirst}0.111 & 31.90 & 0.930 & 0.121 & \cellcolor{tabsecond}32.18 & 0.934 & 0.121 \\
  & Bonsai & \cellcolor{tabfirst}33.27 & \cellcolor{tabfirst}0.953 & \cellcolor{tabfirst}0.175 & 32.56 & 0.947 & \cellcolor{tabthird}0.178 & 32.76 & \cellcolor{tabthird}0.949 & \cellcolor{tabsecond}0.175 & \cellcolor{tabthird}32.86 & 0.944& 0.199 & \cellcolor{tabsecond}32.87 & \cellcolor{tabsecond}0.949 & 0.190 \\
 & \textbf{Average} & \cellcolor{tabfirst}28.21 & \cellcolor{tabfirst}0.835 & \cellcolor{tabsecond}0.183 & 27.74 & 0.821 & \cellcolor{tabthird}0.187 & 27.81 & \cellcolor{tabsecond}0.831 & \cellcolor{tabfirst}0.180 & \cellcolor{tabthird}27.85 & 0.814 & 0.222 & \cellcolor{tabsecond}27.94 & \cellcolor{tabthird}0.824 & 0.209 \\
 \midrule
 \multirow{3}{*}{\rotatebox{90}{\parbox{1.15cm}{\centering\scriptsize Tanks \& Temples}}} & Truck & \cellcolor{tabfirst}26.38 & \cellcolor{tabfirst}0.900 & \cellcolor{tabsecond}0.099 & 25.74 & 0.888 & \cellcolor{tabthird}0.102 & \cellcolor{tabthird}25.88 & \cellcolor{tabsecond}0.893 & \cellcolor{tabfirst}0.092 & 25.82& 0.886 & 0.139 & \cellcolor{tabsecond}26.20 & \cellcolor{tabthird}0.890 & 0.138 \\
  & Train & \cellcolor{tabthird}22.79 & \cellcolor{tabthird}0.842 & \cellcolor{tabsecond}0.158 & 22.01 & 0.825 & 0.165 & 22.59 & \cellcolor{tabfirst}0.851 & \cellcolor{tabfirst}0.149 & \cellcolor{tabsecond}22.82 & 0.831 & 0.189 & \cellcolor{tabfirst}23.06 & \cellcolor{tabsecond}0.848 & \cellcolor{tabthird}0.164 \\
 & \textbf{Average} & \cellcolor{tabsecond}24.59 & \cellcolor{tabsecond}0.871 &\cellcolor{tabsecond}0.129 & 23.88 & 0.856 & \cellcolor{tabthird}0.134 & 24.24 & \cellcolor{tabfirst}0.872 & \cellcolor{tabfirst}0.121 & \cellcolor{tabthird}24.32 & 0.858 & 0.164 & \cellcolor{tabfirst}24.63 & \cellcolor{tabthird}0.869 & 0.151 \\
 \midrule
 \multirow{3}{*}{\rotatebox{90}{\parbox{1.15cm}{\centering\scriptsize Deep Blending}}} & Dr Johnson & \cellcolor{tabfirst}29.93 & \cellcolor{tabfirst}0.911 & \cellcolor{tabsecond}0.228 & \cellcolor{tabsecond}29.84 & 0.902 & \cellcolor{tabthird}0.230 & \cellcolor{tabthird}29.63 & \cellcolor{tabsecond}0.906 & \cellcolor{tabfirst}0.221 & 29.36 & \cellcolor{tabthird}0.904 & 0.243 & 28.00 & 0.893 & 0.273 \\
  & Playroom & \cellcolor{tabsecond}30.39 & \cellcolor{tabsecond}0.910 & \cellcolor{tabthird}0.218 & \cellcolor{tabfirst}30.48 & 0.902 & \cellcolor{tabfirst}0.214 & \cellcolor{tabthird}30.29 & \cellcolor{tabthird}0.904 & \cellcolor{tabsecond}0.218 & 29.93 & 0.902 & 0.241 & 30.22 & \cellcolor{tabfirst}0.913 & 0.247 \\
 & \textbf{Average} & \cellcolor{tabsecond}30.16 & \cellcolor{tabfirst}0.911 & \cellcolor{tabthird}0.223 & \cellcolor{tabfirst}30.16 & 0.902 & \cellcolor{tabsecond}0.222 & \cellcolor{tabthird}29.96 & \cellcolor{tabsecond}0.905 & \cellcolor{tabfirst}0.219 & 29.65 & 0.903 & 0.242 & 29.11 & \cellcolor{tabthird}0.903 & 0.260 \\
 \midrule
 \multirow{12}{*}{\rotatebox{90}{\parbox{1.15cm}{\centering\scriptsize DL3DV}}}& 140 & \cellcolor{tabfirst}28.11 & \cellcolor{tabfirst}0.902 & \cellcolor{tabthird}0.174 & \cellcolor{tabthird}27.90 & \cellcolor{tabsecond}0.899 & \cellcolor{tabsecond}0.171 & 27.89 & \cellcolor{tabthird}0.899 & \cellcolor{tabfirst}0.159& 27.71 & 0.891 & 0.197 & \cellcolor{tabsecond}27.94 & 0.899 & 0.188 \\
  & Car & \cellcolor{tabfirst}30.50 & \cellcolor{tabfirst}0.946 & \cellcolor{tabsecond}0.173 & 30.06 & 0.943 & \cellcolor{tabthird}0.178 & 30.06 & 0.942 & \cellcolor{tabfirst}0.164 & \cellcolor{tabthird}30.23 & \cellcolor{tabthird}0.943 & 0.187 & \cellcolor{tabsecond}30.35 & \cellcolor{tabsecond}0.944 & 0.184 \\
  & Garden & \cellcolor{tabthird}26.07 & \cellcolor{tabthird}0.834 & \cellcolor{tabthird}0.163 & \cellcolor{tabfirst}26.23 & \cellcolor{tabfirst}0.840 & \cellcolor{tabfirst}0.149 & \cellcolor{tabsecond}26.09 & \cellcolor{tabsecond}0.834 & \cellcolor{tabsecond}0.154 & 25.57 & 0.804 & 0.194 & 25.59 & 0.809 & 0.193 \\
  & Greenhouse & \cellcolor{tabfirst}23.57 & \cellcolor{tabfirst}0.816 & \cellcolor{tabfirst}0.184 & \cellcolor{tabsecond}23.23 & \cellcolor{tabsecond}0.803 & \cellcolor{tabsecond}0.192 & 22.67 & 0.779 & \cellcolor{tabthird}0.212 & 22.58 &0.771 & 0.234 & \cellcolor{tabthird}22.84 & \cellcolor{tabthird}0.784 & 0.216 \\
  & Grills & \cellcolor{tabfirst}28.04 & \cellcolor{tabfirst}0.927 & \cellcolor{tabfirst}0.198 & 26.73 & 0.919 & \cellcolor{tabthird}0.208 & 26.78 & \cellcolor{tabthird}0.920 & \cellcolor{tabsecond}0.201 & \cellcolor{tabthird}27.07 & 0.917& 0.228 & \cellcolor{tabsecond}27.82 & \cellcolor{tabsecond}0.925 & 0.217 \\
  & Herbary & \cellcolor{tabfirst}33.01 & \cellcolor{tabfirst}0.959 & \cellcolor{tabsecond}0.097 & \cellcolor{tabsecond}32.81 & \cellcolor{tabsecond}0.958 & \cellcolor{tabthird}0.099 & \cellcolor{tabthird}32.68 & \cellcolor{tabthird}0.957 & \cellcolor{tabfirst}0.095 & 32.52 & 0.954 & 0.109 & 32.68 & 0.957 & 0.108 \\
  & Highrise & \cellcolor{tabfirst}28.12 & \cellcolor{tabfirst}0.916 & \cellcolor{tabsecond}0.120 & 27.67 & \cellcolor{tabthird}0.913 & \cellcolor{tabthird}0.121 & \cellcolor{tabsecond}27.86 & \cellcolor{tabsecond}0.915 & \cellcolor{tabfirst}0.116 & 26.94 & 0.896 & 0.155 & \cellcolor{tabthird}27.71 & 0.910 & 0.138 \\
  & Roomset & \cellcolor{tabfirst}31.97 & \cellcolor{tabfirst}0.951 & \cellcolor{tabfirst}0.077 & 31.67 & \cellcolor{tabsecond}0.950 & \cellcolor{tabsecond}0.079 & \cellcolor{tabsecond}31.90 & \cellcolor{tabthird}0.949 & \cellcolor{tabthird}0.079 & \cellcolor{tabthird}31.82 & 0.948 & 0.087 & 31.74 & 0.949 & 0.085 \\
  & Statue & \cellcolor{tabsecond}33.53 & \cellcolor{tabsecond}0.957 & \cellcolor{tabsecond}0.111 & \cellcolor{tabfirst}33.85 & \cellcolor{tabthird}0.954 & \cellcolor{tabthird}0.116 & 33.15 & \cellcolor{tabfirst}0.957 & \cellcolor{tabfirst}0.106 & 32.42 & 0.939 & 0.151 & \cellcolor{tabthird}33.53 & 0.949 & 0.142 \\
  & Supermarket & \cellcolor{tabfirst}33.07 & \cellcolor{tabfirst}0.933 & \cellcolor{tabsecond}0.173 & \cellcolor{tabsecond}32.65 & \cellcolor{tabthird}0.930 &\cellcolor{tabthird}0.174 & 32.59 & \cellcolor{tabsecond}0.931 & \cellcolor{tabfirst}0.161 & \cellcolor{tabthird}32.60 & 0.926 & 0.189 & 32.50 & 0.928 & 0.191 \\
  & Vasary & \cellcolor{tabfirst}26.40 & \cellcolor{tabfirst}0.849 & \cellcolor{tabfirst}0.185 & \cellcolor{tabthird}25.97 & 0.834 & \cellcolor{tabsecond}0.193& 25.62 & \cellcolor{tabthird}0.834 & \cellcolor{tabthird}0.194 & 25.50 & 0.824& 0.232 & \cellcolor{tabsecond}26.01 & \cellcolor{tabsecond}0.835 & 0.215 \\
 & \textbf{Average} & \cellcolor{tabfirst}29.31 & \cellcolor{tabfirst}0.908 & \cellcolor{tabsecond}0.150 & \cellcolor{tabsecond}28.98 & \cellcolor{tabsecond}0.904 & \cellcolor{tabthird}0.153 & 28.85 & \cellcolor{tabthird}0.902 & \cellcolor{tabfirst}0.149 & 28.63 & 0.892 & 0.178 & \cellcolor{tabthird}28.97 & 0.899 & 0.171 \\
\bottomrule
\end{tabular}}
\caption{Per-scene results at a budget of $\mathrm{2M}$ primitives, SfM initialization.}
\label{tab:results-2000000}
\end{table*}

\begin{table}
\centering
\resizebox{\linewidth}{!}{%
\begin{tabular}{cl|ccc|ccc|ccc}
\toprule
  & & \multicolumn{3}{c}{Ours} & \multicolumn{3}{c}{3DGS-MCMC} & \multicolumn{3}{c}{\begin{tabular}{@{}c@{}}Ours +\\3DGS-MCMC noise\end{tabular}} \\
\cmidrule(lr){3-5} \cmidrule(lr){6-8} \cmidrule(lr){9-11}
  & \textit{Scene} & PSNR & SSIM & LPIPS & PSNR & SSIM & LPIPS & PSNR & SSIM & LPIPS \\
\midrule
 \multirow{10}{*}{\rotatebox{90}{\scriptsize $\mathrm{100k}$}} & Bicycle & \cellcolor{tabfirst}23.14 & \cellcolor{tabfirst}0.587 & \cellcolor{tabfirst}0.423 & \cellcolor{tabsecond}21.28 & \cellcolor{tabsecond}0.458 & \cellcolor{tabsecond}0.537 & \cellcolor{tabthird}20.87 & \cellcolor{tabthird}0.422 & \cellcolor{tabthird}0.592 \\
  & Flowers & \cellcolor{tabfirst}19.93 & \cellcolor{tabfirst}0.468 & \cellcolor{tabfirst}0.490 & \cellcolor{tabthird}18.18 & \cellcolor{tabthird}0.363 & \cellcolor{tabthird}0.568 & \cellcolor{tabsecond}19.67 & \cellcolor{tabsecond}0.434 & \cellcolor{tabsecond}0.519 \\
  & Garden & \cellcolor{tabfirst}24.43 & \cellcolor{tabfirst}0.687 & \cellcolor{tabfirst}0.351 & \cellcolor{tabthird}23.29 & \cellcolor{tabthird}0.601 & \cellcolor{tabthird}0.415 & \cellcolor{tabsecond}23.52 & \cellcolor{tabsecond}0.629 & \cellcolor{tabsecond}0.388 \\
  & Stump & \cellcolor{tabfirst}23.19 & \cellcolor{tabfirst}0.580 & \cellcolor{tabfirst}0.438 & \cellcolor{tabsecond}19.91 & \cellcolor{tabsecond}0.388 & \cellcolor{tabsecond}0.578 & \cellcolor{tabthird}19.44 & \cellcolor{tabthird}0.371 & \cellcolor{tabthird}0.643 \\
  & Treehill & \cellcolor{tabsecond}21.80 & \cellcolor{tabfirst}0.538 & \cellcolor{tabfirst}0.501 & \cellcolor{tabthird}20.49 & \cellcolor{tabthird}0.452 & \cellcolor{tabthird}0.575 & \cellcolor{tabfirst}21.89 & \cellcolor{tabsecond}0.512 & \cellcolor{tabsecond}0.532 \\
  & Room & \cellcolor{tabfirst}29.56 & \cellcolor{tabfirst}0.884 & \cellcolor{tabfirst}0.290 & \cellcolor{tabthird}28.45 & \cellcolor{tabthird}0.860 & \cellcolor{tabthird}0.330 & \cellcolor{tabsecond}28.80 & \cellcolor{tabsecond}0.863 & \cellcolor{tabsecond}0.329 \\
  & Counter & \cellcolor{tabfirst}27.67 & \cellcolor{tabfirst}0.865 & \cellcolor{tabfirst}0.278 & \cellcolor{tabthird}26.27 & \cellcolor{tabthird}0.823 & \cellcolor{tabthird}0.329 & \cellcolor{tabsecond}26.82 & \cellcolor{tabsecond}0.844 & \cellcolor{tabsecond}0.304 \\
  & Kitchen & \cellcolor{tabfirst}29.23 & \cellcolor{tabfirst}0.886 & \cellcolor{tabfirst}0.215 & \cellcolor{tabthird}25.96 & \cellcolor{tabthird}0.775 & \cellcolor{tabthird}0.325 & \cellcolor{tabsecond}27.04 & \cellcolor{tabsecond}0.829 & \cellcolor{tabsecond}0.284 \\
  & Bonsai & \cellcolor{tabfirst}29.51 & \cellcolor{tabfirst}0.897 & \cellcolor{tabfirst}0.281 & \cellcolor{tabthird}27.44 & \cellcolor{tabthird}0.827 & \cellcolor{tabthird}0.358 & \cellcolor{tabsecond}27.47 & \cellcolor{tabsecond}0.837 & \cellcolor{tabsecond}0.352 \\
 & \textbf{Average} & \cellcolor{tabfirst}25.38 & \cellcolor{tabfirst}0.710 & \cellcolor{tabfirst}0.363 & \cellcolor{tabthird}23.48 & \cellcolor{tabthird}0.616 & \cellcolor{tabthird}0.446 & \cellcolor{tabsecond}23.95 & \cellcolor{tabsecond}0.638 & \cellcolor{tabsecond}0.438 \\
\midrule
 \multirow{10}{*}{\rotatebox{90}{\scriptsize 3DGS}} & Bicycle & \cellcolor{tabfirst}25.19 & \cellcolor{tabfirst}0.773 & \cellcolor{tabfirst}0.177 & \cellcolor{tabsecond}25.04 & \cellcolor{tabsecond}0.769 & \cellcolor{tabsecond}0.210 & \cellcolor{tabthird}23.54 & \cellcolor{tabthird}0.603 & \cellcolor{tabthird}0.380 \\
  & Flowers & \cellcolor{tabthird}20.98 & \cellcolor{tabsecond}0.606 & \cellcolor{tabfirst}0.276 & \cellcolor{tabsecond}21.28 & \cellcolor{tabthird}0.603 & \cellcolor{tabsecond}0.306 & \cellcolor{tabfirst}21.81 & \cellcolor{tabfirst}0.614 & \cellcolor{tabthird}0.318 \\
  & Garden & \cellcolor{tabsecond}27.59 & \cellcolor{tabfirst}0.876 & \cellcolor{tabfirst}0.090 & \cellcolor{tabthird}27.47 & \cellcolor{tabsecond}0.871 & \cellcolor{tabthird}0.103 & \cellcolor{tabfirst}27.78 & \cellcolor{tabfirst}0.876 & \cellcolor{tabsecond}0.093 \\
  & Stump & \cellcolor{tabsecond}26.04 & \cellcolor{tabsecond}0.760 & \cellcolor{tabsecond}0.202 & \cellcolor{tabfirst}26.81 & \cellcolor{tabfirst}0.787 & \cellcolor{tabfirst}0.199 & \cellcolor{tabthird}18.44 & \cellcolor{tabthird}0.336 & \cellcolor{tabthird}0.614 \\
  & Treehill & \cellcolor{tabthird}21.94 & \cellcolor{tabsecond}0.629 & \cellcolor{tabfirst}0.277 & \cellcolor{tabsecond}22.54 & \cellcolor{tabfirst}0.634 & \cellcolor{tabsecond}0.320 & \cellcolor{tabfirst}22.58 & \cellcolor{tabthird}0.605 & \cellcolor{tabthird}0.375 \\
  & Room & \cellcolor{tabfirst}32.02 & \cellcolor{tabfirst}0.928 & \cellcolor{tabfirst}0.191 & \cellcolor{tabthird}31.77 & \cellcolor{tabsecond}0.924 & \cellcolor{tabthird}0.207 & \cellcolor{tabsecond}31.80 & \cellcolor{tabsecond}0.924 & \cellcolor{tabsecond}0.206 \\
  & Counter & \cellcolor{tabfirst}29.48 & \cellcolor{tabfirst}0.917 & \cellcolor{tabfirst}0.178 & \cellcolor{tabthird}28.97 & \cellcolor{tabthird}0.909 & \cellcolor{tabsecond}0.197 & \cellcolor{tabsecond}29.02 & \cellcolor{tabsecond}0.910 & \cellcolor{tabsecond}0.197 \\
  & Kitchen & \cellcolor{tabsecond}31.76 & \cellcolor{tabfirst}0.933 & \cellcolor{tabfirst}0.118 & \cellcolor{tabthird}31.73 & \cellcolor{tabthird}0.927 & \cellcolor{tabthird}0.133 & \cellcolor{tabfirst}32.05 & \cellcolor{tabsecond}0.931 & \cellcolor{tabsecond}0.126 \\
  & Bonsai & \cellcolor{tabfirst}32.64 & \cellcolor{tabfirst}0.947 & \cellcolor{tabfirst}0.187 & \cellcolor{tabsecond}32.14 & \cellcolor{tabsecond}0.940 & \cellcolor{tabsecond}0.210 & \cellcolor{tabthird}31.86 & \cellcolor{tabthird}0.936 & \cellcolor{tabthird}0.219 \\
  & \textbf{Average} & \cellcolor{tabsecond}27.52 & \cellcolor{tabfirst}0.819 & \cellcolor{tabfirst}0.188 & \cellcolor{tabfirst}27.53 & \cellcolor{tabsecond}0.818 & \cellcolor{tabsecond}0.210 & \cellcolor{tabthird}26.54 & \cellcolor{tabthird}0.748 & \cellcolor{tabthird}0.281 \\
\bottomrule
\end{tabular}}
\caption{Per-scene results on Mip-NeRF 360 with random initialization, at a budget of $\mathrm{100k}$ primitives (top) and at the primitive counts of the original 3DGS.
In each row, red, orange and yellow mark the best, second and third method; equal values share a color.}
\label{tab:results-random}
\end{table}

\begin{table}
\centering
\resizebox{\linewidth}{!}{%
\begin{tabular}{cl|ccc|ccc}
\toprule
  & & \multicolumn{3}{c}{Ours (EDGS init.)} & \multicolumn{3}{c}{EDGS} \\
\cmidrule(lr){3-5} \cmidrule(lr){6-8}
  & \textit{Scene} & PSNR & SSIM & LPIPS & PSNR & SSIM & LPIPS \\
\midrule
 \multirow{10}{*}{\rotatebox{90}{\parbox{1.15cm}{\centering\scriptsize Mip-NeRF360}}} & Bicycle & \cellcolor{tabfirst}23.85 & \cellcolor{tabfirst}0.637 & \cellcolor{tabsecond}0.397 & \cellcolor{tabsecond}23.71 & \cellcolor{tabsecond}0.636 & \cellcolor{tabfirst}0.392 \\
  & Flowers & \cellcolor{tabfirst}20.61 & \cellcolor{tabfirst}0.508 & \cellcolor{tabsecond}0.471 & \cellcolor{tabsecond}20.12 & \cellcolor{tabsecond}0.496 & \cellcolor{tabfirst}0.467 \\
  & Garden & \cellcolor{tabfirst}24.81 & \cellcolor{tabfirst}0.712 & \cellcolor{tabsecond}0.344 & \cellcolor{tabsecond}24.45 & \cellcolor{tabsecond}0.705 & \cellcolor{tabfirst}0.341 \\
  & Stump & \cellcolor{tabfirst}25.78 & \cellcolor{tabfirst}0.709 & \cellcolor{tabfirst}0.353 & \cellcolor{tabsecond}25.13 & \cellcolor{tabsecond}0.673 & \cellcolor{tabsecond}0.377 \\
  & Treehill & \cellcolor{tabfirst}22.69 & \cellcolor{tabsecond}0.583 & \cellcolor{tabsecond}0.468 & \cellcolor{tabsecond}22.16 & \cellcolor{tabfirst}0.591 & \cellcolor{tabfirst}0.449 \\
  & Room & \cellcolor{tabfirst}31.16 & \cellcolor{tabfirst}0.906 & \cellcolor{tabfirst}0.255 & \cellcolor{tabsecond}29.89 & \cellcolor{tabsecond}0.894 & \cellcolor{tabsecond}0.267 \\
  & Counter & \cellcolor{tabfirst}28.21 & \cellcolor{tabfirst}0.885 & \cellcolor{tabfirst}0.248 & \cellcolor{tabsecond}27.78 & \cellcolor{tabsecond}0.876 & \cellcolor{tabsecond}0.253 \\
  & Kitchen & \cellcolor{tabfirst}30.05 & \cellcolor{tabfirst}0.896 & \cellcolor{tabsecond}0.196 & \cellcolor{tabsecond}28.92 & \cellcolor{tabsecond}0.890 & \cellcolor{tabfirst}0.189 \\
  & Bonsai & \cellcolor{tabfirst}30.63 & \cellcolor{tabfirst}0.918 & \cellcolor{tabfirst}0.253 & \cellcolor{tabsecond}29.63 & \cellcolor{tabsecond}0.905 & \cellcolor{tabsecond}0.258 \\
  & \textbf{Average} & \cellcolor{tabfirst}26.42 & \cellcolor{tabfirst}0.750 & \cellcolor{tabfirst}0.332 & \cellcolor{tabsecond}25.75 & \cellcolor{tabsecond}0.741 & \cellcolor{tabsecond}0.333 \\
 \midrule
 \multirow{3}{*}{\rotatebox{90}{\parbox{1.15cm}{\centering\scriptsize Tanks \& Temples}}} & Truck & \cellcolor{tabfirst}24.81 & \cellcolor{tabfirst}0.851 & \cellcolor{tabfirst}0.213 & \cellcolor{tabsecond}23.80 & \cellcolor{tabsecond}0.830 & \cellcolor{tabsecond}0.222 \\
  & Train & \cellcolor{tabfirst}22.02 & \cellcolor{tabfirst}0.781 & \cellcolor{tabfirst}0.277 & \cellcolor{tabsecond}21.59 & \cellcolor{tabsecond}0.774 & \cellcolor{tabsecond}0.285 \\
  & \textbf{Average} & \cellcolor{tabfirst}23.41 & \cellcolor{tabfirst}0.816 & \cellcolor{tabfirst}0.245 & \cellcolor{tabsecond}22.70 & \cellcolor{tabsecond}0.802 & \cellcolor{tabsecond}0.254 \\
 \midrule
 \multirow{3}{*}{\rotatebox{90}{\parbox{1.15cm}{\centering\scriptsize Deep Blending}}} & Dr Johnson & \cellcolor{tabfirst}29.31 & \cellcolor{tabfirst}0.895 & \cellcolor{tabfirst}0.285 & \cellcolor{tabsecond}28.49 & \cellcolor{tabsecond}0.885 & \cellcolor{tabsecond}0.299 \\
  & Playroom & \cellcolor{tabfirst}30.32 & \cellcolor{tabfirst}0.909 & \cellcolor{tabfirst}0.269 & \cellcolor{tabsecond}29.49 & \cellcolor{tabsecond}0.894 & \cellcolor{tabsecond}0.291 \\
  & \textbf{Average} & \cellcolor{tabfirst}29.81 & \cellcolor{tabfirst}0.902 & \cellcolor{tabfirst}0.277 & \cellcolor{tabsecond}28.99 & \cellcolor{tabsecond}0.889 & \cellcolor{tabsecond}0.295 \\
 \midrule
 \multirow{12}{*}{\rotatebox{90}{\parbox{1.15cm}{\centering\scriptsize DL3DV}}} & 140 & \cellcolor{tabfirst}26.47 & \cellcolor{tabfirst}0.864 & \cellcolor{tabfirst}0.244 & \cellcolor{tabsecond}25.93 & \cellcolor{tabsecond}0.846 & \cellcolor{tabsecond}0.267 \\
  & Car & \cellcolor{tabfirst}29.33 & \cellcolor{tabfirst}0.934 & \cellcolor{tabfirst}0.202 & \cellcolor{tabsecond}28.77 & \cellcolor{tabsecond}0.928 & \cellcolor{tabsecond}0.209 \\
  & Garden & \cellcolor{tabfirst}23.28 & \cellcolor{tabfirst}0.673 & \cellcolor{tabfirst}0.344 & \cellcolor{tabsecond}23.04 & \cellcolor{tabsecond}0.662 & \cellcolor{tabsecond}0.346 \\
  & Greenhouse & \cellcolor{tabfirst}20.19 & \cellcolor{tabfirst}0.630 & \cellcolor{tabfirst}0.377 & \cellcolor{tabsecond}19.33 & \cellcolor{tabsecond}0.580 & \cellcolor{tabsecond}0.418 \\
  & Grills & \cellcolor{tabfirst}26.02 & \cellcolor{tabfirst}0.897 & \cellcolor{tabfirst}0.262 & \cellcolor{tabsecond}24.72 & \cellcolor{tabsecond}0.876 & \cellcolor{tabsecond}0.283 \\
  & Herbary & \cellcolor{tabfirst}30.50 & \cellcolor{tabfirst}0.936 & \cellcolor{tabfirst}0.142 & \cellcolor{tabsecond}29.35 & \cellcolor{tabsecond}0.919 & \cellcolor{tabsecond}0.159 \\
  & Highrise & \cellcolor{tabfirst}25.67 & \cellcolor{tabfirst}0.859 & \cellcolor{tabfirst}0.212 & \cellcolor{tabsecond}24.83 & \cellcolor{tabsecond}0.842 & \cellcolor{tabsecond}0.231 \\
  & Roomset & \cellcolor{tabfirst}30.98 & \cellcolor{tabfirst}0.939 & \cellcolor{tabfirst}0.106 & \cellcolor{tabsecond}29.99 & \cellcolor{tabsecond}0.926 & \cellcolor{tabsecond}0.127 \\
  & Statue & \cellcolor{tabfirst}29.99 & \cellcolor{tabfirst}0.913 & \cellcolor{tabfirst}0.197 & \cellcolor{tabsecond}28.46 & \cellcolor{tabsecond}0.899 & \cellcolor{tabsecond}0.209 \\
  & Supermarket & \cellcolor{tabfirst}30.27 & \cellcolor{tabfirst}0.898 & \cellcolor{tabfirst}0.241 & \cellcolor{tabsecond}29.11 & \cellcolor{tabsecond}0.884 & \cellcolor{tabsecond}0.255 \\
  & Vasary & \cellcolor{tabfirst}24.35 & \cellcolor{tabfirst}0.786 & \cellcolor{tabfirst}0.288 & \cellcolor{tabsecond}23.60 & \cellcolor{tabsecond}0.759 & \cellcolor{tabsecond}0.314 \\
  & \textbf{Average} & \cellcolor{tabfirst}27.00 & \cellcolor{tabfirst}0.848 & \cellcolor{tabfirst}0.238 & \cellcolor{tabsecond}26.10 & \cellcolor{tabsecond}0.829 & \cellcolor{tabsecond}0.256 \\
\bottomrule
\end{tabular}}
\caption{Per-scene results at a budget of $\mathrm{100k}$ primitives; our method initialized from the EDGS point cloud.}
\label{tab:results-edgsinit-100000}
\end{table}

\begin{table}
\centering
\resizebox{\linewidth}{!}{%
\begin{tabular}{cl|ccc|ccc}
\toprule
  & & \multicolumn{3}{c}{Ours (EDGS init.)} & \multicolumn{3}{c}{EDGS} \\
\cmidrule(lr){3-5} \cmidrule(lr){6-8}
  & \textit{Scene} & PSNR & SSIM & LPIPS & PSNR & SSIM & LPIPS \\
\midrule
 \multirow{10}{*}{\rotatebox{90}{\parbox{1.15cm}{\centering\scriptsize Mip-NeRF360}}} & Bicycle & \cellcolor{tabfirst}25.16 & \cellcolor{tabfirst}0.759 & \cellcolor{tabsecond}0.256 & \cellcolor{tabsecond}24.97 & \cellcolor{tabsecond}0.752 & \cellcolor{tabfirst}0.251 \\
  & Flowers & \cellcolor{tabfirst}21.82 & \cellcolor{tabfirst}0.613 & \cellcolor{tabsecond}0.353 & \cellcolor{tabsecond}21.08 & \cellcolor{tabsecond}0.598 & \cellcolor{tabfirst}0.351 \\
  & Garden & \cellcolor{tabfirst}27.18 & \cellcolor{tabfirst}0.847 & \cellcolor{tabfirst}0.162 & \cellcolor{tabsecond}26.45 & \cellcolor{tabsecond}0.831 & \cellcolor{tabsecond}0.164 \\
  & Stump & \cellcolor{tabfirst}27.41 & \cellcolor{tabfirst}0.806 & \cellcolor{tabfirst}0.214 & \cellcolor{tabsecond}26.52 & \cellcolor{tabsecond}0.769 & \cellcolor{tabsecond}0.238 \\
  & Treehill & \cellcolor{tabfirst}23.04 & \cellcolor{tabsecond}0.648 & \cellcolor{tabsecond}0.346 & \cellcolor{tabsecond}22.57 & \cellcolor{tabfirst}0.653 & \cellcolor{tabfirst}0.319 \\
  & Room & \cellcolor{tabfirst}32.52 & \cellcolor{tabfirst}0.930 & \cellcolor{tabfirst}0.195 & \cellcolor{tabsecond}31.48 & \cellcolor{tabsecond}0.923 & \cellcolor{tabsecond}0.205 \\
  & Counter & \cellcolor{tabfirst}29.46 & \cellcolor{tabfirst}0.917 & \cellcolor{tabfirst}0.182 & \cellcolor{tabsecond}29.04 & \cellcolor{tabsecond}0.911 & \cellcolor{tabsecond}0.185 \\
  & Kitchen & \cellcolor{tabfirst}32.16 & \cellcolor{tabfirst}0.932 & \cellcolor{tabfirst}0.122 & \cellcolor{tabsecond}31.29 & \cellcolor{tabsecond}0.927 & \cellcolor{tabsecond}0.123 \\
  & Bonsai & \cellcolor{tabfirst}32.58 & \cellcolor{tabfirst}0.947 & \cellcolor{tabfirst}0.189 & \cellcolor{tabsecond}31.89 & \cellcolor{tabsecond}0.941 & \cellcolor{tabsecond}0.194 \\
  & \textbf{Average} & \cellcolor{tabfirst}27.93 & \cellcolor{tabfirst}0.822 & \cellcolor{tabfirst}0.224 & \cellcolor{tabsecond}27.25 & \cellcolor{tabsecond}0.812 & \cellcolor{tabsecond}0.226 \\
 \midrule
 \multirow{3}{*}{\rotatebox{90}{\parbox{1.15cm}{\centering\scriptsize Tanks \& Temples}}} & Truck & \cellcolor{tabfirst}26.09 & \cellcolor{tabfirst}0.893 & \cellcolor{tabfirst}0.120 & \cellcolor{tabsecond}25.29 & \cellcolor{tabsecond}0.880 & \cellcolor{tabsecond}0.128 \\
  & Train & \cellcolor{tabfirst}22.74 & \cellcolor{tabsecond}0.832 & \cellcolor{tabsecond}0.192 & \cellcolor{tabsecond}22.56 & \cellcolor{tabfirst}0.835 & \cellcolor{tabfirst}0.190 \\
  & \textbf{Average} & \cellcolor{tabfirst}24.41 & \cellcolor{tabfirst}0.863 & \cellcolor{tabfirst}0.156 & \cellcolor{tabsecond}23.92 & \cellcolor{tabsecond}0.857 & \cellcolor{tabsecond}0.159 \\
 \midrule
 \multirow{3}{*}{\rotatebox{90}{\parbox{1.15cm}{\centering\scriptsize Deep Blending}}} & Dr Johnson & \cellcolor{tabfirst}30.04 & \cellcolor{tabfirst}0.913 & \cellcolor{tabfirst}0.237 & \cellcolor{tabsecond}29.34 & \cellcolor{tabsecond}0.905 & \cellcolor{tabsecond}0.245 \\
  & Playroom & \cellcolor{tabsecond}30.29 & \cellcolor{tabfirst}0.914 & \cellcolor{tabfirst}0.233 & \cellcolor{tabfirst}30.33 & \cellcolor{tabsecond}0.907 & \cellcolor{tabsecond}0.244 \\
  & \textbf{Average} & \cellcolor{tabfirst}30.16 & \cellcolor{tabfirst}0.913 & \cellcolor{tabfirst}0.235 & \cellcolor{tabsecond}29.84 & \cellcolor{tabsecond}0.906 & \cellcolor{tabsecond}0.244 \\
 \midrule
 \multirow{12}{*}{\rotatebox{90}{\parbox{1.15cm}{\centering\scriptsize DL3DV}}} & 140 & \cellcolor{tabfirst}27.76 & \cellcolor{tabfirst}0.896 & \cellcolor{tabfirst}0.184 & \cellcolor{tabsecond}27.24 & \cellcolor{tabsecond}0.884 & \cellcolor{tabsecond}0.195 \\
  & Car & \cellcolor{tabfirst}30.13 & \cellcolor{tabfirst}0.943 & \cellcolor{tabsecond}0.181 & \cellcolor{tabsecond}29.72 & \cellcolor{tabsecond}0.939 & \cellcolor{tabfirst}0.180 \\
  & Garden & \cellcolor{tabfirst}25.27 & \cellcolor{tabfirst}0.796 & \cellcolor{tabfirst}0.210 & \cellcolor{tabsecond}24.98 & \cellcolor{tabsecond}0.782 & \cellcolor{tabsecond}0.217 \\
  & Greenhouse & \cellcolor{tabfirst}22.37 & \cellcolor{tabfirst}0.762 & \cellcolor{tabfirst}0.246 & \cellcolor{tabsecond}21.32 & \cellcolor{tabsecond}0.709 & \cellcolor{tabsecond}0.291 \\
  & Grills & \cellcolor{tabfirst}27.44 & \cellcolor{tabfirst}0.922 & \cellcolor{tabfirst}0.211 & \cellcolor{tabsecond}26.09 & \cellcolor{tabsecond}0.909 & \cellcolor{tabsecond}0.226 \\
  & Herbary & \cellcolor{tabfirst}32.50 & \cellcolor{tabfirst}0.956 & \cellcolor{tabfirst}0.106 & \cellcolor{tabsecond}31.65 & \cellcolor{tabsecond}0.949 & \cellcolor{tabsecond}0.111 \\
  & Highrise & \cellcolor{tabfirst}27.51 & \cellcolor{tabfirst}0.907 & \cellcolor{tabfirst}0.139 & \cellcolor{tabsecond}26.92 & \cellcolor{tabsecond}0.897 & \cellcolor{tabsecond}0.149 \\
  & Roomset & \cellcolor{tabfirst}31.93 & \cellcolor{tabfirst}0.951 & \cellcolor{tabfirst}0.082 & \cellcolor{tabsecond}31.35 & \cellcolor{tabsecond}0.945 & \cellcolor{tabsecond}0.089 \\
  & Statue & \cellcolor{tabfirst}33.06 & \cellcolor{tabfirst}0.950 & \cellcolor{tabfirst}0.129 & \cellcolor{tabsecond}31.36 & \cellcolor{tabsecond}0.942 & \cellcolor{tabsecond}0.135 \\
  & Supermarket & \cellcolor{tabfirst}32.43 & \cellcolor{tabfirst}0.927 & \cellcolor{tabsecond}0.187 & \cellcolor{tabsecond}31.71 & \cellcolor{tabsecond}0.922 & \cellcolor{tabfirst}0.186 \\
  & Vasary & \cellcolor{tabfirst}25.86 & \cellcolor{tabfirst}0.835 & \cellcolor{tabfirst}0.211 & \cellcolor{tabsecond}25.00 & \cellcolor{tabsecond}0.814 & \cellcolor{tabsecond}0.231 \\
  & \textbf{Average} & \cellcolor{tabfirst}28.75 & \cellcolor{tabfirst}0.895 & \cellcolor{tabfirst}0.171 & \cellcolor{tabsecond}27.94 & \cellcolor{tabsecond}0.881 & \cellcolor{tabsecond}0.183 \\
\bottomrule
\end{tabular}}
\caption{Per-scene results at a budget of $\mathrm{500k}$ primitives; our method initialized from the EDGS point cloud.}
\label{tab:results-edgsinit-500000}
\end{table}

\begin{table}
    \centering
    \resizebox{\linewidth}{!}{%
    \begin{tabular}{cl|ccc|ccc}
    \toprule
      & & \multicolumn{3}{c}{Ours (iNGP init.)} & \multicolumn{3}{c}{ConeGS} \\
    \cmidrule(lr){3-5} \cmidrule(lr){6-8}
      & \textit{Scene} & PSNR & SSIM & LPIPS & PSNR & SSIM & LPIPS \\
    \midrule
     \multirow{10}{*}{\rotatebox{90}{\parbox{1.15cm}{\centering\scriptsize Mip-NeRF360}}} & Bicycle & \cellcolor{tabsecond}23.97 & \cellcolor{tabsecond}0.637 & \cellcolor{tabsecond}0.399 & \cellcolor{tabfirst}24.08 & \cellcolor{tabfirst}0.655 & \cellcolor{tabfirst}0.378 \\
      & Flowers & \cellcolor{tabfirst}20.68 & \cellcolor{tabfirst}0.511 & \cellcolor{tabsecond}0.471 & \cellcolor{tabsecond}20.59 & \cellcolor{tabsecond}0.509 & \cellcolor{tabfirst}0.465 \\
      & Garden & \cellcolor{tabfirst}24.85 & \cellcolor{tabfirst}0.711 & \cellcolor{tabsecond}0.346 & \cellcolor{tabsecond}24.68 & \cellcolor{tabsecond}0.709 & \cellcolor{tabfirst}0.336 \\
      & Stump & \cellcolor{tabfirst}25.87 & \cellcolor{tabfirst}0.712 & \cellcolor{tabsecond}0.351 & \cellcolor{tabsecond}25.79 & \cellcolor{tabsecond}0.709 & \cellcolor{tabfirst}0.349 \\
      & Treehill & \cellcolor{tabsecond}22.76 & \cellcolor{tabfirst}0.572 & \cellcolor{tabsecond}0.479 & \cellcolor{tabfirst}22.79 & \cellcolor{tabsecond}0.569 & \cellcolor{tabfirst}0.476 \\
      & Room & \cellcolor{tabfirst}31.17 & \cellcolor{tabfirst}0.904 & \cellcolor{tabsecond}0.260 & \cellcolor{tabsecond}30.95 & \cellcolor{tabsecond}0.903 & \cellcolor{tabfirst}0.254 \\
      & Counter & \cellcolor{tabfirst}28.25 & \cellcolor{tabfirst}0.882 & \cellcolor{tabsecond}0.251 & \cellcolor{tabsecond}28.11 & \cellcolor{tabsecond}0.879 & \cellcolor{tabfirst}0.251 \\
      & Kitchen & \cellcolor{tabfirst}30.12 & \cellcolor{tabsecond}0.895 & \cellcolor{tabsecond}0.197 & \cellcolor{tabsecond}29.82 & \cellcolor{tabfirst}0.897 & \cellcolor{tabfirst}0.183 \\
      & Bonsai & \cellcolor{tabfirst}30.66 & \cellcolor{tabsecond}0.917 & \cellcolor{tabsecond}0.255 & \cellcolor{tabsecond}30.63 & \cellcolor{tabfirst}0.920 & \cellcolor{tabfirst}0.241 \\
     & \textbf{Average} & \cellcolor{tabfirst}26.48 & \cellcolor{tabsecond}0.749 & \cellcolor{tabsecond}0.334 & \cellcolor{tabsecond}26.38 & \cellcolor{tabfirst}0.750 & \cellcolor{tabfirst}0.326 \\
     \midrule
     \multirow{3}{*}{\rotatebox{90}{\parbox{1.15cm}{\centering\scriptsize Tanks \& Temples}}} & Truck & \cellcolor{tabfirst}24.95 & \cellcolor{tabsecond}0.849 & \cellcolor{tabsecond}0.217 & \cellcolor{tabsecond}24.88 & \cellcolor{tabfirst}0.850 & \cellcolor{tabfirst}0.201 \\
      & Train & \cellcolor{tabfirst}22.10 & \cellcolor{tabfirst}0.780 & \cellcolor{tabfirst}0.276 & \cellcolor{tabsecond}21.85 & \cellcolor{tabsecond}0.772 & \cellcolor{tabsecond}0.279 \\
     & \textbf{Average} & \cellcolor{tabfirst}23.53 & \cellcolor{tabfirst}0.814 & \cellcolor{tabsecond}0.247 & \cellcolor{tabsecond}23.37 & \cellcolor{tabsecond}0.811 & \cellcolor{tabfirst}0.240 \\
     \midrule
     \multirow{3}{*}{\rotatebox{90}{\parbox{1.15cm}{\centering\scriptsize Deep Blending}}} & Dr Johnson & \cellcolor{tabfirst}29.39 & \cellcolor{tabfirst}0.894 & \cellcolor{tabfirst}0.288 & \cellcolor{tabsecond}29.35 & \cellcolor{tabsecond}0.891 & \cellcolor{tabsecond}0.293 \\
      & Playroom & \cellcolor{tabsecond}29.97 & \cellcolor{tabfirst}0.903 & \cellcolor{tabfirst}0.275 & \cellcolor{tabfirst}30.26 & \cellcolor{tabsecond}0.901 & \cellcolor{tabsecond}0.278 \\
     & \textbf{Average} & \cellcolor{tabsecond}29.68 & \cellcolor{tabfirst}0.898 & \cellcolor{tabfirst}0.282 & \cellcolor{tabfirst}29.80 & \cellcolor{tabsecond}0.896 & \cellcolor{tabsecond}0.285 \\
     \midrule
     \multirow{12}{*}{\rotatebox{90}{\parbox{1.15cm}{\centering\scriptsize DL3DV}}} & 140 & \cellcolor{tabsecond}26.55 & \cellcolor{tabsecond}0.865 & \cellcolor{tabsecond}0.245 & \cellcolor{tabfirst}26.60 & \cellcolor{tabfirst}0.867 & \cellcolor{tabfirst}0.239 \\
      & Car & \cellcolor{tabfirst}29.31 & \cellcolor{tabfirst}0.934 & \cellcolor{tabfirst}0.202 & \cellcolor{tabsecond}28.99 & \cellcolor{tabsecond}0.931 & \cellcolor{tabsecond}0.208 \\
      & Garden & \cellcolor{tabsecond}23.41 & \cellcolor{tabsecond}0.679 & \cellcolor{tabsecond}0.340 & \cellcolor{tabfirst}23.47 & \cellcolor{tabfirst}0.682 & \cellcolor{tabfirst}0.333 \\
      & Greenhouse & \cellcolor{tabsecond}20.31 & \cellcolor{tabsecond}0.635 & \cellcolor{tabfirst}0.373 & \cellcolor{tabfirst}20.39 & \cellcolor{tabfirst}0.635 & \cellcolor{tabsecond}0.373 \\
      & Grills & \cellcolor{tabfirst}26.09 & \cellcolor{tabfirst}0.898 & \cellcolor{tabsecond}0.263 & \cellcolor{tabsecond}25.84 & \cellcolor{tabsecond}0.898 & \cellcolor{tabfirst}0.259 \\
      & Herbary & \cellcolor{tabsecond}30.51 & \cellcolor{tabsecond}0.935 & \cellcolor{tabsecond}0.144 & \cellcolor{tabfirst}30.69 & \cellcolor{tabfirst}0.936 & \cellcolor{tabfirst}0.138 \\
      & Highrise & \cellcolor{tabfirst}25.89 & \cellcolor{tabsecond}0.861 & \cellcolor{tabsecond}0.209 & \cellcolor{tabsecond}25.80 & \cellcolor{tabfirst}0.864 & \cellcolor{tabfirst}0.203 \\
      & Roomset & \cellcolor{tabfirst}30.78 & \cellcolor{tabfirst}0.938 & \cellcolor{tabsecond}0.106 & \cellcolor{tabsecond}30.60 & \cellcolor{tabsecond}0.937 & \cellcolor{tabfirst}0.106 \\
      & Statue & \cellcolor{tabsecond}30.55 & \cellcolor{tabsecond}0.915 & \cellcolor{tabsecond}0.201 & \cellcolor{tabfirst}30.81 & \cellcolor{tabfirst}0.919 & \cellcolor{tabfirst}0.191 \\
      & Supermarket & \cellcolor{tabsecond}30.36 & \cellcolor{tabsecond}0.898 & \cellcolor{tabsecond}0.243 & \cellcolor{tabfirst}30.45 & \cellcolor{tabfirst}0.900 & \cellcolor{tabfirst}0.234 \\
      & Vasary & \cellcolor{tabsecond}24.53 & \cellcolor{tabfirst}0.786 & \cellcolor{tabsecond}0.287 & \cellcolor{tabfirst}24.60 & \cellcolor{tabsecond}0.786 & \cellcolor{tabfirst}0.279 \\
     & \textbf{Average} & \cellcolor{tabfirst}27.12 & \cellcolor{tabfirst}0.850 & \cellcolor{tabsecond}0.238 & \cellcolor{tabsecond}27.11 & \cellcolor{tabfirst}0.850 & \cellcolor{tabfirst}0.233 \\
    \bottomrule
    \end{tabular}}
    \caption{Per-scene results at a budget of $\mathrm{100k}$ primitives; our method initialized from the ConeGS (iNGP) point cloud. Both methods train for 30k iterations.}
    \label{tab:conegs-comparison-100k}
    \end{table}

\begin{table}
    \centering
    \resizebox{\linewidth}{!}{%
    \begin{tabular}{cl|ccc|ccc}
    \toprule
      & & \multicolumn{3}{c}{Ours (iNGP init.)} & \multicolumn{3}{c}{ConeGS} \\
    \cmidrule(lr){3-5} \cmidrule(lr){6-8}
      & \textit{Scene} & PSNR & SSIM & LPIPS & PSNR & SSIM & LPIPS \\
    \midrule
     \multirow{10}{*}{\rotatebox{90}{\parbox{1.15cm}{\centering\scriptsize Mip-NeRF360}}} & Bicycle & \cellcolor{tabfirst}25.30 & \cellcolor{tabfirst}0.760 & \cellcolor{tabsecond}0.260 & \cellcolor{tabsecond}25.26 & \cellcolor{tabsecond}0.758 & \cellcolor{tabfirst}0.242 \\
      & Flowers & \cellcolor{tabfirst}21.97 & \cellcolor{tabfirst}0.622 & \cellcolor{tabsecond}0.349 & \cellcolor{tabsecond}21.46 & \cellcolor{tabsecond}0.604 & \cellcolor{tabfirst}0.347 \\
      & Garden & \cellcolor{tabfirst}27.13 & \cellcolor{tabfirst}0.843 & \cellcolor{tabsecond}0.163 & \cellcolor{tabsecond}26.86 & \cellcolor{tabsecond}0.837 & \cellcolor{tabfirst}0.157 \\
      & Stump & \cellcolor{tabfirst}27.43 & \cellcolor{tabfirst}0.808 & \cellcolor{tabfirst}0.212 & \cellcolor{tabsecond}26.96 & \cellcolor{tabsecond}0.794 & \cellcolor{tabsecond}0.214 \\
      & Treehill & \cellcolor{tabfirst}23.22 & \cellcolor{tabfirst}0.639 & \cellcolor{tabfirst}0.367 & \cellcolor{tabsecond}22.79 & \cellcolor{tabsecond}0.608 & \cellcolor{tabsecond}0.372 \\
      & Room & \cellcolor{tabfirst}32.29 & \cellcolor{tabfirst}0.927 & \cellcolor{tabfirst}0.200 & \cellcolor{tabsecond}31.95 & \cellcolor{tabsecond}0.923 & \cellcolor{tabsecond}0.206 \\
      & Counter & \cellcolor{tabfirst}29.37 & \cellcolor{tabfirst}0.914 & \cellcolor{tabfirst}0.187 & \cellcolor{tabsecond}28.93 & \cellcolor{tabsecond}0.906 & \cellcolor{tabsecond}0.194 \\
      & Kitchen & \cellcolor{tabfirst}31.98 & \cellcolor{tabfirst}0.930 & \cellcolor{tabfirst}0.123 & \cellcolor{tabsecond}31.41 & \cellcolor{tabsecond}0.929 & \cellcolor{tabsecond}0.124 \\
      & Bonsai & \cellcolor{tabfirst}32.64 & \cellcolor{tabfirst}0.946 & \cellcolor{tabfirst}0.190 & \cellcolor{tabsecond}32.02 & \cellcolor{tabsecond}0.942 & \cellcolor{tabsecond}0.191 \\
     & \textbf{Average} & \cellcolor{tabfirst}27.93 & \cellcolor{tabfirst}0.821 & \cellcolor{tabsecond}0.228 & \cellcolor{tabsecond}27.52 & \cellcolor{tabsecond}0.811 & \cellcolor{tabfirst}0.227 \\
     \midrule
     \multirow{3}{*}{\rotatebox{90}{\parbox{1.15cm}{\centering\scriptsize Tanks \& Temples}}} & Truck & \cellcolor{tabfirst}26.06 & \cellcolor{tabfirst}0.890 & \cellcolor{tabfirst}0.124 & \cellcolor{tabsecond}25.71 & \cellcolor{tabsecond}0.883 & \cellcolor{tabsecond}0.124 \\
      & Train & \cellcolor{tabfirst}22.72 & \cellcolor{tabfirst}0.835 & \cellcolor{tabfirst}0.188 & \cellcolor{tabsecond}22.18 & \cellcolor{tabsecond}0.819 & \cellcolor{tabsecond}0.197 \\
     & \textbf{Average} & \cellcolor{tabfirst}24.39 & \cellcolor{tabfirst}0.862 & \cellcolor{tabfirst}0.156 & \cellcolor{tabsecond}23.95 & \cellcolor{tabsecond}0.851 & \cellcolor{tabsecond}0.161 \\
     \midrule
     \multirow{3}{*}{\rotatebox{90}{\parbox{1.15cm}{\centering\scriptsize Deep Blending}}} & Dr Johnson & \cellcolor{tabfirst}29.98 & \cellcolor{tabfirst}0.910 & \cellcolor{tabfirst}0.240 & \cellcolor{tabsecond}29.81 & \cellcolor{tabsecond}0.905 & \cellcolor{tabsecond}0.245 \\
      & Playroom & \cellcolor{tabsecond}30.17 & \cellcolor{tabfirst}0.909 & \cellcolor{tabfirst}0.236 & \cellcolor{tabfirst}30.60 & \cellcolor{tabsecond}0.908 & \cellcolor{tabsecond}0.241 \\
     & \textbf{Average} & \cellcolor{tabsecond}30.07 & \cellcolor{tabfirst}0.909 & \cellcolor{tabfirst}0.238 & \cellcolor{tabfirst}30.20 & \cellcolor{tabsecond}0.907 & \cellcolor{tabsecond}0.243 \\
     \midrule
     \multirow{12}{*}{\rotatebox{90}{\parbox{1.15cm}{\centering\scriptsize DL3DV}}} & 140 & \cellcolor{tabfirst}27.73 & \cellcolor{tabfirst}0.896 & \cellcolor{tabfirst}0.189 & \cellcolor{tabsecond}27.49 & \cellcolor{tabsecond}0.891 & \cellcolor{tabsecond}0.192 \\
      & Car & \cellcolor{tabfirst}30.13 & \cellcolor{tabfirst}0.943 & \cellcolor{tabfirst}0.181 & \cellcolor{tabsecond}29.74 & \cellcolor{tabsecond}0.940 & \cellcolor{tabsecond}0.187 \\
      & Garden & \cellcolor{tabfirst}25.38 & \cellcolor{tabfirst}0.800 & \cellcolor{tabsecond}0.204 & \cellcolor{tabsecond}25.36 & \cellcolor{tabsecond}0.799 & \cellcolor{tabfirst}0.200 \\
      & Greenhouse & \cellcolor{tabfirst}22.53 & \cellcolor{tabfirst}0.765 & \cellcolor{tabfirst}0.243 & \cellcolor{tabsecond}22.33 & \cellcolor{tabsecond}0.756 & \cellcolor{tabsecond}0.249 \\
      & Grills & \cellcolor{tabfirst}27.36 & \cellcolor{tabfirst}0.921 & \cellcolor{tabfirst}0.213 & \cellcolor{tabsecond}26.56 & \cellcolor{tabsecond}0.915 & \cellcolor{tabsecond}0.220 \\
      & Herbary & \cellcolor{tabfirst}32.43 & \cellcolor{tabfirst}0.956 & \cellcolor{tabfirst}0.108 & \cellcolor{tabsecond}32.20 & \cellcolor{tabsecond}0.953 & \cellcolor{tabsecond}0.109 \\
      & Highrise & \cellcolor{tabfirst}27.78 & \cellcolor{tabfirst}0.909 & \cellcolor{tabfirst}0.137 & \cellcolor{tabsecond}27.24 & \cellcolor{tabsecond}0.903 & \cellcolor{tabsecond}0.141 \\
      & Roomset & \cellcolor{tabfirst}31.71 & \cellcolor{tabfirst}0.949 & \cellcolor{tabfirst}0.082 & \cellcolor{tabsecond}31.38 & \cellcolor{tabsecond}0.947 & \cellcolor{tabsecond}0.085 \\
      & Statue & \cellcolor{tabfirst}33.57 & \cellcolor{tabfirst}0.951 & \cellcolor{tabfirst}0.133 & \cellcolor{tabsecond}32.96 & \cellcolor{tabsecond}0.947 & \cellcolor{tabsecond}0.136 \\
      & Supermarket & \cellcolor{tabfirst}32.44 & \cellcolor{tabfirst}0.927 & \cellcolor{tabsecond}0.190 & \cellcolor{tabsecond}32.12 & \cellcolor{tabsecond}0.924 & \cellcolor{tabfirst}0.190 \\
      & Vasary & \cellcolor{tabfirst}25.92 & \cellcolor{tabfirst}0.832 & \cellcolor{tabfirst}0.215 & \cellcolor{tabsecond}25.63 & \cellcolor{tabsecond}0.823 & \cellcolor{tabsecond}0.218 \\
     & \textbf{Average} & \cellcolor{tabfirst}28.82 & \cellcolor{tabfirst}0.895 & \cellcolor{tabfirst}0.172 & \cellcolor{tabsecond}28.46 & \cellcolor{tabsecond}0.891 & \cellcolor{tabsecond}0.175 \\
    \bottomrule
    \end{tabular}}
    \caption{Per-scene results at a budget of $\mathrm{500k}$ primitives; our method initialized from the ConeGS (iNGP) point cloud. Both methods train for 30k iterations.}
    \label{tab:conegs-comparison-500k}
    \end{table}

\begin{table*}[t]
\centering
\small
\setlength{\tabcolsep}{4pt}
\begin{tabular}{lcccccccccccc}
\toprule
 & \multicolumn{3}{c}{$\mathrm{100k}$} & \multicolumn{3}{c}{$\mathrm{500k}$} & \multicolumn{3}{c}{$\mathrm{1M}$} & \multicolumn{3}{c}{$\mathrm{2M}$} \\
\cmidrule(lr){2-4} \cmidrule(lr){5-7} \cmidrule(lr){8-10} \cmidrule(lr){11-13}
Share & PSNR & SSIM & LPIPS & PSNR & SSIM & LPIPS & PSNR & SSIM & LPIPS & PSNR & SSIM & LPIPS \\
\midrule
5\% & 26.24 & 0.747 & \textbf{0.333} & 27.72 & 0.814 & 0.231 & 27.99 & 0.826 & 0.203 & 28.09 & 0.831 & 0.186 \\
7.5\% & 26.28 & \textbf{0.748} & 0.334 & 27.78 & 0.817 & \textbf{0.230} & 28.04 & 0.829 & \textbf{0.201} & 28.14 & 0.834 & 0.183 \\
10\% (ours) & \textbf{26.32} & \textbf{0.748} & 0.336 & 27.84 & 0.819 & 0.231 & 28.10 & 0.831 & 0.202 & 28.21 & 0.835 & 0.183 \\
12.5\% & 26.31 & \textbf{0.748} & 0.337 & 27.85 & \textbf{0.820} & 0.231 & 28.13 & 0.832 & 0.202 & 28.24 & 0.838 & \textbf{0.182} \\
15\% & 26.30 & 0.746 & 0.339 & \textbf{27.86} & \textbf{0.820} & 0.233 & \textbf{28.15} & \textbf{0.833} & 0.202 & \textbf{28.27} & \textbf{0.839} & \textbf{0.182} \\
\bottomrule
\end{tabular}
\caption{\textbf{Relocation Share}: Our method with different shares of primitives relocated per relocation step before the share falls off from iteration 10k (Section~\ref{sec:method:population}; 10\% in all other experiments), as means over the 9 Mip-NeRF 360 scenes~\cite{barron2022mipnerf360}. Bold marks the best value in each column.}
\label{tab:explore}
\end{table*}

\begin{table}[t]
\centering
\resizebox{\linewidth}{!}{%
\begin{tabular}{llcccc}
\toprule
 & & \multicolumn{4}{c}{$\Delta$ Ours $-$ 3DGS-MCMC} \\
\cmidrule(lr){3-6}
\textit{Metric} & \textit{Region} & $\mathrm{100k}$ & $\mathrm{500k}$ & $\mathrm{1M}$ & $\mathrm{2M}$ \\
\midrule
\multirow{2}{*}{PSNR $\uparrow$} & Background & \cellcolor[RGB]{255,181,181}$+1.34$ & \cellcolor[RGB]{255,179,179}$+1.38$ & \cellcolor[RGB]{255,189,189}$+1.20$ & \cellcolor[RGB]{255,198,198}$+1.04$ \\
 & Foreground & \cellcolor[RGB]{255,230,230}$+0.45$ & \cellcolor[RGB]{255,235,235}$+0.36$ & \cellcolor[RGB]{255,241,241}$+0.25$ & \cellcolor[RGB]{255,245,245}$+0.18$ \\
\midrule
\multirow{2}{*}{SSIM $\uparrow$} & Background & \cellcolor[RGB]{255,179,179}$+0.015$ & \cellcolor[RGB]{255,194,194}$+0.012$ & \cellcolor[RGB]{255,210,210}$+0.009$ & \cellcolor[RGB]{255,221,221}$+0.007$ \\
 & Foreground & \cellcolor[RGB]{255,204,204}$+0.010$ & \cellcolor[RGB]{255,223,223}$+0.006$ & \cellcolor[RGB]{255,235,235}$+0.004$ & \cellcolor[RGB]{255,247,247}$+0.002$ \\
\midrule
\multirow{2}{*}{LPIPS $\downarrow$} & Background & \cellcolor[RGB]{255,184,184}$-0.024$ & \cellcolor[RGB]{255,179,179}$-0.025$ & \cellcolor[RGB]{255,190,190}$-0.022$ & \cellcolor[RGB]{255,200,200}$-0.018$ \\
 & Foreground & \cellcolor[RGB]{255,218,218}$-0.012$ & \cellcolor[RGB]{255,206,206}$-0.016$ & \cellcolor[RGB]{255,216,216}$-0.013$ & \cellcolor[RGB]{255,228,228}$-0.009$ \\
\bottomrule
\end{tabular}}
\caption{\textbf{Background vs. Foreground, Indoor}: the same comparison as Table~\ref{tab:background} on the indoor Mip-NeRF 360 scenes. The background of each scene is defined as the farthest $25\%$ of all its test pixels by Depth Anything V2~\cite{yang2024depthanythingv2} depth of the ground-truth test images, aligned per image to the observed COLMAP~\cite{schoenberger2016sfm,schoenberger2016mvs} points. We also exclude a $\pm 5\%$ quantile band around the cut and eroding both regions by $3$ pixels. We report the differences of the means over room, counter, kitchen and bonsai.}
\label{tab:background-indoor}
\end{table}

\providecommand{\cmpdir}{figures/varbudget}
\newlength{\cmpw}
\newlength{\cmph}

\newcommand{\cmpscene}[7]{%
  \begin{scope}[yshift={-#1*(2*\cmph+7.5pt)}]
    \node (view) at (0, 0) {\includegraphics[width=\cmpw]{\cmpdir/#2/view.jpg}};
    \node (gt) at (0, -1) {\includegraphics[width=\cmpw]{\cmpdir/#2/gt.jpg}};
    \draw[red, thick, shift={(view.south west)}, x=\cmpw, y=\cmph] (#5) rectangle +(#6, #6);
    \path (gt.south west) -- (view.north west) node[midway, sloped, above=3pt, font=\footnotesize, text depth=0.25ex]
      {#3 \textcolor{black!55}{(#4)}};
    \foreach \m/\base/\ours [count=\i] in {#7} {
      \node (b) at (\i, 0) {\includegraphics[width=\cmpw]{\cmpdir/#2/\m_base.jpg}};
      \node[tag] at (b.south west) {\base};
      \node (o) at (\i, -1) {\includegraphics[width=\cmpw]{\cmpdir/#2/\m_ours.jpg}};
      \node[tag] at (o.south west) {Ours \ours};
    }
  \end{scope}}

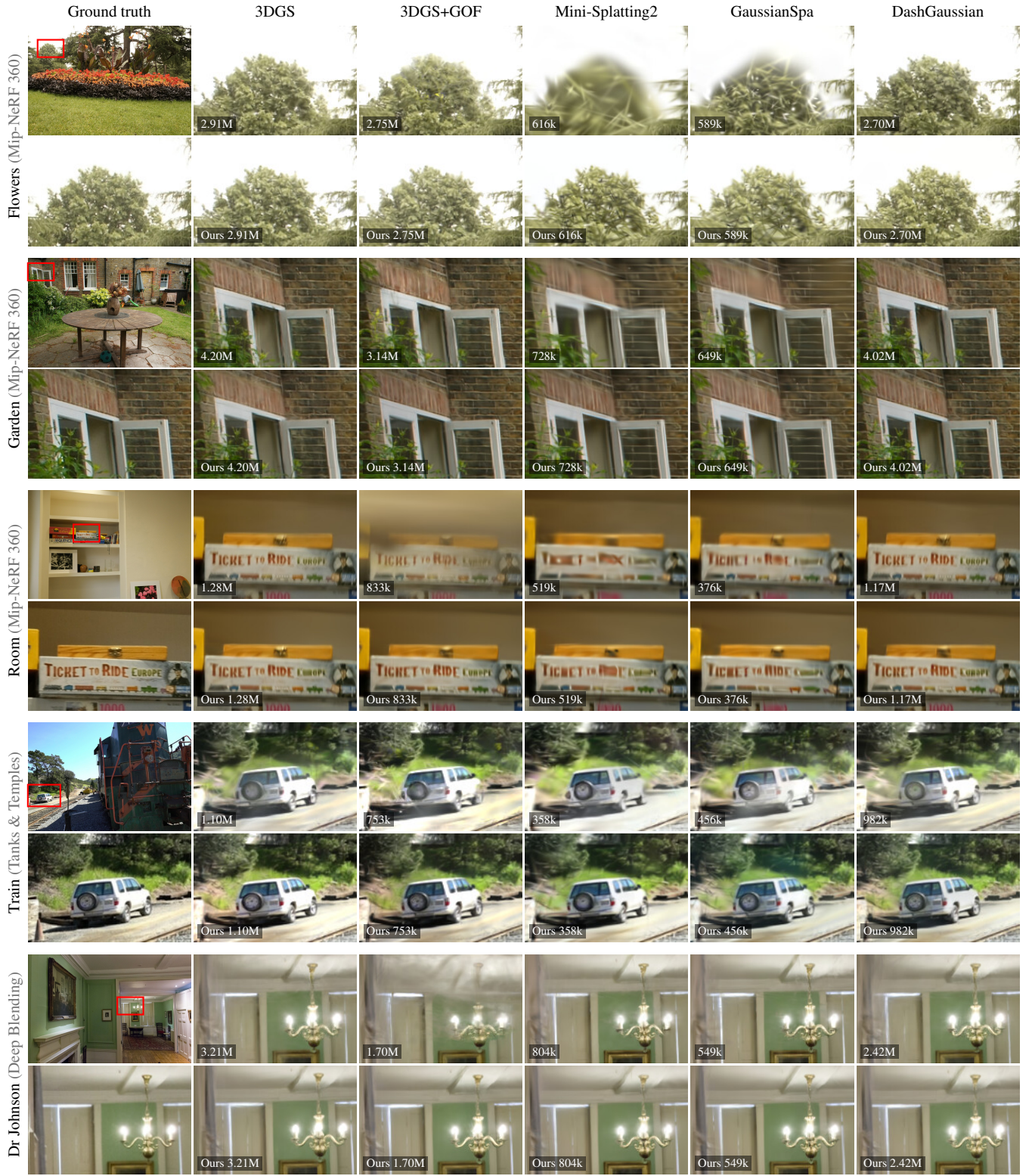
\begin{figure*}[p]
\begin{minipage}[c][\textheight][c]{\textwidth}
\centering
\setlength{\cmpw}{\dimexpr(\textwidth-1.2em-7.5pt)/6\relax}
\setlength{\cmph}{\dimexpr\cmpw*2/3\relax}
\begin{tikzpicture}[x=\cmpw+1.5pt, y=\cmph+1.5pt, every node/.style={inner sep=0pt},
    tag/.style={anchor=south west, xshift=2pt, yshift=2pt, inner sep=1.5pt, fill=black, fill opacity=0.5,
                text opacity=1, text=white, font=\scriptsize}]
  \foreach \name [count=\i from 0] in {Ground truth, 3DGS, 3DGS+GOF, Mini-Splatting2, GaussianSpa, DashGaussian}
    \node[above=3pt, font=\footnotesize, text depth=0.25ex] at (\i, 0.5) {\name};
  \cmpscene{0}{flowers}{Flowers}{Mip-NeRF 360}{0.06,0.72}{0.16}
    {3dgs/2.91M/2.91M, gof/2.75M/2.75M, msv2/616k/616k, spa/589k/589k, dash/2.70M/2.70M}
  \cmpscene{1}{garden}{Garden}{Mip-NeRF 360}{0.00,0.80}{0.16}
    {3dgs/4.20M/4.20M, gof/3.14M/3.14M, msv2/728k/728k, spa/649k/649k, dash/4.02M/4.02M}
  \cmpscene{2}{room}{Room}{Mip-NeRF 360}{0.28,0.53}{0.16}
    {3dgs/1.28M/1.28M, gof/833k/833k, msv2/519k/519k, spa/376k/376k, dash/1.17M/1.17M}
  \cmpscene{3}{train}{Train}{Tanks \& Temples}{0.00,0.23}{0.2}
    {3dgs/1.10M/1.10M, gof/753k/753k, msv2/358k/358k, spa/456k/456k, dash/982k/982k}
  \cmpscene{4}{drjohnson}{Dr Johnson}{Deep Blending}{0.55,0.45}{0.16}
    {3dgs/3.21M/3.21M, gof/1.70M/1.70M, msv2/804k/804k, spa/549k/549k, dash/2.42M/2.42M}
\end{tikzpicture}
\caption{\textbf{Comparison with variable-budget methods.} For each scene, the top row shows the baseline and
the bottom row ours trained with the primitive budget of that baseline.}
\label{fig:supp-variable-budget}
\end{minipage}
\end{figure*}

\begin{table*}[t]
\centering
\small
\setlength{\tabcolsep}{4pt}
\begin{tabular}{llcccccccccccc}
\toprule
 &  & \multicolumn{3}{c}{$\mathrm{100k}$} & \multicolumn{3}{c}{$\mathrm{500k}$} & \multicolumn{3}{c}{$\mathrm{1M}$} & \multicolumn{3}{c}{$\mathrm{2M}$} \\
\cmidrule(lr){3-5} \cmidrule(lr){6-8} \cmidrule(lr){9-11} \cmidrule(lr){12-14}
Dataset & Method & PSNR & SSIM & LPIPS & PSNR & SSIM & LPIPS & PSNR & SSIM & LPIPS & PSNR & SSIM & LPIPS \\
\midrule
\multirow{2}{*}{MipNeRF 360} & Ours & \textbf{26.32} & \textbf{0.747} & \textbf{0.336} & \textbf{27.84} & \textbf{0.819} & \textbf{0.231} & \textbf{28.10} & \textbf{0.831} & \textbf{0.202} & \textbf{28.21} & \textbf{0.835} & \textbf{0.183} \\
 & Revising & 24.24 & 0.691 & 0.379 & 26.57 & 0.768 & 0.273 & 26.93 & 0.788 & 0.238 & 27.13 & 0.800 & 0.213 \\
\midrule
\multirow{2}{*}{Tanks\&Temples} & Ours & \textbf{23.47} & \textbf{0.815} & \textbf{0.251} & \textbf{24.30} & \textbf{0.860} & \textbf{0.164} & \textbf{24.62} & \textbf{0.868} & \textbf{0.142} & \textbf{24.59} & \textbf{0.871} & \textbf{0.129} \\
 & Revising & 20.84 & 0.760 & 0.298 & 23.06 & 0.832 & 0.193 & 23.47 & 0.844 & 0.168 & 23.78 & 0.852 & 0.153 \\
\midrule
\multirow{2}{*}{Deep Blending} & Ours & \textbf{29.55} & \textbf{0.899} & \textbf{0.283} & \textbf{30.06} & \textbf{0.912} & \textbf{0.241} & \textbf{30.03} & \textbf{0.911} & \textbf{0.230} & \textbf{30.16} & \textbf{0.911} & \textbf{0.223} \\
 & Revising & 29.34 & 0.894 & 0.294 & 29.90 & 0.904 & 0.254 & 29.92 & 0.905 & 0.243 & 29.90 & 0.905 & 0.235 \\
\bottomrule
\end{tabular}
\caption{\textbf{Comparison with Revising Densification}: Our method against Revising Densification at Gaussian budgets of 100k, 500k, 1M and 2M, as means over the 9 Mip-NeRF 360 scenes, the 2 Tanks\&Temples scenes (Train, Truck) and the 2 Deep Blending scenes (Playroom, DrJohnson). Bold marks the best value in each column pair.}
\label{tab:revising_comparison}
\end{table*}

\end{document}